\documentclass{article}

\def\VenueIsCOLM{}
\makeatletter
\@ifundefined{IsPreprint}{%
  \@ifundefined{IsFinal}{\def\IsFinal{}}{}%
}{}
\makeatother

\ifdefined\IsFinal
  \usepackage[final]{venues/colm2026/colm2026_conference}
\else\ifdefined\IsPreprint
  \usepackage[preprint]{venues/colm2026/colm2026_conference}
\else
  \usepackage[submission]{venues/colm2026/colm2026_conference}
\fi\fi

\usepackage{microtype}
\usepackage{hyperref}
\usepackage{url}
\usepackage{amssymb} %
\usepackage{lineno}
\usepackage{placeins}  %
\usepackage{afterpage}  %
\usepackage{tabularx} %

\newcommand{\PaperTitle}{Capability Provenance in Language Models:\\
A Case Study in Social Reasoning}

\usepackage{amsmath}
\usepackage{amssymb}       %
\usepackage{booktabs}
\usepackage{makecell}
\usepackage{multirow}
\usepackage{xfp}
\usepackage{siunitx}
\usepackage{graphicx}
\usepackage{twemojis}     %
\usepackage{fontawesome5} %

\newcommand{\PaperWebsiteURL}{https://hcai-lab-gt.github.io/capabilibara/}
\newcommand{\PaperCodeURL}{https://github.com/HCAI-Lab-GT/capabilibara}
\newcommand{\PaperHFURL}{https://huggingface.co/spaces/HCAI-Lab-GT/capabilibara}
\newcommand{\PaperDiscordURL}{https://discord.gg/ZnDx35F3z7}
\usepackage{stfloats}
\usepackage{placeins}      %
\usepackage{enumitem}
\usepackage{tikz}
\usetikzlibrary{arrows.meta,positioning,fit}
\usepackage{subcaption}

\ifdefined\DeclareUnicodeCharacter
  \DeclareUnicodeCharacter{0019}{'}
\fi

\usepackage{xspace}      %
\usepackage{xcolor}      %
\usepackage{pifont}      %
\usepackage{enumitem}    %

\newcommand{\eg}{e.g.,\xspace}      %
\providecommand{\num}[1]{#1}

\newif\ifdraft
\drafttrue  %

\usepackage{mlpaper_visual_style}
\definecolor{SocialTDAPurple}{HTML}{762A83}
\definecolor{SocialTDAGold}{HTML}{8C6D1F}
\definecolor{SocialTDAOrange}{HTML}{A14F00}
\definecolor{SocialTDAGreen}{HTML}{1B7837}
\definecolor{SocialTDATeal}{HTML}{006D5B}
\definecolor{SocialTDASlate}{HTML}{4B5563}
\definecolor{SocialTDAUnlearning}{HTML}{7A5638}
\definecolor{SocialTDAGray}{HTML}{575757}
\definecolor{SocialTDAControlGray}{HTML}{737373}
\definecolor{SocialTDAPositive}{HTML}{4575B4}
\definecolor{SocialTDAPositiveLight}{HTML}{67A9CF}
\definecolor{SocialTDANeutral}{HTML}{F7F7F7}
\definecolor{SocialTDANegativeLight}{HTML}{EF8A62}
\definecolor{SocialTDANegative}{HTML}{D73027}
\definecolor{SocialTDADark}{HTML}{222222}
\definecolor{SocialTDALightGray}{HTML}{F5F5F5}
\definecolor{SocialTDABorderGray}{HTML}{D0D0D0}
\definecolor{MLBlue}{HTML}{4575B4}
\definecolor{MLGreen}{HTML}{1B7837}
\definecolor{MLAmber}{HTML}{A14F00}
\definecolor{MLRed}{HTML}{D73027}
\definecolor{MLPurple}{HTML}{762A83}
\definecolor{MLTeal}{HTML}{006D5B}
\definecolor{MLSlate}{HTML}{4B5563}
\definecolor{MLBrown}{HTML}{7A5638}
\definecolor{MLGray}{HTML}{575757}
\definecolor{MLLightGray}{HTML}{F5F5F5}
\definecolor{MLBorderGray}{HTML}{D0D0D0}
\tikzset{
  mlpv/socialreasoning/.style={draw=SocialTDAPurple!70!black, fill=SocialTDAPurple!10, text=black},
  mlpv/socialknowledge/.style={draw=SocialTDAGold!70!black, fill=SocialTDAGold!10, text=black},
  mlpv/stemreasoning/.style={draw=SocialTDAOrange!70!black, fill=SocialTDAOrange!10, text=black},
  mlpv/stemknowledge/.style={draw=SocialTDAGreen!70!black, fill=SocialTDAGreen!10, text=black},
  mlpv/attribution/.style={draw=SocialTDASlate!70!black, fill=SocialTDASlate!10, text=black},
  mlpv/methodstage/.style={draw=SocialTDASlate!70!black, fill=SocialTDASlate!10, text=black},
  mlpv/influencepositive/.style={draw=SocialTDAPositive!70!black, fill=SocialTDAPositive!10, text=black},
  mlpv/influencenegative/.style={draw=SocialTDANegative!70!black, fill=SocialTDANegative!10, text=black},
  mlpv/influenceneutral/.style={draw=SocialTDAGray!70!black, fill=SocialTDAGray!10, text=black},
  mlpv/validationtargeted/.style={draw=SocialTDAUnlearning!70!black, fill=SocialTDAUnlearning!10, text=black},
  mlpv/validationcontrol/.style={draw=SocialTDAControlGray!70!black, fill=SocialTDAControlGray!10, text=black},
}

\NewDocumentCommand{\SocialReasoning}{m}{\socialreasoningconcept{#1}}
\NewDocumentCommand{\SocialKnowledge}{m}{\socialknowledgeconcept{#1}}
\NewDocumentCommand{\STEMReasoning}{m}{\stemreasoningconcept{#1}}
\NewDocumentCommand{\STEMKnowledge}{m}{\stemknowledgeconcept{#1}}
\NewDocumentCommand{\AttributionConcept}{m}{\attributionconcept{#1}}

\NewDocumentCommand{\TargetedValidation}{m}{\validationtargetedconcept{#1}}
\NewDocumentCommand{\ControlValidation}{m}{\validationcontrolconcept{#1}}

\NewDocumentCommand{\SocialTDABadge}{O{SocialTDASlate} m}{%
  \tcbox[on line, boxsep=0.8pt, left=2pt, right=2pt, top=0.35pt,
    bottom=0.35pt, arc=0.8pt, boxrule=.25pt, colback=#1!5,
    colframe=#1!62!black]{\scriptsize\textsf{#2}}%
}

\NewDocumentCommand{\ValidationBadge}{}{\SocialTDABadge[SocialTDAUnlearning]{Unlearning validation}}

\NewDocumentCommand{\RQBadge}{m}{\SocialTDABadge{RQ#1}}

\NewDocumentCommand{\PositiveAttribution}{m}{\textcolor{SocialTDAPositive}{\textbf{#1}}}
\NewDocumentCommand{\NegativeAttribution}{m}{\textcolor{SocialTDANegative}{\textbf{#1}}}

\NewDocumentCommand{\VisualSwatch}{m}{%
  \raisebox{0.08ex}{%
    \tikz\filldraw[draw=#1!72!black, fill=#1!70, line width=0.25pt]
      (0,0) rectangle (0.72em,0.72em);%
  }%
}

\NewDocumentCommand{\ValidationTargetMarker}{}{%
  \raisebox{0.08ex}{%
    \tikz\filldraw[draw=SocialTDAUnlearning!85!black, fill=SocialTDAUnlearning,
      line width=0.25pt] (0,0) rectangle (0.72em,0.72em);%
  }%
}
\NewDocumentCommand{\ValidationControlMarker}{}{%
  \raisebox{0.05ex}{%
    \tikz\draw[draw=SocialTDAControlGray, fill=white, line width=0.55pt]
      (0.36em,0.36em) circle (0.36em);%
  }%
}
\NewDocumentCommand{\LegendItem}{m m}{\mbox{\VisualSwatch{#1}\hspace{0.35em}\textsf{#2}}}
\NewDocumentCommand{\LegendMarkerItem}{m m}{\mbox{#1\hspace{0.35em}\textsf{#2}}}

\newenvironment{SocialTDAFigureLegend}{%
  \par\smallskip
  \begin{center}
  \begin{minipage}{0.98\linewidth}
  \footnotesize\centering
}{%
  \end{minipage}
  \end{center}
  \vspace{-2pt}%
}

\NewDocumentCommand{\InfluenceSignLegend}{}{%
  \begin{SocialTDAFigureLegend}
    \begin{tabular}{@{}c@{}}
    \textbf{Influence sign:}\quad
    \LegendItem{SocialTDAPositive}{Positive or supportive}\quad
    \LegendItem{SocialTDANegative}{Negative or suppressive}\quad
    \LegendItem{SocialTDAGray}{Near zero or neutral}
    \end{tabular}
  \end{SocialTDAFigureLegend}%
}

\NewDocumentCommand{\GammaHeatmapSignLegend}{}{%
  \begin{SocialTDAFigureLegend}
    \begin{tabular}{@{}c@{}}
    \textbf{Validation outcome:}\quad
    \LegendItem{SocialTDAPositive}{Accuracy damage ($\gamma>0$)}\quad
    \LegendItem{SocialTDANegative}{Accuracy increase ($\gamma<0$)}
    \end{tabular}
  \end{SocialTDAFigureLegend}%
}

\NewDocumentCommand{\GammaSignLegend}{}{%
  \begin{SocialTDAFigureLegend}
    \begin{tabular}{@{}c@{}}
    \textbf{Validation outcome:}\quad
    \LegendItem{SocialTDAUnlearning}{Accuracy damage}\quad
    \LegendItem{SocialTDAControlGray}{Accuracy increase}\quad
    \LegendItem{SocialTDALightGray}{Near zero}
    \end{tabular}
  \end{SocialTDAFigureLegend}%
}

\NewDocumentCommand{\ValidationLegend}{}{%
  \begin{SocialTDAFigureLegend}
    \begin{tabular}{@{}c@{}}
    \textbf{Unlearning contrast:}\quad
    \LegendMarkerItem{\ValidationTargetMarker}{Influence-targeted}\quad
    \LegendMarkerItem{\ValidationControlMarker}{Random or control}
    \end{tabular}
  \end{SocialTDAFigureLegend}%
}

\NewDocumentCommand{\EfficiencyLegend}{}{%
  \begin{SocialTDAFigureLegend}
    \begin{tabular}{@{}c@{}}
    \textbf{Efficiency sign:}\quad
    \LegendItem{SocialTDAUnlearning}{Faster with influence-targeted selection}\quad
    \LegendItem{SocialTDAControlGray}{Slower than random baseline}\\[-1pt]
    \LegendItem{SocialTDALightGray}{Near zero}
    \end{tabular}
  \end{SocialTDAFigureLegend}%
}

\NewDocumentCommand{\TopicControlLegend}{}{%
\begin{SocialTDAFigureLegend}
  \begin{tabular}{@{}c@{}}
  \LegendItem{SocialTDAExpOne}{Single-bin topic runs ($n=\WebOrgNumTopics{}$; dots\,=\,individual topics)}\quad
  \LegendMarkerItem{{\large$\bigstar$}}{Global random control}
  \end{tabular}
\end{SocialTDAFigureLegend}%
}

\NewDocumentCommand{\BenchmarkRoleTableCell}{m m m}{%
  \cellcolor{#1!7}%
  \begin{minipage}[c][2.45em][c]{\linewidth}
    \centering
    \textbf{#2}\\[-1pt]{\scriptsize\textcolor{#1!88!black}{#3}}%
  \end{minipage}%
}
\NewDocumentCommand{\SocialReasoningTableCell}{}{\BenchmarkRoleTableCell{SocialTDAPurple}{SocialIQA}{social reasoning}}
\NewDocumentCommand{\SocialKnowledgeTableCell}{}{\BenchmarkRoleTableCell{SocialTDAGold}{MMLU Social Sciences}{social knowledge}}
\NewDocumentCommand{\STEMReasoningTableCell}{}{\BenchmarkRoleTableCell{SocialTDAOrange}{ARC-Challenge}{STEM reasoning}}
\NewDocumentCommand{\STEMKnowledgeTableCell}{}{\BenchmarkRoleTableCell{SocialTDAGreen}{MMLU STEM}{STEM knowledge}}
\NewDocumentCommand{\BenchmarkTargetTableCell}{m m}{%
  \cellcolor{#1!7}%
  \begin{minipage}[c][2.6em][c]{\linewidth}
    \raggedright\textcolor{#1!88!black}{\textbf{#2}}%
  \end{minipage}%
}
\newcommand{\RQProfileHeader}[1]{\rowcolor{#1!8}}
\newcommand{\RQProfileGroupRow}[2][11]{%
  \ifnum#1=8
    \RQProfileGroupRowEight{#2}%
  \else
    \RQProfileGroupRowEleven{#2}%
  \fi
}
\newcommand{\RQProfileGroupRowEight}[1]{%
  & \textcolor{SocialTDASlate!85!black}{\textit{#1}} & & & & & &\\[-1pt]%
}
\newcommand{\RQProfileGroupRowEleven}[1]{%
  & \textcolor{SocialTDASlate!85!black}{\textit{#1}} & & & & & & & & &\\[-1pt]%
}

\newenvironment{SocialTDATakeaway}[1]{%
  \begin{tcolorbox}[mlboxbase, colback=SocialTDASlate!4,
    colframe=SocialTDASlate!68!black, unbreakable, top=3pt, bottom=3pt,
    before skip=4pt, after skip=4pt, title={#1}]%
}{%
  \end{tcolorbox}%
}

\newenvironment{SocialTDADesignBox}[1]{%
  \begin{tcolorbox}[mlboxbase, colback=SocialTDALightGray,
    colframe=SocialTDASlate!62!black, unbreakable, top=3pt, bottom=3pt,
    before skip=4pt, after skip=4pt, title={#1}]%
}{%
  \end{tcolorbox}%
}

\newenvironment{SocialTDAAppendixCallout}[1]{%
  \begin{tcolorbox}[mlboxbase, colback=SocialTDASlate!3,
    colframe=SocialTDASlate!62!black, title={#1}]%
}{%
  \end{tcolorbox}%
}

\newenvironment{SocialTDACallout}[1]{%
  \begin{SocialTDAAppendixCallout}{#1}%
}{%
  \end{SocialTDAAppendixCallout}%
}

\newcommand{\CorpusRawSlots}{3.87B}                     %
\newcommand{\CorpusUniqueDocs}{\num{1263587168}}        %
\newcommand{\CorpusUniqueDocsApprox}{1.26B}             %
\newcommand{\CorpusBinnedDocs}{\num{1098646162}}        %
\newcommand{\CorpusDedupRate}{67.3}                     %
\newcommand{\CorpusTrainingTokens}{6T}                  %
\newcommand{\CorpusNumShards}{\num{65718}}              %

\newcommand{\WebOrgNumFormats}{24}
\newcommand{\WebOrgNumTopics}{24}
\newcommand{\WebOrgNumBins}{\fpeval{\WebOrgNumFormats*\WebOrgNumTopics}}
\newcommand{\BergsonVersion}{0.9.0}

\newcommand{\AttrDocCount}{\num{5678621}}
\newcommand{\AttrTokensB}{10.5}
\newcommand{\AttrDocsPerBin}{\num{10000}}
\newcommand{\AttrShardCount}{316}                      %
\newcommand{\AttrDocsPerShard}{\num{17996}}
\newcommand{\AttrGPUHoursExplore}{156.1}
\newcommand{\AttrBinsUnderfilled}{17}
\newcommand{\PrecondSampleSize}{\num{100000}}

\newcommand{\ComputeRefGPU}{H200 144\,GB}

\newcommand{\ComputeEnrichRaw}{\num{10000}}             %
\newcommand{\ComputeEnrichEquiv}{\num{8000}}            %
\newcommand{\ComputeAttrIndexRaw}{\num{20000}}
\newcommand{\ComputeAttrIndexEquiv}{\num{20000}}
\newcommand{\ComputeAttrQueryRaw}{20}
\newcommand{\ComputeAttrQueryEquiv}{20}
\newcommand{\ComputeAttrPrecondRaw}{192}                %
\newcommand{\ComputeAttrPrecondEquiv}{192}              %
\newcommand{\ComputeAttrCalibRaw}{\num{1500}}
\newcommand{\ComputeAttrCalibEquiv}{\num{1500}}
\newcommand{\ComputeUnlearnRaw}{\num{6000}}
\newcommand{\ComputeUnlearnEquiv}{\num{6000}}
\newcommand{\ComputeEvalRaw}{600}
\newcommand{\ComputeEvalEquiv}{500}
\newcommand{\ComputeTotalRaw}{\num{40000}}
\newcommand{\ComputeTotalEquiv}{\num{37000}}

\newcommand{\MainFigureWidth}{0.98\linewidth}

\newcommand{\SingleHeatmapFigureWidth}{\MainFigureWidth}
\newcommand{\StackedHeatmapPanelWidth}{0.88\linewidth}
\newcommand{\StackedBarPanelWidth}{\MainFigureWidth}

\newcommand{\TallFigureMaxHeight}{0.78\textheight}

\newcommand{\RQOneFigureWidth}{\MainFigureWidth}

\usepackage{titletoc} %

\newif\ifMarksVisible
\MarksVisibletrue
\ifdefined\MarksHidden\MarksVisiblefalse\fi

\ifdefined\IsPreprint \def\IsNamed{} \fi
\ifdefined\IsFinal \def\IsNamed{} \fi

\ifMarksVisible
  \newcommand{\Mark}[1]{{\scriptsize\textcolor{blue}{[Mark: #1]}}}
  \newcommand{\MarkLeft}[1]{{\scriptsize\textcolor{blue}{[$\leftarrow$Mark: #1]}}}
  \newcommand{\MarkRight}[1]{{\scriptsize\textcolor{blue}{[Mark: #1$\rightarrow$]}}}
\else
  \newcommand{\Mark}[1]{}
  \newcommand{\MarkLeft}[1]{}
  \newcommand{\MarkRight}[1]{}
\fi

\ifdefined\IsDraft
  \usepackage{eso-pic}
  \AddToShipoutPictureBG{%
    \begin{tikzpicture}[remember picture, overlay]%
      \foreach \i in {0,1,2,3}{%
        \foreach \j in {0,1,2,3,4}{%
          \node[rotate=30, text=black!8, font=\large\bfseries, align=center]
            at ([xshift={\i*7cm-2cm}, yshift={-\j*5.75cm+1cm}]current page.north west)
            {UNDER REVIEW\\DO NOT DISTRIBUTE};%
        }%
      }%
    \end{tikzpicture}%
  }
\fi

\definecolor{darkblue}{rgb}{0, 0, 0.5}
\hypersetup{colorlinks=true, citecolor=darkblue, linkcolor=darkblue, urlcolor=darkblue,
  pdftitle={Capability Provenance in Language Models: A Case Study in Social Reasoning},
  pdfauthor={Glenn Matlin, Chandreyi Chakraborty, Saehee Eom, Mika Okamoto, Rayan Castilla, Louis Jaburi, Alvin Deng, Taywon Min, Lucia Quirke, Stella Biderman, Mark Riedl}}

\title{\PaperTitle}

\ifdefined\IsNamed
  \ifdefined\VenueIsCOLM
    \newcommand{\AffGT}{\twemoji{bee}}          %
    \newcommand{\AffMATS}{\twemoji{fire}}       %
    \newcommand{\AffEleuther}{\twemoji{dove}}   %
    \newcommand{\AffKAIST}{\twemoji{robot}}     %
    \newcommand{\AffGTSafety}{\twemoji{shield}} %
    \author{%
      Glenn Matlin$^{\AffGT\,\AffMATS\,\AffGTSafety}$\thanks{Corresponding author: \texttt{glenn@gatech.edu}} \quad
      Chandreyi Chakraborty$^{\AffGT}$ \\
      Saehee Eom$^{\AffGT\,\dagger}$ \quad Mika Okamoto$^{\AffGT\,\dagger}$ \quad Rayan Castilla$^{\AffGT\,\dagger}$ \\
      Louis Jaburi$^{\AffEleuther}$ \quad Alvin Deng$^{\AffEleuther}$ \quad
      Taywon Min$^{\AffKAIST\,\AffMATS}$ \quad Lucia Quirke$^{\AffEleuther}$ \\
      Stella Biderman$^{\AffEleuther}$ \quad Mark Riedl$^{\AffGT}$ \\[4pt]
      \AffGT\,Georgia Institute of Technology, College of Computing \quad
      \AffMATS\,MATS Program \\
      \AffEleuther\,EleutherAI \quad \AffKAIST\,KAIST AI \quad
      \AffGTSafety\,Georgia Tech AI Safety Initiative \\[2pt]
      {\small $^{\dagger}$Equal contribution} \\[5pt]
      {\small
        \href{\PaperWebsiteURL}{\faGlobe~Website} \quad
        \href{\PaperCodeURL}{\faGithub~Code} \quad
        \href{\PaperHFURL}{\twemoji{1f917}~HuggingFace} \quad
        \href{\PaperDiscordURL}{\faDiscord~Discord}}%
    }
  \fi
\else
  \ifdefined\VenueIsCOLM
    \author{Anonymous Authors}
  \fi
\fi

\begin{document}

\newcommand\blfootnote[1]{%
  \begingroup
  \renewcommand\thefootnote{}\footnote{#1}%
  \addtocounter{footnote}{-1}%
  \endgroup
}

\maketitle
\ifdefined\IsFinal
  \ifdefined\VenueIsSciFM\lhead{Published as a workshop paper at Sci-FM@COLM 2026}%
  \else\lhead{Published as a conference paper at COLM 2026}\fi
\else\ifdefined\IsPreprint
  \lhead{Preprint. Under review.}
\else
  \lhead{Under review as a conference paper at COLM 2026}
\fi\fi
\FloatBarrier
\suppressfloats[t]

\ifcolmsubmission
\linenumbers
\fi

\begin{abstract}
We use training-data attribution as an interpretable tool for capability discovery, mapping which regions of the pretraining corpus support social reasoning versus STEM reasoning in OLMo3-7B.
Training-data attribution measures how strongly each training document influences a model's predictions on a benchmark, but document-level scores are too noisy to identify which corpus regions support which capabilities.
We compute gradient-based attribution (TrackStar via Bergson) over a working set drawn from the de-duplicated Dolma3 mix, aggregate influence across WebOrganizer's \WebOrgNumFormats{}-format $\times$ \WebOrgNumTopics{}-topic taxonomy (\WebOrgNumBins{}~bins), and contrast benchmark pairs in a $2{\times}2$ design that varies domain (social vs.\ STEM) and capability type (reasoning vs.\ knowledge): SocialIQA and MMLU Social Sciences against ARC-Challenge and MMLU STEM.
Social and STEM reasoning draw on qualitatively distinct corpus regions, and the contrast is sharper at the reasoning level than at the knowledge level.
Targeted machine unlearning provides partial causal validation: forgetting high-attribution topics (e.g., Literature for SocialIQA) degrades the aligned benchmark more than within-topic random baselines.
We release the code and aggregate artifacts at \url{https://github.com/HCAI-Lab-GT/capabilibara} and \url{https://huggingface.co/HCAI-Lab-GT}.

\end{abstract}

\ifdefined\VenueIsCOLM
\vspace{-10pt}
\begin{center}
    \includegraphics[width=\linewidth,height=0.24\textheight,keepaspectratio]{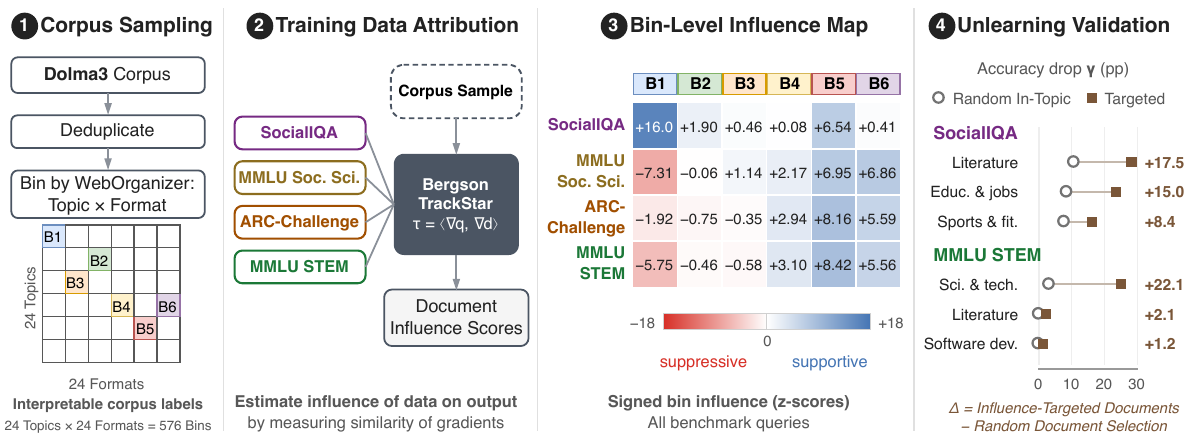}
    \captionof{figure}{Capability provenance pipeline. \textbf{1)}~Dolma3 stratified into 576 WebOrganizer bins. \textbf{2)}~Bergson/TrackStar attributes probes to bins. \textbf{3)}~Signed z-scores over all benchmark queries, as in Figure~\ref{fig:rq1-format-concentration}, for bins B1 Lit.$\times$Cust.\ Support, B2 Social$\times$Q\&A, B3 Travel$\times$Q\&A, B4 Sci.\&Tech.$\times$Academic, B5 Industrial$\times$Docs, B6 Politics$\times$Docs. \textbf{4)}~Targeted unlearning beats random controls.}
    \label{fig:overview}
\end{center}
\vspace{0.1em}

\fi

\section{Introduction}

Training-data attribution (TDA) methods estimate how much each training example influences a model's predictions---quantifying which documents in the pretraining corpus matter most for a given behavior. Recent work using influence estimates has shown that knowledge-oriented tasks tend to rely on fewer documents, while reasoning tasks draw support from broader regions of the training corpus~\citep{ruis_procedural_2024}.
However, these analyses operate at the document level: they show how influence is distributed across training documents, but not \emph{where in the corpus} distinct capabilities, such as social versus STEM reasoning, find their support.

Identifiable corpus regions can be inspected or reweighted, so a map from capabilities onto those regions is usable for auditing and for shaping behavior. We treat aggregate gradient-based attribution as a coarse mechanistic decomposition, complementary to internal-state interpretability because it analyzes inputs rather than activations. Document-level estimates support corpus-level claims only after they are aggregated in a structured taxonomy.

We study capability provenance in the open OLMo ecosystem~\citep{soldaini_dolma_2024,groeneveld_olmo_2024} using TrackStar~\citep{chang_scalable_2024} gradient-based attribution, implemented in the Bergson library~\citep{quirke2026bergson}, over a stratified working set of \AttrDocCount{} documents (\AttrTokensB{}B tokens; \AttrDocsPerBin{}/bin) from the de-duplicated Dolma3 mix of \CorpusUniqueDocs{} (${\sim}$\CorpusUniqueDocsApprox{}) unique documents.
WebOrganizer~\citep{wettig2025weborganizer} assigns each document a topic (e.g., Social Life, Literature, Software Development) and a format (e.g., Q\&A Forum, Customer Support, Documentation).
Our attribution unit is the resulting topic--format bin (\WebOrgNumTopics{} $\times$ \WebOrgNumFormats{} = \WebOrgNumBins{} bins), such as Social Life $\times$ Q\&A Forum or Literature $\times$ Customer Support (full taxonomy in Appendix~\ref{app:taxonomy}).
We choose the four primary benchmarks as a controlled OLMES-compatible $2{\times}2$: domain (social vs.\ STEM) crossed with capability type (reasoning vs.\ knowledge)~\citep{gu2025olmes}.
The reasoning cells are \SocialReasoning{SocialIQA}~\citep{sap2019socialiqacommonsensereasoningsocial} and \STEMReasoning{ARC-Challenge}~\citep{clark2018think}, which ask likelihood-scored multiple-choice questions over social commonsense narratives and multi-step scientific problems; the knowledge cells are \SocialKnowledge{MMLU Social Sciences} and \STEMKnowledge{MMLU STEM}~\citep{hendrycks2021measuringmassivemultitasklanguage}.
Because all four targets are fixed-choice, above-chance OLMo3 evaluations under the same protocol, this pairing lets us ask whether corpus regions separate by domain, by capability type, or by their interaction (Figure~\ref{fig:overview}).

The resulting profiles separate SocialIQA from the three comparison tasks at the topic and format levels. We then use targeted unlearning to test whether high-attribution regions are functionally necessary, and extend attribution to a held-out suite of social-reasoning probes spanning theory of mind, moral judgment, bias, and pragmatics.

We frame the study around three research questions, progressing from description to comparison to unlearning validation.

\begin{itemize}[leftmargin=1em, topsep=0pt,itemsep=0pt]
    \item \RQBadge{1} Where does each benchmark's attribution concentrate in the topic$\times$format taxonomy, do social and STEM benchmarks show distinct profiles, and what kinds of text characterize the high-influence bins?
    \item \RQBadge{2} Do social- and STEM-reasoning benchmarks draw attribution from different corpus regions, and does the same hold for the knowledge pair?
    \item \RQBadge{3} Are high-attribution corpus regions validated by targeted unlearning? Does forgetting them degrade the aligned benchmark more than matched random controls?
\end{itemize}

Our findings show that SocialIQA has a broader and sharply distinct training-data profile from social-domain knowledge and STEM reasoning, extending evidence that different capabilities can depend on different pretraining evidence~\citep{ruis_procedural_2024,wang2024generalization,razeghi2024backtracking}. Lexically, its high-influence bins carry a bimodal signature in which short interpersonal text and long-form documentation both contribute.
Influence-targeted unlearning validates this association most clearly for SocialIQA, while weaker, null, and reversed effects on the comparison tasks delimit where attribution identifies functionally necessary documents~\citep{zhao2024deciphering,bu2025ngdiff}.
The held-out probes share a provenance signature that SocialIQA captures only weakly, so the causally validated result samples one region of a broader capability space.

\section{Related Work}\label{sec:related}

Training-data attribution identifies which corpus parts shape model behavior. Classical influence functions estimate up-weighting effects without retraining~\citep{koh2017understanding}; later work introduced scalable gradient-based approximations such as TracIn~\citep{pruthi2020estimating} and TRAK~\citep{park2023trak}, with improved inverse-Hessian methods for large models~\citep{schioppa2022scaling,grosse2023studying}. Subsequent research traces model outputs or factual knowledge back to training data~\citep{akyurek2022towards,chang_scalable_2024} and builds retrieval or indexing systems for corpus-scale attribution~\citep{lin2024token,liu2024infini,liu2025olmotrace}. Concurrently, prior work identifies limitations of example-level estimates: brittleness to approximation choices~\citep{basu2020influence,grosse2023studying}, failure to capture interactions between training examples~\citep{guu2023simfluence}, and inconsistent cross-task behavior~\citep{jiao2025date}.

Prior corpus-scale capability-provenance work either characterizes top-$k$ influential documents \emph{post hoc} via language models or human annotation~\citep{ruis_procedural_2024, razeghi2024backtracking}, or unlearns putative dataset components on closed-data models via proxy datasets~\citep{zhao2024deciphering}. We instead apply structured taxonomic decomposition (WebOrganizer;~\citealp{wettig2025weborganizer}) \emph{before} inspection, on the released pretraining corpus; to our knowledge this is the first application of gradient-based corpus-scale TDA to \emph{capability provenance} with taxonomic aggregation on a fully open data ecosystem.

Our approach relies on an open-model ecosystem whose pretraining corpus, training pipeline, and checkpoints are all released at the document level, such as Dolma~\citep{soldaini_dolma_2024} and OLMo3~\citep{olmo3}. \emph{Open-weight} and \emph{open-data} are distinct: models that publish weights without their pretraining corpora~\citep{longpre_largescale_audit_2024} cannot support document-level attribution. Among fully open families (OLMo, Pythia~\citep{biderman_pythia_suite_2023}, the Common Pile-trained models~\citep{kandpal_common_pile_2025}), OLMo3 currently has the most mature evaluation methodology (OLMES;~\citealp{gu2025olmes}).

Attribution scores are associational, and machine unlearning offers an intervention to complement them. Recent work explores unlearning for analyzing training data effects: NGDiff~\citep{bu2025ngdiff} forgets targeted data via gradient-normalized objectives without catastrophic loss of general ability, and~\citet{zhao2024deciphering} show that unlearning specific pretraining slices reveals which data components drive downstream performance. We adopt bin-level unlearning to validate our strongest attribution findings.

We ask whether different capability families—such as social reasoning, social-domain knowledge, and quantitative reasoning—draw support from systematically different regions of pretraining data.
Recent tracing work shows that capabilities depend on distinct types of pretraining evidence~\citep{ruis_procedural_2024}, memorization and generalization can be analyzed via attribution~\citep{wang2024generalization}, and mathematical reasoning can be traced to specific data sources~\citep{razeghi2024backtracking}. We extend this line of work by asking whether such distinctions emerge at the aggregate corpus level. Complementary data-centric literature selects and audits pretraining corpora at scale~\citep{albalak_data_selection_survey_2024,elazar_wimbd_2024}.
Capability-level provenance maps supply the evidence such curation efforts
currently lack.

\section{Method}\label{sec:method}

We study capability provenance through data attribution, using targeted benchmarks
as queries against an open-data model, complemented by validation via
machine unlearning. Our setting is the OLMo3-7B/Dolma3 ecosystem, where we
run the pipeline end to end, from corpus profiling to causal validation.

\subsection{Corpus Construction}\label{sec:corpus}

Training data attribution requires a model family whose pretraining corpus
and checkpoints are both released. We therefore center the study on the
OLMo3-7B open-data ecosystem~\citep{olmo3}: OLMo3-7B Base is the object of
pretraining-data attribution over Dolma3, while OLMo3-7B Instruct supplies the benchmark
completions and query gradients used for scoring. This split follows the
base-document/instruct-query design of~\citet{ruis_procedural_2024}:
document gradients and corpus-side curvature are computed on Base so the
index remains tied to pretraining data, whereas query gradients are computed
on Instruct because it follows the OLMES multiple-choice and chat-style
prompts more reliably.

The 6T token mix was generated with quality-aware upsampling, in which higher-quality data is replicated more frequently in the training mixture to increase its effective weight~\citep{soldaini_dolma_2024}. We de-duplicated the dataset to avoid duplicate attributions from upsampled documents. We then classify the Common Crawl and olmOCR portions of the de-duplicated corpus (\CorpusBinnedDocs{} documents, ${\sim}87\%$ of the population) with topic and format classifiers from WebOrganizer~\citep{wettig2025weborganizer}, yielding taxonomies of \WebOrgNumTopics{} topics $\times$ \WebOrgNumFormats{} formats, resulting in \WebOrgNumBins{} bins. The remaining code, mathematics, and reference families (Stack-Edu, FineMath, arXiv, Wikipedia) are retained in the population count but not taxonomized, since the web-oriented WebOrganizer taxonomy does not apply to them. These bins are the units of analysis for every corpus-level comparison that follows.

We estimate influence on a stratified sample of the de-duplicated dataset in which every taxonomy bin is equally represented. The Dolma3 corpus~\citep{soldaini_dolma_2024} exhibits substantial skew, so naive sampling would under-sample low-volume bins and yield unstable gradient estimates (see Appendices~\ref{app:corpus-dist} and~\ref{app:sampling}). We therefore construct the working set by sampling uniformly from each bin, which puts all bins on a common footing for comparison.

\subsection{Training Data Attribution}\label{sec:attribution}

With the stratified corpus established, we define the attribution queries. Each benchmark sample is a query to the attribution pipeline. The four core benchmarks below are the primary queries; a held-out suite of social-reasoning probes reuses the same query pipeline for the generalization analysis (construction and results in Appendix~\ref{app:heldout-suite}).

We adopt a $2 \times 2$ contrastive design crossing domain (social vs.\ STEM)
with capability type (reasoning vs.\ knowledge), shown in
Table~\ref{tab:contrastive-design}: SocialIQA for social commonsense
reasoning~\citep{sap2019socialiqacommonsensereasoningsocial}, ARC-Challenge
for science reasoning~\citep{clark2018think}, and MMLU subdomains for
social-science and STEM
knowledge~\citep{hendrycks2021measuringmassivemultitasklanguage}. The design
distinguishes domain-specific effects, capability-specific effects, and
domain-general patterns shared across all benchmarks. All four benchmarks are
part of the OLMo3 base evaluation suite, evaluated with
OLMES~\citep{gu2025olmes} following the exact protocol used during model
development (Appendix~\ref{app:attribution}).

\begin{table}[t]
\centering
\footnotesize
\caption{Contrastive benchmark design: domain (social vs.\ STEM) crossed with capability type (reasoning vs.\ knowledge). Cell tint follows the manuscript's semantic benchmark palette; citations for the four benchmarks appear at first mention in the introduction.}
\label{tab:contrastive-design}
\setlength{\tabcolsep}{3pt}
\renewcommand{\arraystretch}{1.08}
\begin{tabularx}{0.98\columnwidth}{@{}>{\bfseries\raggedright\arraybackslash}p{0.17\columnwidth}>{\centering\arraybackslash}X>{\centering\arraybackslash}X@{}}
\toprule
\multicolumn{1}{c}{Domain} & \textbf{Reasoning} & \textbf{Knowledge} \\
\cmidrule(lr){1-1}\cmidrule(lr){2-2}\cmidrule(lr){3-3}
\midrule
Social &
  \SocialReasoningTableCell{} &
  \SocialKnowledgeTableCell{} \\
\midrule
STEM &
  \STEMReasoningTableCell{} &
  \STEMKnowledgeTableCell{} \\
\bottomrule
\end{tabularx}
\end{table}

\begin{SocialTDADesignBox}{Unit of analysis}
A WebOrganizer \AttributionConcept{bin} is one topic$\times$format cell.  For
each benchmark, the attribution method assigns that bin a signed, within-benchmark
$z$-score: \PositiveAttribution{positive} values indicate supportive influence
on the benchmark query gradients, while \NegativeAttribution{negative} values
indicate suppressive or contrasting influence.
\end{SocialTDADesignBox}

With the benchmarks established, we compute document-level attribution scores
using TrackStar~\citep{chang_scalable_2024} as implemented in Bergson~\citep{quirke2026bergson}.
Gradient-based training-data attribution estimates how strongly each training
document affects a model's behavior on a downstream query by comparing the gradient it would induce during training with the gradient the model produces on the query. A document whose gradients consistently align
with a benchmark's query gradients is judged to have positive influence on
that benchmark, and one whose gradients consistently oppose them is judged
to have negative influence. TrackStar is a scalable variant that projects
these gradients into a low-rank subspace so the
comparison runs at corpus scale. We build the corpus gradient index from
OLMo3-7B Base and reduce benchmark query gradients from OLMo3-7B Instruct. The influence
score compares pretraining-document gradients against the
instruction-tuned model's task gradients, under the approximation used by
\citet{ruis_procedural_2024} that the supervised-instruction-tuning
second-order term is the identity. The output is a signed influence score per
document--benchmark pair. Implementation details are in
Appendix~\ref{app:attribution}.

\subsection{Influence Aggregation}\label{sec:aggregation}

We aggregate attribution over taxonomy bins because individual document
scores can be brittle under influence-function approximations and sensitive
to example interactions~\citep{basu2020influence,grosse2023studying,guu2023simfluence,jiao2025date}.
The WebOrganizer bins provide fixed, semantically named aggregation units~\citep{wettig2025weborganizer},
and the resulting patterns are stable enough to interpret at corpus-region level.
We aggregate with the mean rather than the median because the within-bin
influence distribution is long-tailed and the tail carries the signal that
drives the targeted unlearning forget sets; the paired unlearning statistics
reported later use medians, where the goal is instead to resist outlier runs.

Formally, for each benchmark $\beta$ and bin $k$, we compute the mean influence $\bar{s}$:

\[
\bar{s}_{\beta,k} = \frac{1}{|D_k|} \sum_{j \in D_k} \frac{1}{|Q_\beta|} \sum_{q \in Q_\beta} s(j, q),
\]

where $s(j, q)$ is the attribution score between training document $j$ and benchmark
query $q$, $Q_\beta$ is the set of queries for benchmark $\beta$, and $D_k$ is the set
of documents in bin $k$.

This produces a $\WebOrgNumBins{} \times 4$ matrix of bin-level influence values, used in all subsequent analyses.

For each pair of benchmarks $(\beta, \beta')$, we compute the per-bin difference $\Delta z_k = z_\beta(k) - z_{\beta'}(k)$, where $z_\beta(k)$ is the standardized bin-level influence for benchmark $\beta$ at bin $k$ (computed via the aggregation rule above), identifying bins that are discriminative between benchmarks versus shared. We report bins where $|\Delta z|$ exceeds the bootstrap-derived noise floor as discriminative; confidence intervals are derived from 1{,}000-iteration resampling over documents within bin and shown in the corresponding figures.

For each benchmark $\beta$, we partition the eval queries $Q_\beta$ into a \emph{correct cohort} $Q_\beta^+$ (queries the model answers correctly under standard OLMES scoring) and an \emph{incorrect cohort} $Q_\beta^-$. We then compute per-bin influence aggregates separately for each cohort using the same bin-level aggregation rule, yielding $z_\beta^+(k)$ and $z_\beta^-(k)$. The correctness differential $\Delta z^{\mathrm{c}}_\beta(k) = z_\beta^+(k) - z_\beta^-(k)$ identifies bins whose content disproportionately drives the model's \emph{correct} predictions on $\beta$, separating provenance-of-capability (positive $\Delta z^{\mathrm{c}}$) from provenance-of-error (negative $\Delta z^{\mathrm{c}}$), with bootstrap 95\% confidence intervals as above.

To characterize the text in high-influence bins, we compute an open, LIWC-adjacent profile over the working set: in-repo pronoun, mental-state, function-word, and cognition counters, social/affective/topical categories from the open-source Empath analyzer~\citep{fast_empath_understanding_2016}, and direction-preserving proxies for the four LIWC summary variables (Analytic, Clout, Authentic, Tone). These are not LIWC.app outputs; the crosswalk, lexicons, and $z$-scoring baselines, with the expanded characterization, are detailed in Appendix~\ref{app:bin-characterization}.

\subsection{Validation via Targeted Machine Unlearning}\label{sec:unlearning}

While attribution identifies associational patterns, it does not
show that the identified data is necessary for a given capability. We
complement our influence analysis with targeted machine unlearning, treating
topic-level forgetting as a diagnostic intervention, following prior uses of
unlearning to analyze pretraining-data effects~\citep{zhao2024deciphering}.
Because the causal claim concerns pretraining data,
unlearning is applied to OLMo3-7B Base rather than the instruction-tuned
checkpoint; this avoids treating post-training behavior as a direct
forget-set target. Thus, the capability profiles reported for attribution
use the deployment-realistic Instruct checkpoint, whereas the unlearning
degradation $\gamma$ is a base-model measurement. Following the NGDiff
framework~\citep{bu2025ngdiff}, we extend unlearning to the granular
WebOrganizer taxonomy.

We structure these interventions around two axes: selection strategy (random
vs.\ influence-guided) and data scope (single-topic vs.\ global). All
unlearning runs train a rank-8 LoRA adapter on OLMo3-7B Base with the NGDiff
gradient-normalized objective~\citep{bu2025ngdiff}, then merge the adapter back into the base
checkpoint for evaluation. We report raw accuracy damage, where positive values
mean the unlearned checkpoint is less accurate than the Base checkpoint:

\[
\gamma = A_{\mathrm{baseline}} - A_{\mathrm{unlearned}}
\]

Detailed optimization parameters, resampling strategies, and the randomized-text stopping criterion are provided in Appendix~\ref{app:unlearning}.

\subsubsection{Experimental Conditions}\label{sec:unlearn-conditions}

The exploratory sweep runs, for each of the \WebOrgNumTopics{} topics, a \textbf{random baseline} ($k{=}2{,}000$ documents sampled uniformly in-topic) against \textbf{influence-guided} top-$k$ selection for the target benchmark, plus a \textbf{global random control} with no topic alignment; this single-topic sweep and its net-effect correction are analyzed in Appendix~\ref{app:unlearning}.

The paper's primary causal contrast is a paired, seed-replicated refinement of the same two axes. For each topic we forget the top-200 documents by per-document influence for the target benchmark, paired against 1{,}000 random documents from the same topic, with a flat 9{,}000-document retain set and three seeds per condition. We test the influence-versus-random difference with paired statistics at the topic level, which avoids inflating statistical power from repeated draws within the same topic (\S\ref{sec:results-unlearning}; full protocol in Appendix~\ref{app:unlearning}).

\section{Results}\label{sec:results}

Unless otherwise noted, all influence scores reported in this section are $z$-score standardized within each benchmark over WebOrganizer bins. For each benchmark, we standardize the per-bin signed mean influence scores to have a mean of 0 and a standard deviation of 1 across bins; ``absolute influence'' refers to the absolute value of these standardized scores.

The working set contains \AttrDocCount{} documents (${\sim}$\AttrTokensB{}B tokens) stratified across \WebOrgNumBins{} WebOrganizer~\citep{wettig2025weborganizer} bins (\AttrDocsPerBin{} per bin; see Appendix~\ref{app:bin-fill} for bin fill rates).

\subsection{Attribution Profiles Across Topics and Formats}\label{sec:results-profiles}

\begin{figure}[tb]
  \centering
  \includegraphics[width=\RQOneFigureWidth]{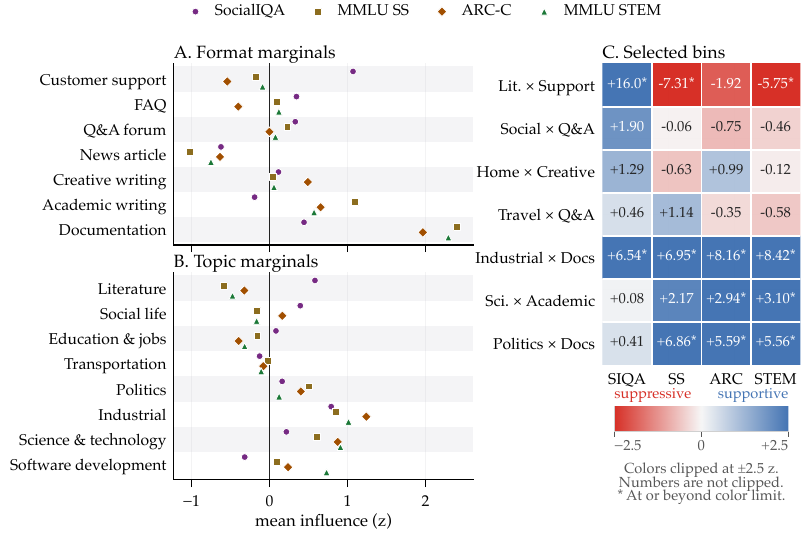}
  \caption{\SocialReasoning{SocialIQA} is the provenance-structure outlier, with
  a bin-level profile that correlates with each comparison benchmark at only
  $r\!\le\!0.21$ versus $r=0.76$--$0.86$ among the three
  (Appendix~\ref{app:profile-divergence}).
  All panels use all benchmark queries. \textbf{A--B} show mean signed influence ($z$)
  by format and topic on a shared axis, rows sorted by the
  SocialIQA-minus-comparison contrast;
  \emph{Documentation} drives the comparison benchmarks, whereas
  \emph{Customer Support} and \emph{Literature} are positive only for
  SocialIQA, which also leads \emph{Social Life} by a wide margin. \textbf{C}
  gives seven topic--format cells ordered by the same contrast, spanning the
  range from SocialIQA-exclusive to comparison-exclusive support.
  \emph{Literature~$\times$~Customer Support} is extreme for SocialIQA
  ($+16.0$) yet negative for all three comparison benchmarks ($-7.31$,
  $-1.92$, $-5.75$), whereas \emph{Politics~$\times$~Documentation} inverts
  that pattern; colors are capped at $|z|=2.5$, and asterisks mark cells at or
  beyond the cap. Full heatmaps in
  Figure~\ref{fig:influence-2x2}.}
  \label{fig:rq1-format-concentration}
\end{figure}

Figure~\ref{fig:rq1-format-concentration} summarizes the qualitative pattern that distinguishes social-reasoning attribution from the other three benchmarks. The social-knowledge and STEM benchmarks concentrate their strongest \PositiveAttribution{positive influence} in documentation-like formats, while \SocialReasoning{SocialIQA}'s strongest format marginal is \emph{Customer Support}, an interpersonal, dialogue-heavy format. At the topic level, \emph{Literature} is \PositiveAttribution{positive} for SocialIQA but \NegativeAttribution{negative} for all three comparison benchmarks, and SocialIQA leads every comparison benchmark on \emph{Social Life}: SocialIQA's support comes from informal, interpersonal text, and the comparison benchmarks' support comes from formal, expository text. Individual topic$\times$format cells hold too few documents to carry claims on their own, so we read cell-level views (Panel~C) as illustrations; the full signed heatmaps are in Figure~\ref{fig:influence-2x2} (Appendix~\ref{app:format-marginals}).

SocialIQA's positive influence is distributed far more evenly across topics and formats than the other three benchmarks, whose strong positive mass concentrates in the Documentation column while News Article is a consistent negative (Appendix~\ref{app:format-marginals}). The single strongest SocialIQA cell, Literature~$\times$~Customer Support, sits where the topic and format marginals predict it: an interpersonal format inside a narrative topic.

The per-topic and per-format marginals, reported in Appendix~\ref{app:format-marginals}, show that the observed patterns are not isolated to a single extreme bin. The near-neutral \emph{Creative Writing} and \emph{Transportation} rows are useful negative checks: the social-reasoning signal is not simply any narrative-looking format or broad human-centered topic. A whole-profile comparison confirms the effect is not driven by one bin, with SocialIQA's 576-bin influence profile correlating with the three comparison benchmarks at $r\!\le\!0.21$ versus $r=0.76$--$0.86$ among them, making it the only benchmark with a significant profile-divergence statistic (Appendix~\ref{app:profile-divergence}).

At the topic level, the clearest social-reasoning divergence is the Literature sign flip: it has positive mean influence for SocialIQA while remaining among the most negatively influential topics for the other three benchmarks, and it is the strongest SocialIQA--ARC-Challenge topic contrast in Appendix~\ref{app:paired-topics}. This places more of SocialIQA's positive attribution in Literature-labeled and narrative/interpersonal corpus regions. In contrast, ARC-Challenge, MMLU Social Sciences, and MMLU STEM display more similar topic profiles, with the Industrial and Science \& Technology topics contributing strongly positive influence across those benchmarks; Science \& Technology contributes substantially less for SocialIQA.

At the format level, marginal influence concentrates in fewer categories at larger magnitudes than at the topic level, while preserving the same contrast between SocialIQA and the other benchmarks. Documentation is the highest-influence format for ARC-Challenge, MMLU Social Sciences, and MMLU STEM, with the next-strongest formats (Academic Writing for ARC-Challenge and MMLU Social Sciences; Legal Notices for MMLU STEM) at clearly smaller magnitudes. SocialIQA again has the most distinct profile: Customer Support leads its format marginals by a wide margin, with Documentation, FAQ, and Q\&A Forum close behind (Appendix~\ref{app:format-marginals}).

\begin{SocialTDATakeaway}{RQ1 attribution profile takeaway}
\SocialReasoning{SocialIQA}'s positive influence is diffuse and
narrative/interpersonal; \SocialKnowledge{MMLU Social Sciences},
\STEMReasoning{ARC-Challenge}, and \STEMKnowledge{MMLU STEM} concentrate in
documentation-like and technical regions.
\end{SocialTDATakeaway}

\FloatBarrier

\subsection{Cross-Domain Contrastive Analysis}\label{sec:results-contrastive}

The paired-topic heatmaps in Appendix~\ref{app:paired-topics} show a sharper social--STEM separation for reasoning (SocialIQA--ARC-Challenge) than for knowledge (MMLU Social Sciences--MMLU STEM), while preserving the same direction: Software Development tilts STEM and Education \& Jobs tilts social. The asymmetry is quantitative as well as visual: the reasoning pair's top-3 topic contrasts span $|\Delta z| \le 0.91$ versus $\le 0.63$ for the knowledge pair, roughly $1.4\times$ the dynamic range, so in this controlled $2{\times}2$ design capability type differentiates corpus provenance more than domain does.

\begin{SocialTDATakeaway}{RQ2 contrast takeaway}
The social--STEM split is sharper for reasoning than knowledge:
\SocialReasoning{SocialIQA}$-$\STEMReasoning{ARC-Challenge} separates corpus
regions more strongly than \SocialKnowledge{MMLU Social Sciences}$-$\STEMKnowledge{MMLU STEM},
though the direction is preserved.
\end{SocialTDATakeaway}

Three additional analyses, detailed in the appendix, refine this picture. A correctness partition reveals a reasoning-vs-knowledge divergence: the text most associated with correct answers on both reasoning benchmarks (\SocialReasoning{SocialIQA}, \STEMReasoning{ARC-Challenge}) is dialogue-rich, interpersonal writing, and the same text is associated with errors on both knowledge benchmarks (\SocialKnowledge{MMLU Social Sciences}, \STEMKnowledge{MMLU STEM}); the sharpest single example is the \emph{Literature~$\times$~Customer Support} cell (Appendix~\ref{app:correctness}). A wider held-out suite of social, Theory-of-Mind, moral, and pragmatic probes runs through the same pipeline, analyzed in Appendix~\ref{app:heldout-suite}. Lexical profiling shows \SocialReasoning{SocialIQA}'s high-influence bins are bimodal: dialogue-rich interpersonal text alongside long-form documentation (Appendix~\ref{app:bin-characterization}).

To test how far the SocialIQA lens generalizes, we extend attribution to the
OLMES-gated held-out suite of Appendix~\ref{app:heldout-suite}: nine probes spanning
theory of mind, moral and normative judgment, social bias, and pragmatics,
scored against the same gradient index and aggregated over the same
\WebOrgNumBins{}-bin taxonomy. Seven of the probes share a provenance block
(mean within-block $r = 0.70$) whose supportive bins carry a documentary,
formal-register signature, while SocialIQA sits outside it ($r = 0.22$ to the
block, $0.20$ across all nine held-out probes;
Figure~\ref{fig:probe-similarity-full},
Appendix~\ref{app:heldout-probe-geometry}). A held-out unlearning test contributes one causal data point beyond the four-benchmark
design (Appendix~\ref{subsec:heldout-tom-unlearning}). We read this as scoping the SocialIQA result: social reasoning is not monolithic, and the causally validated finding anchors one region of a broader space of socially relevant capabilities, each with its own corpus support.

\subsection{Empirical Validation via Machine Unlearning}\label{sec:results-unlearning}

We validate the attribution results with machine unlearning, comparing influence-targeted document removal against in-topic random removal across the four benchmarks.

In the faithful replication, which unlearns the top-200 documents by per-document influence within each of the \WebOrgNumTopics{} topics, influence-targeted unlearning produces larger accuracy drops than random in-topic unlearning most clearly for \textbf{SocialIQA}: across the available matched topic--seed cells, the paired difference $\gamma_{\mathrm{influence}} - \gamma_{\mathrm{random}}$ is significantly positive (Wilcoxon BH-adjusted $p \approx 10^{-5}$; full per-benchmark statistics in Appendix~\ref{app:unlearning-results}, Table~\ref{tab:paired-significance} and Figure~\ref{fig:unlearning-paired}). \textbf{MMLU STEM} is weaker: it reaches significance on the pooled Wilcoxon test, but not on the paired $t$-test over topic means, and its bootstrap median interval crosses zero. \textbf{MMLU Social Sciences} does not reach significance, while \textbf{ARC-Challenge} has a negative median paired difference, meaning influence-targeted forgetting is slightly less damaging than random in-topic forgetting under this construction. High-attribution documents thus identify capability-relevant corpus regions, not merely broad topic membership, with SocialIQA as the clearest case.

Appendix~\ref{app:unlearning-results} reports the full unlearning suite, including convergence-speed and influence--drop diagnostics, which we treat as supporting evidence rather than separate main claims.

\begin{SocialTDATakeaway}{RQ3 validation takeaway}
\TargetedValidation{Influence-targeted unlearning} reduces
\SocialReasoning{SocialIQA} accuracy more than \ControlValidation{same-topic
random controls}. On the comparison tasks the paired effect weakens,
disappears, or reverses, so attribution does not uniformly isolate
functionally necessary documents across benchmarks.
\end{SocialTDATakeaway}

\section{Discussion}\label{sec:discussion}

What generalizes across our settings is a relationship between influence targeting and selective, benchmark-specific damage.
Within OLMo3-7B, the knowledge-oriented benchmarks draw concentrated support
from documentation-led corpus regions and correlate with one another, while
the social probes decompose by probe (Appendix~\ref{app:heldout-suite}).
\emph{Which} corpus support is functionally necessary, down to which benchmark
shows it, is probe-sensitive; the causal tests support the selectivity itself. We accordingly
read attribution maps per probe and benchmark, and reserve broader claims for
the contrasts that replicate.

\section{Conclusions and Future Work}

Within our OLMo3-7B/Dolma3 setting, SocialIQA draws positive attribution from a broader, more evenly distributed set of topic$\times$format regions than the comparison tasks, with strong influence from both interactional and expository writing. Influence-targeted unlearning validates this result most clearly for SocialIQA.

A fixed taxonomy turns noisy document-level scores into corpus units a reader can compare or perturb. Influence estimates stay approximate; the taxonomy is what makes corpus-level interpretation more stable than a standalone document ranking.

Direct extensions include repeating the pipeline on a larger model from the same family with an identical unlearning protocol, testing whether causal selectivity and category associations survive scale, and applying it to other fully open pretraining ecosystems to test how capability attribution distributions vary with corpus composition.

\section*{Limitations}

Training-data attribution estimates influence; it does not prove causal
necessity for individual documents, and large web corpora---heterogeneous,
source-correlated, and shaped by filtering and upsampling
decisions~\citep{soldaini_dolma_2024}---strain the assumptions behind
influence approximations~\citep{basu2020influence,grosse2023studying}.
Bin-level aggregation mitigates the resulting instability without eliminating
it, and the bins are classifier-defined compressions rather than exact
semantic partitions~\citep{wettig2025weborganizer}; we read them as structured
summaries and test the strongest findings with targeted unlearning
(\S\ref{sec:unlearning}).

Our provenance claims are anchored to the de-duplicated Dolma3
population~\citep{soldaini_dolma_2024}, but the empirical analysis operates on
the stratified working set drawn from its WebOrganizer-binned portion
(\S\ref{sec:method}). Results characterize benchmark-relevant structure within
that working set; they should not be read as exhaustively characterizing every
document in the full corpus.
Corpus-frequency weighting changes the question from per-bin capability signal
to population contribution. A post-hoc token-mass weighting check preserves the
sign of each bin's influence but changes which bins dominate population-level
totals, because high-volume bins can outweigh lower-volume high-$z$ bins. For
example, SocialIQA's strongest balanced bin remains Literature~$\times$~Customer
Support, while the largest token-weighted positive contributions come from
higher-mass technical/tutorial bins. We therefore report the main results as
balanced taxonomy-level contrasts and treat corpus-weighted summaries as a
separate population-scope diagnostic.

The core causal validation rests on the four primary benchmarks. The held-out
probe suite broadens the picture, but it is attribution-first evidence: apart
from the single held-out Theory-of-Mind unlearning test
(Appendix~\ref{subsec:heldout-tom-unlearning}), the probes carry no matching
unlearning validation, and we interpret them only under the tiering policy of
Appendix~\ref{app:heldout-suite}. Other social-reasoning benchmarks (\eg ToMi and
BigToM~\citep{gandhi_understanding_social_2023}) may reveal different
attribution patterns.

The unlearning experiments test whether removing high-attribution topics harms
benchmark performance, but they do not identify the mechanisms by which
training documents shape model behavior: topic-level forgetting establishes that
a corpus region is functionally necessary without decomposing that dependence
into features, patterns, or individual examples.

Our deepest corpus-profile analyses use a single open-data ecosystem
(OLMo3-7B/Dolma3), for the reason given in \S\ref{sec:related}: document-level
TDA requires a released pretraining corpus, and OLMo3 is the best-instrumented
family that qualifies. Whether the high-level attribution patterns hold across
other open-data models trained on web-scale corpora remains open; at
\ComputeTotalEquiv{} H200-equivalent GPU-hours for the present analysis,
testing this is a separate follow-up study rather than an incremental
experiment.

\ifdefined\VenueIsCOLM
  \ifdefined\IsNamed
     \section*{Author Contributions}

\noindent Glenn Matlin (corresponding author) led the project as first author;
Chandreyi Chakraborty was its second-largest contributor. Saehee Eom, Mika
Okamoto, and Rayan Castilla contributed equally as joint third authors. The
matrix below lists each author's roles in the CRediT taxonomy.

{%
\def\CY{\tikz[baseline=-0.6ex]\fill[SocialTDAPurple] (0,0) circle (2.3pt);}%
\def\CN{\tikz[baseline=-0.6ex]\fill[SocialTDABorderGray] (0,0) circle (0.8pt);}%
\def\CR#1{\multicolumn{1}{c}{\rotatebox{90}{\scriptsize #1}}}%
\begin{center}
\footnotesize
\setlength{\tabcolsep}{0pt}
\renewcommand{\arraystretch}{1.12}
\begin{tabular}{@{}l@{\hspace{1.2em}}*{14}{>{\centering\arraybackslash}p{2.05em}}@{}}
 & \CR{Conceptualization} & \CR{Data curation} & \CR{Formal analysis}
 & \CR{Funding acquisition} & \CR{Investigation} & \CR{Methodology}
 & \CR{Project administration} & \CR{Resources} & \CR{Software}
 & \CR{Supervision} & \CR{Validation} & \CR{Visualization}
 & \CR{Writing -- original draft} & \CR{Writing -- review \& editing} \\
\addlinespace[3pt]
Glenn Matlin          & \CY & \CY & \CY & \CY & \CY & \CY & \CY & \CY & \CY & \CY & \CY & \CY & \CY & \CY \\
Chandreyi Chakraborty & \CY & \CN & \CY & \CN & \CY & \CY & \CY & \CN & \CN & \CY & \CN & \CY & \CY & \CY \\
\addlinespace[4pt]
Saehee Eom            & \CN & \CY & \CN & \CN & \CY & \CN & \CN & \CN & \CY & \CN & \CY & \CN & \CN & \CY \\
Mika Okamoto          & \CN & \CN & \CN & \CN & \CY & \CY & \CN & \CN & \CN & \CN & \CN & \CY & \CY & \CY \\
Rayan Castilla        & \CN & \CN & \CN & \CN & \CY & \CN & \CN & \CN & \CN & \CN & \CN & \CY & \CN & \CY \\
\addlinespace[4pt]
Louis Jaburi          & \CN & \CN & \CN & \CN & \CN & \CY & \CN & \CN & \CN & \CN & \CN & \CN & \CN & \CY \\
Alvin Deng            & \CN & \CY & \CN & \CN & \CY & \CY & \CN & \CN & \CN & \CN & \CN & \CY & \CN & \CY \\
Taywon Min            & \CN & \CN & \CN & \CN & \CN & \CN & \CN & \CN & \CN & \CN & \CN & \CN & \CN & \CY \\
Lucia Quirke          & \CN & \CN & \CN & \CN & \CN & \CN & \CN & \CY & \CY & \CN & \CN & \CN & \CN & \CN \\
Stella Biderman       & \CN & \CN & \CN & \CN & \CN & \CN & \CN & \CN & \CN & \CY & \CN & \CN & \CN & \CN \\
Mark Riedl            & \CN & \CN & \CN & \CY & \CN & \CN & \CN & \CN & \CN & \CY & \CN & \CN & \CN & \CY \\
\end{tabular}
\end{center}
}

  \fi

  \ifdefined\IsNamed
    \ifdim\pagetotal>0pt\relax\newpage\fi
    \section*{Acknowledgments}

This research was supported in part through research cyberinfrastructure
resources and services provided by the Partnership for an Advanced Computing
Environment (PACE) at the Georgia Institute of Technology, Atlanta, Georgia,
USA. RRID:SCR\_027619. We additionally thank the PACE team for the operational
support that enabled the attribution and unlearning experiments in this work.

We are grateful to the \href{https://www.matsprogram.org/}{ML Alignment \&
Theory Scholars (MATS) Program} for funding and program support throughout this
research, and in particular to Ryan Kidd and Lauren Vaughn for their mentorship,
logistical coordination, and ongoing encouragement.

Glenn Matlin thanks the \href{https://www.existence.org/}{Berkeley Existential
Risk Initiative (BERI)} for funding support of this research, and the
\href{https://www.aisi.dev/}{Georgia Tech AI Safety Initiative} for its
support and research community.

We thank \href{https://modal.com/}{Modal Labs} and CEO Charles Frye for additional compute support and
for providing infrastructure that accelerated the unlearning validation
experiments.

  \fi

\section*{Reproducibility Statement}%
\phantomsection%
\label{sec:reproducibility}

All experiments use the OLMo3-7B/Dolma3 open-data ecosystem~\citep{olmo3}:
OLMo3-7B Base anchors corpus indexing and unlearning against the de-duplicated
\CorpusTrainingTokens{} Dolma3 training mix~\citep{soldaini_dolma_2024}
(${\sim}$\CorpusUniqueDocsApprox{} unique documents), while OLMo3-7B
Instruct supplies the benchmark completions and query gradients for
attribution scoring. The reproducibility target is the full analysis
workflow over the sampled working set (\AttrDocCount{} documents,
\AttrTokensB{}B tokens across \WebOrgNumBins{} bins), not attribution over
the entire upstream population. Attribution uses
TrackStar~\citep{chang_scalable_2024} via Bergson v\BergsonVersion{};
unlearning uses
NGDiff~\citep{bu2025ngdiff} with LoRA (rank~8). The end-to-end pipeline---taxonomy enrichment of ${\sim}$\CorpusUniqueDocsApprox{}
documents, gradient index construction, attribution scoring across six benchmarks,
and \WebOrgNumTopics{}-topic unlearning---consumed approximately \ComputeTotalEquiv{} \ComputeRefGPU-equivalent GPU-hours
(Appendix~\ref{app:compute}). All pipeline parameters, compute accounting, and
storage requirements are fully specified in
Appendices~\ref{app:attribution}, \ref{app:unlearning}, and~\ref{app:compute}.

Following dataset and model documentation norms~\citep{bender2018data,gebru2021datasheets,mitchell2019model},
we release the attribution and aggregation code, sampling scripts,
bin-level manifests, the aggregate $\WebOrgNumBins{} \times 4$ influence
matrix, available unlearning adapters, and analysis scripts at
\url{https://github.com/HCAI-Lab-GT/capabilibara} and
\url{https://huggingface.co/HCAI-Lab-GT}. This release excludes document-level
influence scores and raw training examples. Together with the publicly
available Dolma3 corpus and OLMo3 weights, these materials specify the
analysis and provide its reported aggregate results. Recomputing attribution
and unlearning results requires the original model checkpoints and substantial
compute. Detailed resource accounting is in Appendix~\ref{app:compute}.

  \section*{Ethics Statement}

This study analyzes the relationship between pretraining data and model
capabilities using publicly available resources: the open-weight OLMo3
model~\citep{olmo3} and the openly released Dolma3 corpus~\citep{soldaini_dolma_2024}.
No new data was collected, and no human subjects were involved.

Our methodology operates at the level of aggregate corpus regions (WebOrganizer
taxonomy bins) rather than individual documents; like datasheets and model
cards~\citep{bender2018data,gebru2021datasheets,mitchell2019model}, the release
discloses structure and withholds raw examples.
This reduces the risk of surfacing or amplifying specific training examples. The unlearning experiments modify
model behavior through parameter updates and do not alter the underlying
training data.

Training-data attribution tools could in principle be used
to identify memorized or sensitive content in pretraining corpora~\citep{carlini_extracting_training_2021}.
We mitigate this by releasing only aggregate bin-level statistics and influence
matrices, and by paraphrasing any qualitative
examples drawn from high-influence bins. All released artifacts operate at
corpus level and do not support individual-document retrieval.

The benchmarks used in this study (SocialIQA, MMLU Social Sciences, MMLU STEM, ARC-Challenge) are established public evaluation resources~\citep{sap2019socialiqacommonsensereasoningsocial,hendrycks2021measuringmassivemultitasklanguage,clark2018think}, as are the probes in the held-out suite (Appendix~\ref{app:heldout-suite}).
Our contrastive design probes the structure of capability provenance; it does not evaluate or make claims about the social reasoning abilities of any population or demographic group.
In particular, the attribution profiles we report for the bias-related probes (BBQ~\citep{parrish_bbq_2022}) are descriptive provenance measurements of where in the corpus those probes draw influence, not a fairness audit of the model; StereoSet~\citep{nadeem_stereoset_measuring_2021} appears only as native-likelihood bias evidence, never as a direct-answer attribution surface.
We make no claim about the model's bias levels, and none of our results certify its behavior toward any group.

\fi

\bibliographystyle{venues/colm2026/colm2026_conference}
\bibliography{references,references_extra}

\clearpage
\appendix
\section*{Appendix Contents}
\startcontents[appendix]
\printcontents[appendix]{}{1}{\setcounter{tocdepth}{2}}
\vspace{1em}

\section{WebOrganizer Taxonomy}\label{app:taxonomy}

\begin{figure}[htbp]
  \centering
  \begin{tikzpicture}[
    cell/.style={draw, minimum width=0.8cm, minimum height=0.5cm, inner sep=0pt, line width=0.3pt},
    highlighted/.style={cell, fill=gray!30, line width=0.7pt},
    rowlabel/.style={anchor=east, font=\footnotesize, inner sep=2pt},
    collabel/.style={anchor=south west, font=\footnotesize, rotate=35, inner sep=2pt},
    dots/.style={font=\footnotesize, anchor=center}
  ]
    \def\colA{0}
    \def\colB{0.8}
    \def\colC{1.6}
    \def\colD{2.4}
    \def\colE{3.2}
    \def\rowA{0}
    \def\rowB{-0.5}
    \def\rowC{-1.0}
    \def\rowD{-1.5}
    \def\rowE{-2.0}

    \node[collabel] at (\colA, 0.3) {Documentation};
    \node[collabel] at (\colB, 0.3) {News Article};
    \node[collabel] at (\colC, 0.3) {Q\&A Forum};
    \node[collabel] at (\colD, 0.3) {Tutorial};
    \node[collabel] at (\colE, 0.3) {$\cdots$};

    \node[rowlabel] at (-0.45, \rowA) {Educ.\ \& Jobs};
    \node[rowlabel] at (-0.45, \rowB) {Politics};
    \node[rowlabel] at (-0.45, \rowC) {Social Life};
    \node[rowlabel] at (-0.45, \rowD) {Health};
    \node[rowlabel] at (-0.45, \rowE) {$\vdots$};

    \foreach \r in {\rowA, \rowB, \rowD} {
      \foreach \c in {\colA, \colB, \colC, \colD} {
        \node[cell] at (\c, \r) {};
      }
      \node[dots] at (\colE, \r) {$\cdots$};
    }
    \foreach \c in {\colA, \colB, \colD} {
      \node[cell] at (\c, \rowC) {};
    }
    \node[highlighted] at (\colC, \rowC) {};
    \node[dots] at (\colE, \rowC) {$\cdots$};

    \foreach \c in {\colA, \colB, \colC, \colD} {
      \node[dots] at (\c, \rowE) {$\vdots$};
    }
    \node[dots] at (\colE, \rowE) {$\ddots$};
  \end{tikzpicture}
  \caption{Schematic of the WebOrganizer cross-product taxonomy. Each of
  \WebOrgNumTopics{} topic categories pairs with each of \WebOrgNumFormats{}
  format categories to yield \WebOrgNumBins{} bins; the shaded cell
  illustrates one example bin (Social~Life~$\times$~Q\&A~Forum). Empty cells
  indicate that this figure conveys taxonomy \emph{structure}; bin
  populations are visualized separately in the joint count panels in
  Figure~\ref{fig:heatmap-docs}.}\label{fig:taxonomy-schematic}
\end{figure}
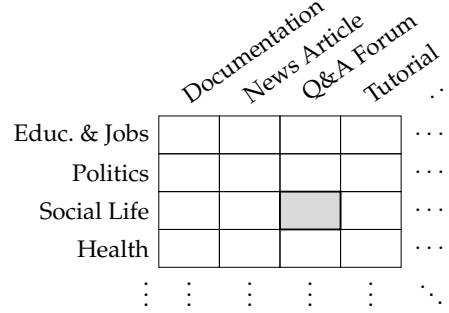

We use the WebOrganizer taxonomy~\citep{wettig2025weborganizer}, comprising
\WebOrgNumTopics{} topic categories and \WebOrgNumFormats{} format categories.
Wettig et al.\ constructed the taxonomy by reviewing existing web taxonomies
(Curlie.org, Google AdSense, Wikipedia ontology) and iteratively refining
category definitions: Llama-3.1-405B-Instruct classified CommonCrawl samples,
and the authors manually reviewed annotations to refine boundaries.
Classification of our de-duplicated Dolma3 corpus uses the official trained
WebOrganizer classifiers (gte-base-en-v1.5 backbone; see
Appendix~\ref{app:taxonomy-enrichment} for throughput, hardware, and reported
accuracy). Topic and format definitions, adapted from the original paper and
reproduced here for reference, appear in Tables~\ref{tab:topic-defs}
and~\ref{tab:format-defs}. Figure~\ref{fig:taxonomy-schematic} summarizes
the topic--format cross-product used throughout the bin-level analyses.

\begin{SocialTDACallout}{Taxonomy unit}
Throughout the paper, a WebOrganizer \AttributionConcept{bin} means one
topic$\times$format cell. Bin-level attribution aggregates noisy document
scores into these repeated units so the same cells can be compared across
\SocialReasoning{social reasoning}, \SocialKnowledge{social knowledge},
\STEMReasoning{STEM reasoning}, and \STEMKnowledge{STEM knowledge}.
\end{SocialTDACallout}

\begin{table}[htbp]
\centering
\small
\caption{WebOrganizer topic categories and descriptions (adapted from
\citealt{wettig2025weborganizer}).}\label{tab:topic-defs}
\begin{tabular}{lp{9cm}}
\toprule
\rowcolor{SocialTDALightGray}
\textbf{Topic} & \textbf{Description} \\
\midrule
Adult & Adult content \\
Art \& Design & Includes architecture \\
Crime \& Law & Includes law enforcement; financial crime falls under Finance \& Business \\
Education \& Jobs & Pedagogy, training, certification, academia \\
Entertainment & Music, movies, TV, celebrities, humor, nightlife \\
Fashion \& Beauty & Clothing, accessories, cosmetics \\
Finance \& Business & Taxes, investments, insurance, corporate communication, marketing, HR \\
Food \& Dining & Recipes, groceries, beverages, restaurants \\
Games & Video games, board games, gambling \\
Hardware & Computer hardware, phones, televisions, consumer electronics \\
Health & Medicine, wellness, mental health, veterinary science, nutritional science \\
History & Includes geography, archaeology \\
Home \& Hobbies & Real estate, furniture, home improvement, DIY, gardening, pets \\
Industrial & Mining, agriculture, manufacturing, utilities, construction \\
Literature & Literary criticism, linguistics, philosophy, humanities \\
Politics & Social issues, campaigns, legislative process, geopolitics, activism \\
Religion & Includes spirituality \\
Science \& Technology & Physics, chemistry, biology, mathematics, engineering \\
Social Life & Family, friends, relationships, community \\
Software & Use of software and the internet \\
Software Development & Algorithms, coding, web development \\
Sports \& Fitness & Martial arts, motor sports, outdoor activities \\
Transportation & Cars, taxis, public transit, aviation, shipping, logistics \\
Travel & Hospitality, hotels, sight-seeing, cruises \\
\bottomrule
\end{tabular}
\end{table}

\begin{table}[htbp]
\centering
\small
\caption{WebOrganizer format categories and descriptions (adapted from
\citealt{wettig2025weborganizer}).}\label{tab:format-defs}
\begin{tabular}{lp{9cm}}
\toprule
\rowcolor{SocialTDALightGray}
\textbf{Format} & \textbf{Description} \\
\midrule
About (Organization) & Organizational self-description or introduction \\
About (Personal) & Personal website self-description or profile \\
Academic Writing & Research papers, abstracts, theses, literature reviews \\
Audio Transcript & Written record of spoken language (interviews, lectures, podcasts) \\
Comment Section & Discussion forum with multiple posts or comments \\
Content Listing & Navigation page: sitemap, catalog, search results \\
Creative Writing & Short stories, novel chapters, poetry, song lyrics \\
Customer Support & Troubleshooting guides for a general audience \\
Documentation & Technical writing, API docs, README files, source code \\
FAQ & Frequently Asked Questions format \\
Knowledge Article & Objective, neutral style; moderated platform or reputable source \\
Legal Notices & Terms of service, disclaimers, privacy policies, licenses \\
Listicle & List-format articles (``Top 10'' lists, Buzzfeed-style) \\
News Article & Journalism on current events by news organizations \\
News (Organization) & Press releases, organizational blog posts \\
Nonfiction Writing & Long reads, profiles, editorials, essays, memoirs \\
Personal Blog & Individual writing on personal experiences and opinions \\
Product Page & Descriptions and promotions for products or services \\
Q\&A Forum & Explicit question-and-answer format (Quora, Stack Exchange) \\
Spam / Ads & SEO keyword stuffing, short ads, or no apparent purpose \\
Structured Data & Multiple data entries with common structure (tables, JSON, CSV) \\
Truncated & Incomplete content: paywalled, login-required, or multimedia descriptions \\
Tutorial & Recipes, DIY instructions, WikiHow, how-to guides \\
User Review & User-posted reviews (Yelp, TripAdvisor) \\
\bottomrule
\end{tabular}
\end{table}

\FloatBarrier

\section{Corpus Characterization}\label{app:corpus}

\subsection{De-duplication and Provenance}

OLMo3-7B~\citep{olmo3} was trained on the \CorpusTrainingTokens{} Dolma3
mix~\citep{soldaini_dolma_2024} (\texttt{allenai/dolma3\_mix-6T-1025-7B}),
distributed across \CorpusNumShards{} shards.
The mix uses quality-aware upsampling, so the same document can appear in multiple
shards (rationale and impact on attribution discussed in
\S\ref{sec:method}). We de-duplicated at the document level using a Bloom filter over
document identifiers (0.1\% false-positive rate, 10 hash functions);
Table~\ref{tab:dedup} reports raw and de-duplicated shard counts.

\begin{table}[!htbp]
\centering
\caption{De-duplication summary for the Dolma3 \CorpusTrainingTokens{} training mix.}\label{tab:dedup}
\begin{tabular}{lr}
\toprule
\rowcolor{SocialTDALightGray}
\textbf{Metric} & \textbf{Value} \\
\midrule
Total shards & \CorpusNumShards{} \\
Total document slots (with upsampling) & ${\sim}$\CorpusRawSlots{} \\
Unique documents after de-duplication & \CorpusUniqueDocs{} \\
Duplication rate & \CorpusDedupRate{}\% \\
Bloom filter size & 2.27\,GB \\
\bottomrule
\end{tabular}
\end{table}

\paragraph{Source composition.}
The de-duplicated corpus draws from six source categories
(percentages reported by \citealp{soldaini_dolma_2024}):
Common Crawl (76\%), olmOCR PDFs (14\%), Stack-Edu (7\%), FineMath (2.6\%),
arXiv (0.9\%), and Wikipedia (0.04\%).

\subsection{Corpus Distribution}\label{app:corpus-dist}

Before sampling, we analyzed all \WebOrgNumBins{}
WebOrganizer~\citep{wettig2025weborganizer} bins in the binned (Common Crawl + olmOCR)
portion of the de-duplicated corpus (\CorpusBinnedDocs{} documents). All bins are populated (zero
empty bins), but the distribution is highly skewed: the largest bin contains
over 100{,}000$\times$ more documents than the smallest, motivating our
stratified sampling design (\S\ref{app:sampling}).
Table~\ref{tab:concentration} quantifies the distributional skew: the Gini
coefficient is 0.68 for documents and 0.73 for tokens, and only ${\sim}$236
of \WebOrgNumBins{} bins carry appreciable document mass (effective bins, computed as
$\exp(H)$ where $H$ is Shannon entropy).

\begin{table}[!htbp]
\centering
\small
\setlength{\tabcolsep}{5pt}
\renewcommand{\arraystretch}{1.06}
\caption{Corpus concentration metrics for the de-duplicated Dolma3 training
mix across 576 WebOrganizer bins. The Gini coefficient and effective-bin
counts quantify the distributional skew that motivates stratified sampling.}
\label{tab:concentration}
\begin{tabular}{@{}lr@{}}
\toprule
\textbf{Metric} & \textbf{Value} \\
\midrule
Total bins (\WebOrgNumTopics{}~$\times$~\WebOrgNumFormats{}) & \WebOrgNumBins{} \\
Empty bins (0 docs) & 0 \\
Near-empty bins ($<$100 docs) & 0 \\
Near-empty bins ($<$1{,}000 docs) & 3 \\
Gini coefficient (documents) & 0.6837 \\
Gini coefficient (tokens) & 0.7337 \\
Effective bins, documents (exp H) & 236.0 \\
Effective bins, tokens (exp H) & 158.6 \\
\midrule
Top-10 bin token share & 27.8\% \\
Top-50 bin token share & 53.1\% \\
Top-100 bin token share & 71.9\% \\
\bottomrule
\end{tabular}
\end{table}

\begin{figure*}[!htbp]
\centering
\begin{subfigure}{\StackedBarPanelWidth}
  \centering
  \includegraphics[width=\linewidth]{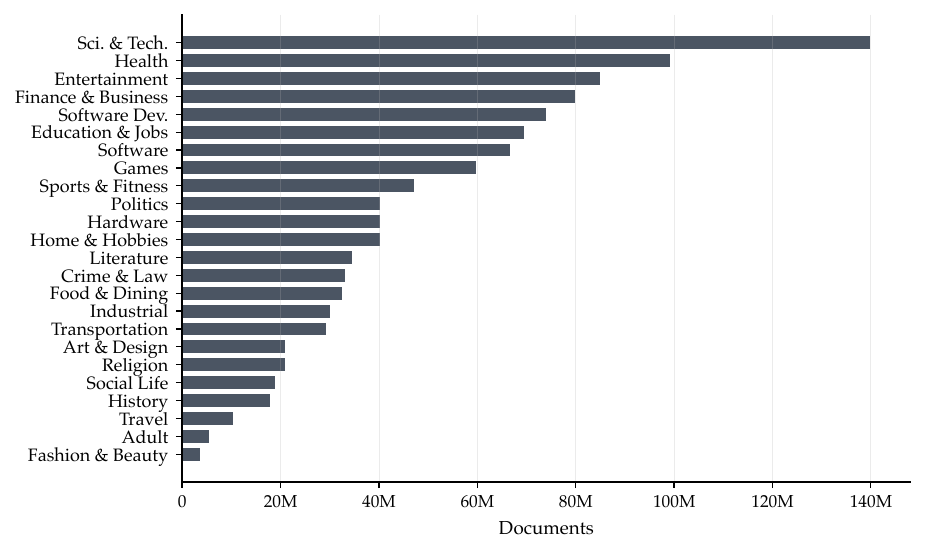}
  \caption{Document count.}
\end{subfigure}\\[4pt]
\begin{subfigure}{\StackedBarPanelWidth}
  \centering
  \includegraphics[width=\linewidth]{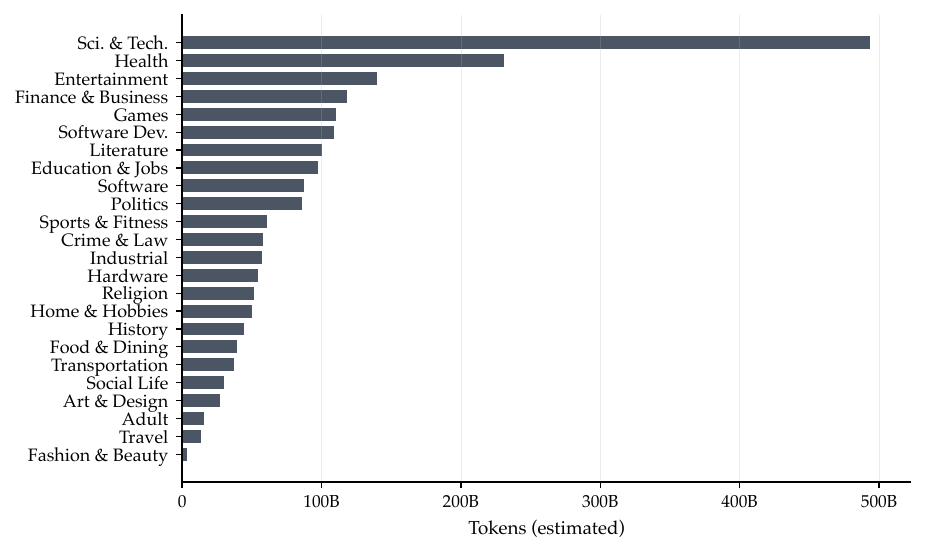}
  \caption{Token count.}
\end{subfigure}
\caption{Marginal distribution by WebOrganizer topic in the WebOrganizer-binned
portion of the de-duplicated Dolma3 corpus (\CorpusBinnedDocs{} documents).
Both panels show all 24 topics, sorted by count within each
panel. The distributions are highly skewed: \emph{Science~\&~Tech.},
\emph{Health}, \emph{Entertainment}, and \emph{Finance~\&~Business}
dominate document counts, while token counts also reflect topic-specific
document length.}\label{fig:topic-marginals}
\end{figure*}

\begin{figure*}[!htbp]
\centering
\begin{subfigure}{\StackedBarPanelWidth}
  \centering
  \includegraphics[width=\linewidth,height=0.36\textheight,keepaspectratio]{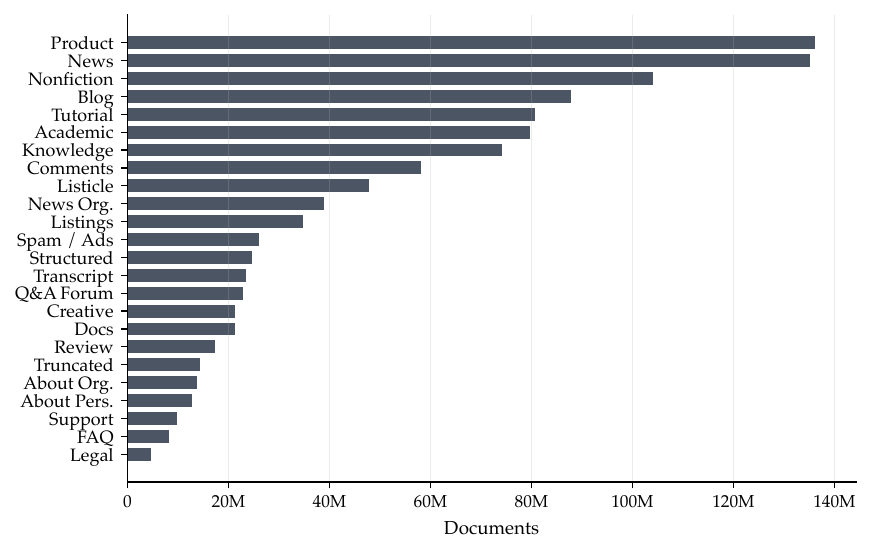}
  \caption{Document count.}
\end{subfigure}\\[4pt]
\begin{subfigure}{\StackedBarPanelWidth}
  \centering
  \includegraphics[width=\linewidth,height=0.36\textheight,keepaspectratio]{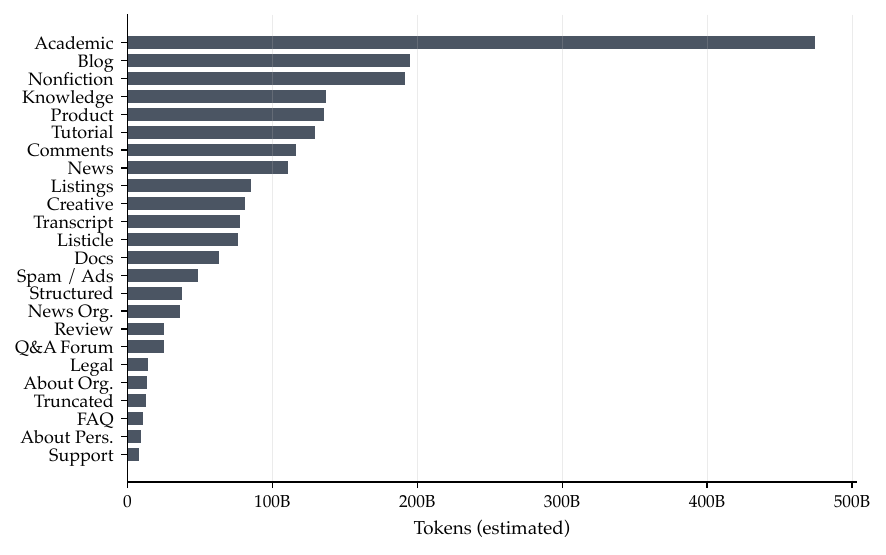}
  \caption{Token count.}
\end{subfigure}
\caption{Marginal distribution by WebOrganizer format in the WebOrganizer-binned
portion of the de-duplicated Dolma3 corpus (\CorpusBinnedDocs{} documents).
Both panels show all 24 formats, sorted by count within each
panel. \emph{Product Page}, \emph{News Article}, \emph{Nonfiction Writing}, and
\emph{Personal Blog} dominate document counts; token counts additionally reflect
format-specific length variation.}\label{fig:format-marginals}
\end{figure*}

\begin{figure}[!p]
\centering
\includegraphics[width=\StackedHeatmapPanelWidth,height=0.36\textheight,keepaspectratio]{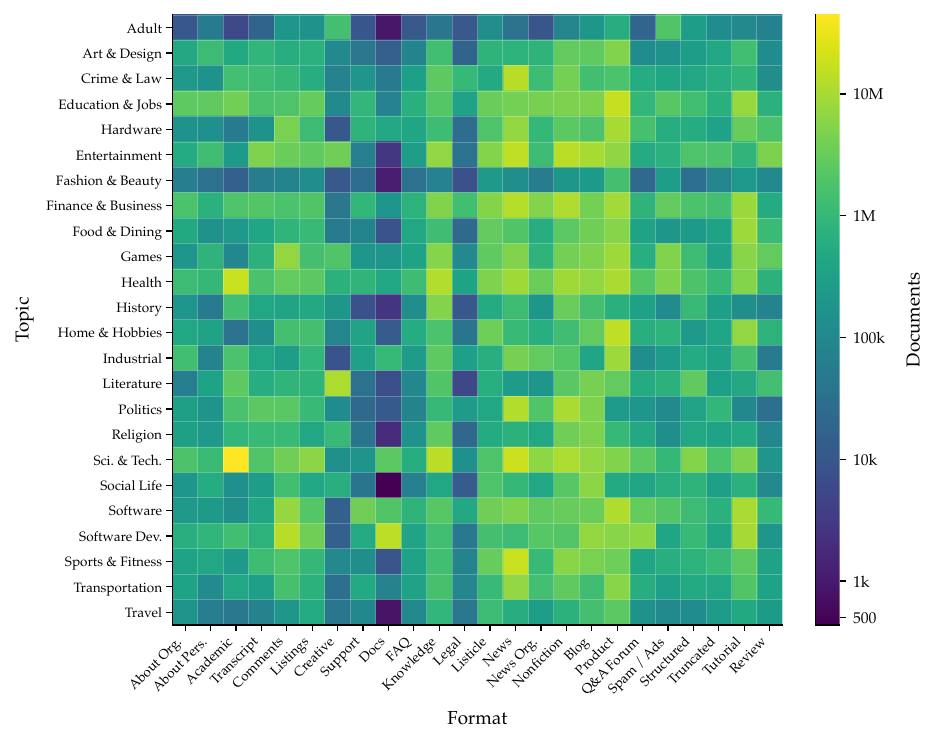}
\vspace{-0.35em}

\includegraphics[width=\StackedHeatmapPanelWidth,height=0.36\textheight,keepaspectratio]{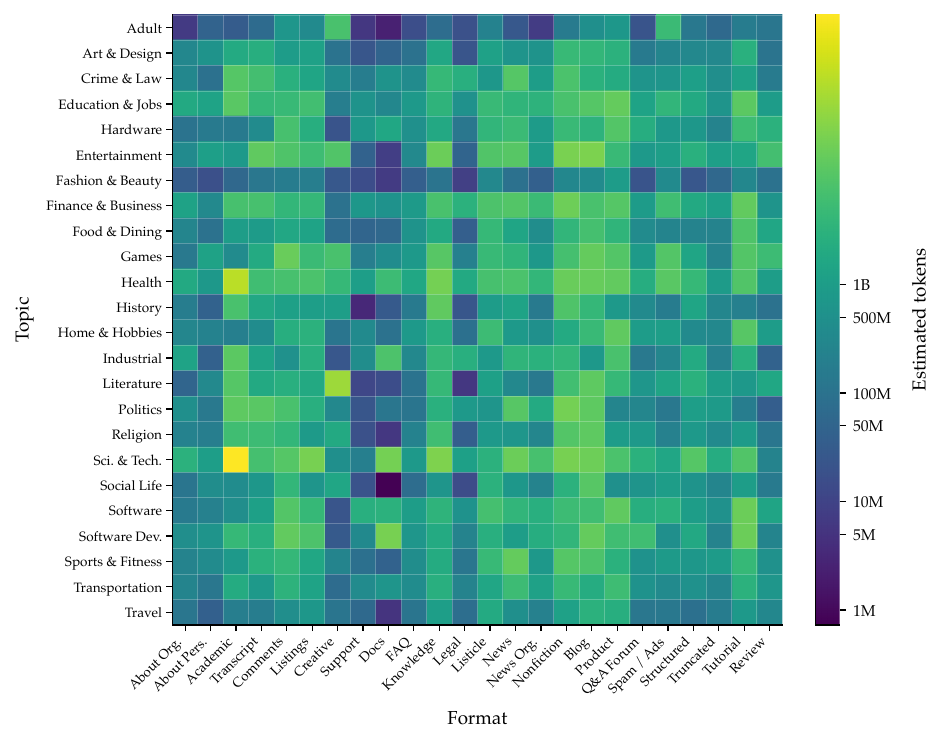}
\vspace{-0.35em}

\caption{Joint topic--format distribution across the 576 WebOrganizer bins. \textbf{(a)}~Document counts ($\log_{10}$ scale). \textbf{(b)}~Token counts ($\log_{10}$ scale). Token counts amplify the skew because longer-form formats (Academic, Docs) carry disproportionate per-document token mass.}
\label{fig:heatmap-docs}
\label{fig:heatmap-tokens}
\end{figure}

Figures~\ref{fig:topic-marginals}, \ref{fig:format-marginals},
and~\ref{fig:heatmap-tokens} visualize these properties across the
WebOrganizer taxonomy. The marginal distributions
reveal substantial concentration: a handful of topics (Science~\&~Technology,
Health, Entertainment, Finance~\&~Business) and formats (Product Page,
News Article, Nonfiction Writing, Personal Blog) dominate counts, while combinations
such as Social Life~$\times$~Documentation or Adult~$\times$~Legal Notices
contribute orders of magnitude fewer documents. The joint heatmaps expose
the full \WebOrgNumTopics$\times$\WebOrgNumFormats{} structure, showing that
skew compounds when topic and format interact. Table~\ref{tab:top-bins}
contrasts the twenty bins that carry the most token mass with the twenty
that carry the least, making the heavy-head/empty-tail asymmetry concrete:
the top 20 account for roughly 36\% of all training tokens, while the
bottom 20 contribute essentially zero.

\FloatBarrier

\begin{table*}[!htbp]
\centering
\caption{Top-20 and bottom-20 bins by token count in the de-duplicated Dolma3
corpus. The top 20 bins (of \WebOrgNumBins{}) carry ${\sim}$36\% of all
training tokens while the bottom 20 collectively carry essentially zero
(${<}0.01$\%), exposing the corpus's heavy-head/empty-tail asymmetry. The
high-mass bins concentrate in technical and reasoning-aligned topic--format
combinations; the empty-tail bins concentrate in documentation and
administrative formats (Documentation, Customer Support, Legal Notices) across
lower-volume topics. Cf.~Table~\ref{tab:concentration} for
aggregate concentration metrics.}\label{tab:top-bins}
\footnotesize

\begin{minipage}[t]{\linewidth}
  \centering
  \subcaption{Top 20 by token mass}
  \begin{tabular}{rllrrr}
\toprule
\rowcolor{SocialTDALightGray}
\textbf{Rank} & \textbf{Topic} & \textbf{Format} & \textbf{Docs} & \textbf{Tokens (B)} & \textbf{Share (\%)} \\
\midrule
1 & Science \& Technology & Academic Writing & 44.44M & 304.60 & 14.39 \\
2 & Health & Academic Writing & 16.57M & 81.51 & 3.85 \\
3 & Literature & Creative Writing & 10.71M & 46.64 & 2.20 \\
4 & Science \& Technology & Knowledge Article & 13.30M & 24.85 & 1.17 \\
5 & Entertainment & Personal Blog & 9.63M & 23.88 & 1.13 \\
6 & Entertainment & Nonfiction Writing & 13.58M & 21.99 & 1.04 \\
7 & Software Development & Documentation & 14.27M & 21.98 & 1.04 \\
8 & Science \& Technology & Nonfiction Writing & 10.85M & 21.62 & 1.02 \\
9 & Science \& Technology & Content Listing & 5.90M & 21.52 & 1.02 \\
10 & Politics & Nonfiction Writing & 10.23M & 20.52 & 0.97 \\
11 & Health & Knowledge Article & 11.82M & 20.40 & 0.96 \\
12 & Science \& Technology & Documentation & 2.47M & 19.88 & 0.94 \\
13 & Finance \& Business & Nonfiction Writing & 11.11M & 17.89 & 0.84 \\
14 & Science \& Technology & Personal Blog & 6.75M & 17.20 & 0.81 \\
15 & Entertainment & Knowledge Article & 6.94M & 16.68 & 0.79 \\
16 & Science \& Technology & News Article & 17.94M & 16.67 & 0.79 \\
17 & Software & Tutorial & 10.40M & 16.25 & 0.77 \\
18 & Software Development & Tutorial & 9.41M & 16.25 & 0.77 \\
19 & Games & Comment Section & 6.87M & 15.98 & 0.76 \\
20 & Health & Nonfiction Writing & 8.59M & 15.90 & 0.75 \\
\bottomrule
\end{tabular}
\end{minipage}

\vspace{6pt}

\begin{minipage}[t]{\linewidth}
  \centering
  \subcaption{Bottom 20 by token mass}
  \begin{tabular}{rllrrr}
\toprule
\rowcolor{SocialTDALightGray}
\textbf{Rank} & \textbf{Topic} & \textbf{Format} & \textbf{Docs} & \textbf{Tokens (B)} & \textbf{Share (\%)} \\
\midrule
557 & Religion & Customer Support & 37K & 0.02 & 0.00 \\
558 & Adult & Legal Notices & 11K & 0.02 & 0.00 \\
559 & Fashion \& Beauty & About (Personal) & 32K & 0.02 & 0.00 \\
560 & Adult & FAQ & 11K & 0.02 & 0.00 \\
561 & Literature & Documentation & 7.2K & 0.02 & 0.00 \\
562 & Fashion \& Beauty & Customer Support & 27K & 0.02 & 0.00 \\
563 & Social Life & Legal Notices & 13K & 0.01 & 0.00 \\
564 & Literature & Customer Support & 34K & 0.01 & 0.00 \\
565 & Fashion \& Beauty & Legal Notices & 7.7K & 0.01 & 0.00 \\
566 & Entertainment & Documentation & 2.9K & 0.01 & 0.00 \\
567 & Adult & News (Organization) & 9.0K & 0.01 & 0.00 \\
568 & Fashion \& Beauty & Documentation & 1.1K & 0.01 & 0.00 \\
569 & Adult & About (Organization) & 10K & 0.01 & 0.00 \\
570 & Adult & Customer Support & 9.5K & 0.01 & 0.00 \\
571 & Literature & Legal Notices & 5.4K & 0.01 & 0.00 \\
572 & Religion & Documentation & 1.9K & 0.01 & 0.00 \\
573 & Travel & Documentation & 816 & 0.01 & 0.00 \\
574 & History & Customer Support & 7.7K & 0.00 & 0.00 \\
575 & Adult & Documentation & 935 & 0.00 & 0.00 \\
576 & Social Life & Documentation & 436 & 0.00 & 0.00 \\
\bottomrule
\end{tabular}
\end{minipage}
\end{table*}

\subsection{Sampling Strategy}\label{app:sampling}

To mitigate the distributional skew documented above, we employ stratified
sampling with a fixed per-bin budget. This design is not intended to preserve
the natural Dolma3 topic--format proportions. Instead, it makes a conditional
capability-provenance estimate possible: given a corpus region, which
capabilities does that region support or interfere with? A proportional sample
at the same working-set budget would be dominated by high-mass corpus regions
and would leave many low-volume cells too sparse for stable
influence-function estimates. In the manifest-backed representative draw used
for the diagnostics below, all \WebOrgNumBins{} bins are represented, but 426 fall below the
\AttrDocsPerBin{}-document-per-bin working-set target. Populating the smallest cells proportionally
would approach sampling the whole corpus, which is computationally infeasible
for this attribution pipeline.

\begin{figure}[!p]
\centering
\includegraphics[width=\StackedHeatmapPanelWidth,height=0.36\textheight,keepaspectratio]{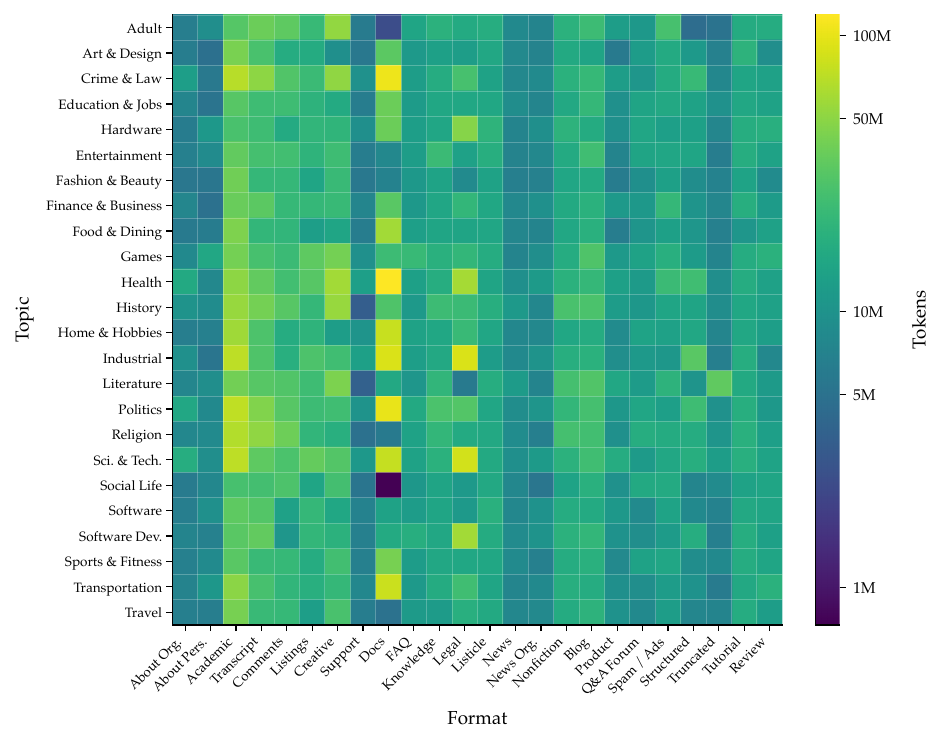}
\vspace{-0.35em}

\includegraphics[width=\StackedHeatmapPanelWidth,height=0.36\textheight,keepaspectratio]{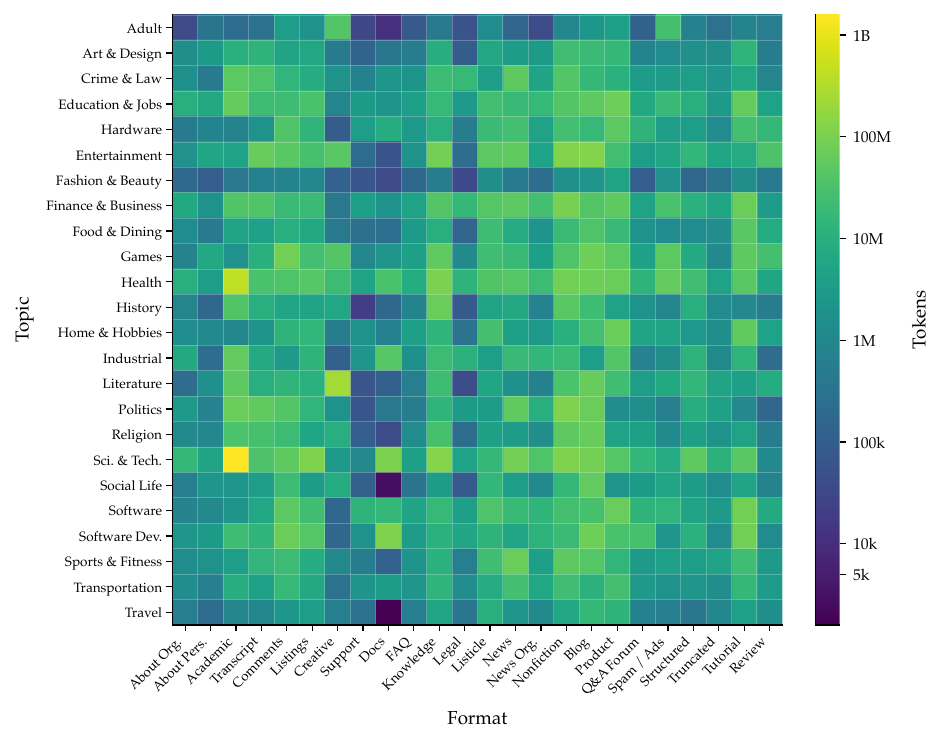}
\vspace{-0.35em}

\caption{Stratified versus representative sampling across WebOrganizer bins (token counts, $\log_{10}$ scale). \textbf{(a)}~The stratified working set gives each populated taxonomy cell comparable measurement budget. \textbf{(b)}~A manifest-backed representative draw with the same \AttrDocCount{}-document budget follows the natural corpus skew: all \WebOrgNumBins{} bins are covered, but 426 fall below the \AttrDocsPerBin{}-document-per-bin working-set target.}\label{fig:sampling-comparison}
\end{figure}

\begin{figure}[!htbp]
\centering
\includegraphics[width=\SingleHeatmapFigureWidth]{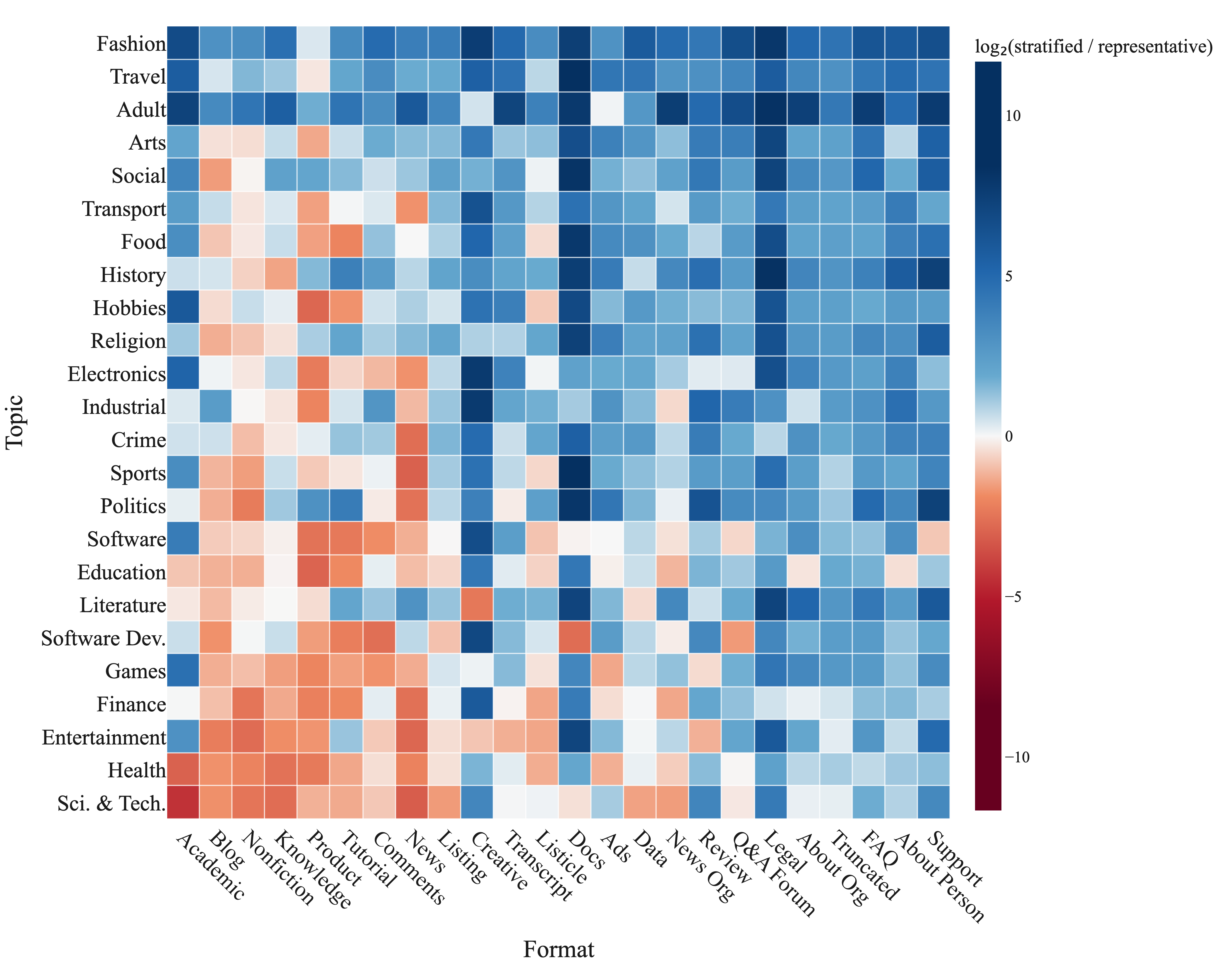}
\caption{Difference between stratified and representative sampling allocations ($\log_2$ share ratio). Positive values indicate bins up-sampled by stratification; negative values indicate bins down-sampled relative to their corpus proportion. The representative side is a manifest-backed draw with the same document budget as the stratified working set; cells with zero share in both samples render at the midpoint of the diverging scale.}\label{fig:sampling-diff}
\end{figure}

Figures~\ref{fig:sampling-comparison} and~\ref{fig:sampling-diff} visualize
this trade-off. The stratified allocation puts taxonomy cells on comparable
footing so that small but semantically meaningful regions can be measured; the
representative allocation answers a different question, namely which corpus
regions dominate total attribution mass under the natural corpus distribution.

\FloatBarrier

\subsection{Realized Sample Fill Rates}\label{app:bin-fill}

The stratified sampling design targets \AttrDocsPerBin{} documents per bin.
Table~\ref{tab:bin-pop} summarizes realized fill rates: \AttrBinsUnderfilled{}
of \WebOrgNumBins{} bins are underfilled, concentrated in the Documentation
and Legal Notices format columns where the upstream Dolma3 population simply
contains fewer than \AttrDocsPerBin{} documents for those topic combinations.
All underfilled bins still contribute to analysis; their influence estimates
are noisier due to smaller sample sizes but are not excluded.

\begin{table}[!htbp]
\centering
\caption{Realized bin fill rates for the \AttrDocsPerBin{}-document-per-bin
stratified sample across 576 WebOrganizer bins.}\label{tab:bin-pop}
\begin{tabular}{lr}
\toprule
\rowcolor{SocialTDALightGray}
\textbf{Category} & \textbf{Bins} \\
\midrule
Full (\AttrDocsPerBin{} docs realized) & 559 \\
Near-full (5{,}000--9{,}999 docs) & 10 \\
Partial (1{,}000--4{,}999 docs) & 4 \\
Sparse ($<$1{,}000 docs) & 3 \\
\midrule
\rowcolor{SocialTDASlate!4}
\textbf{Total bins} & \textbf{576} \\
Underfilled (below target) & \AttrBinsUnderfilled{} \\
Smallest realized bin & Social Life / Documentation (436 docs) \\
\bottomrule
\end{tabular}
\end{table}

\subsection{Document Length Statistics}\label{app:doc-length}

\begin{figure}[!htbp]
\centering
\includegraphics[width=\MainFigureWidth]{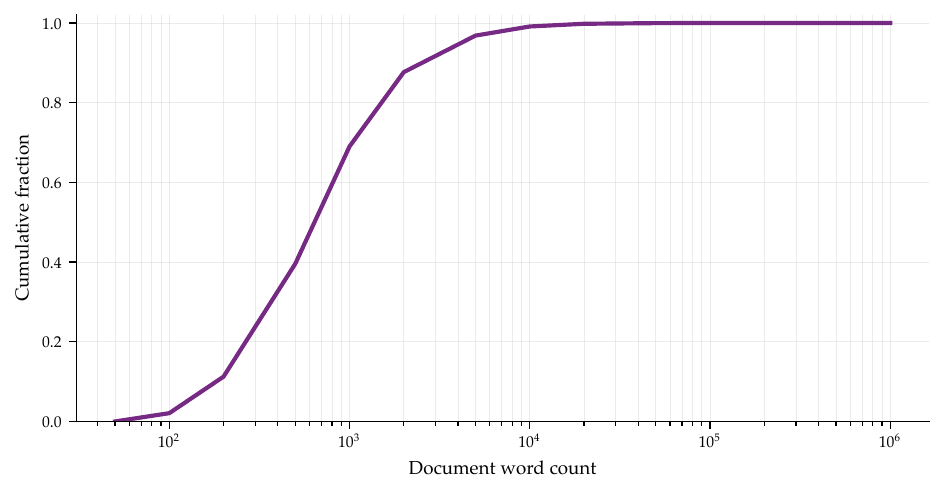}
\caption{Empirical cumulative distribution (ECDF) of document word counts over an 86.7M-document EDA sample of the de-duplicated Dolma3 corpus, plotted on a log-scaled $x$-axis. The curve's steep climb between 100 and 2{,}000 words confirms that most documents are short, while the remaining upper tail stretches to $10^5$ words, motivating per-document normalization in the attribution pipeline (\S\ref{sec:method}).}\label{fig:word-count-hist}
\end{figure}

\begin{figure}[!htbp]
\centering
\includegraphics[width=\SingleHeatmapFigureWidth]{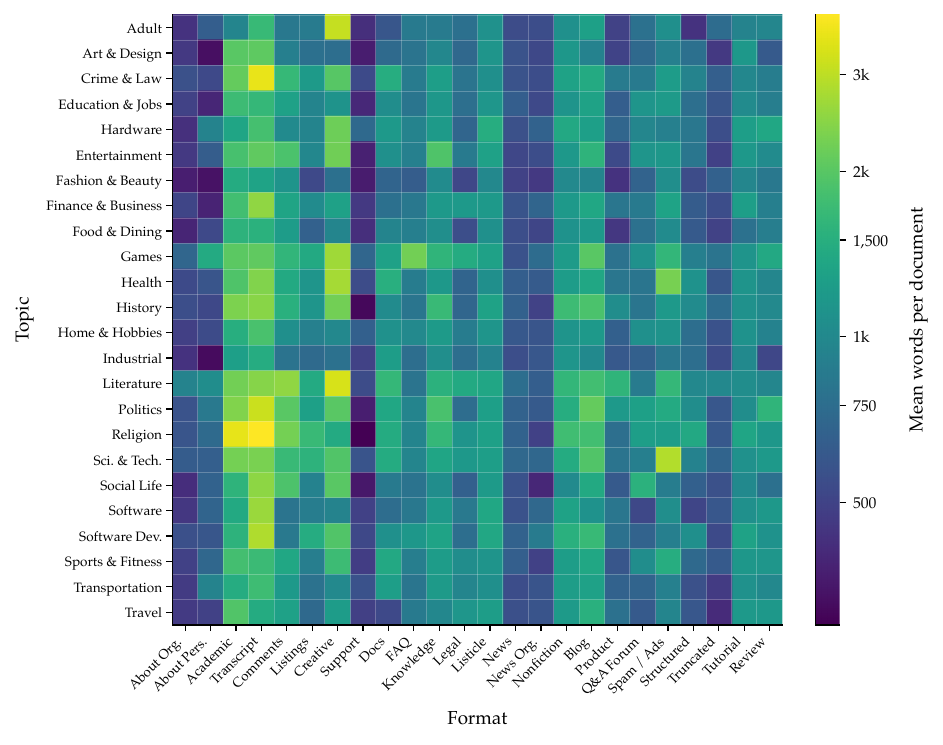}
\caption{Joint topic--format heatmap of mean document word count on a log color scale, computed over the same 86.7M-document EDA sample as Figure~\ref{fig:word-count-hist}. Colorbar ticks report mean words per document. Length variation across bins exceeds an order of magnitude, motivating per-document normalization in the attribution pipeline.}\label{fig:word-mean-heatmap}
\end{figure}

Document length varies substantially across the WebOrganizer taxonomy and may confound attribution patterns if longer documents accumulate more gradient signal. Figure~\ref{fig:word-count-hist} shows the ECDF of document word counts across the de-duplicated corpus; the joint topic--format heatmap in Figure~\ref{fig:word-mean-heatmap} reveals that mean document length varies by more than an order of magnitude across taxonomy cells, motivating per-document normalization in the attribution pipeline.

\subsection{Influence Score Distributions}\label{app:influence-diag}

\begin{figure}[!htbp]
\centering
\includegraphics[width=\MainFigureWidth,height=\TallFigureMaxHeight,keepaspectratio]{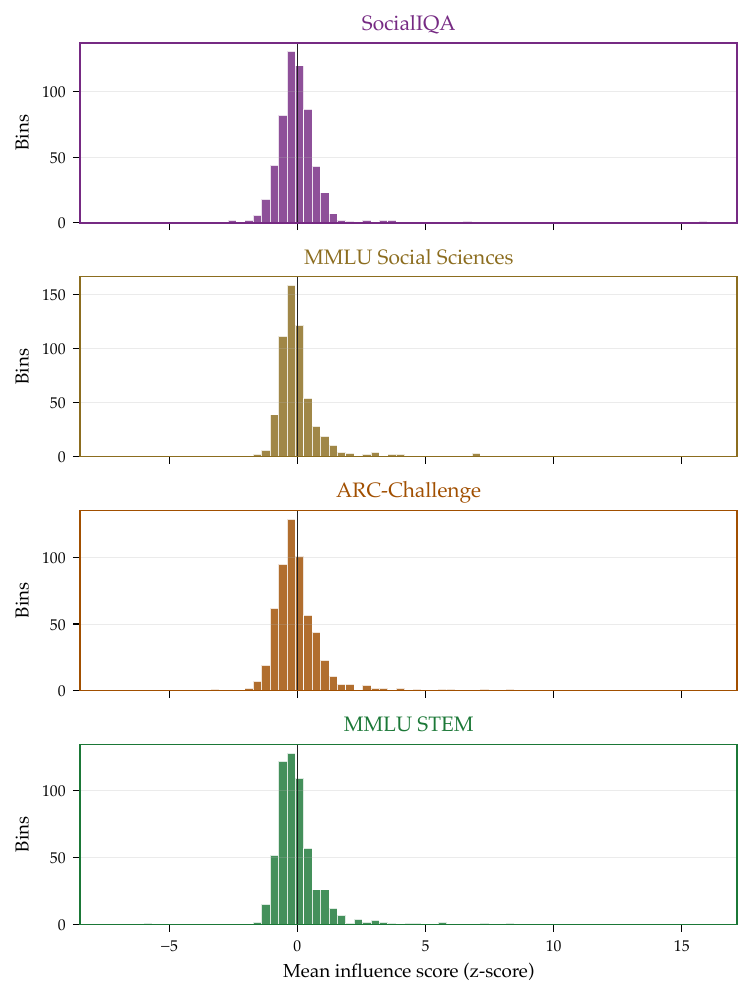}
\caption{Per-benchmark influence score distributions across all 576 WebOrganizer bins. Distributions are centered near zero with comparable spread, while benchmark-specific behavior appears in the tails.}\label{fig:influence-hist-overlay}
\end{figure}

Figure~\ref{fig:influence-hist-overlay} characterizes the influence score distribution across the four primary benchmarks before any aggregation, confirming approximate symmetry around zero with benchmark-specific tail behavior. Topic-level cross-benchmark summaries appear in Appendix~\ref{app:format-marginals}.

\FloatBarrier

\section{Attribution Pipeline Details}\label{app:attribution}

\subsection{Background: TrackStar and Bergson}

We compute gradient-based training-data attribution following
TrackStar~\citep{chang_scalable_2024} as implemented in Bergson. TrackStar
projects per-example gradients into a low-dimensional space using
Rademacher random projections and computes influence as

\[
s(j, i) = \sum_{k \in \mathcal{K}} \hat{g}_k(x_j)^\top \hat{g}_k(q_i),
\]

where $\hat{g}_k(\cdot)$ is the unit-normalized projected gradient at module
$k$. \S\ref{app:precond} describes the mixed preconditioner used in
practice; the unpreconditioned form above is retained only for exploratory
runs.

\subsection{Query Construction}

\begin{table}[htbp]
\centering
\caption{Core benchmarks used as attribution queries in the main design.}\label{tab:query-benchmarks}
\setlength{\tabcolsep}{5pt}
\begin{tabular}{>{\raggedright\arraybackslash}p{3.0cm} r
                >{\raggedright\arraybackslash}p{3.2cm}
                >{\raggedright\arraybackslash}p{3.0cm}}
\toprule
\rowcolor{SocialTDALightGray}
\textbf{Benchmark} & \textbf{Queries} & \textbf{Evaluation format} & \textbf{Query type} \\
\midrule
\rowcolor{SocialTDAPurple!8}
\SocialReasoning{SocialIQA}~\citep{sap2019socialiqacommonsensereasoningsocial} & \num{10000} & 3-shot MC, Accuracy & \SocialReasoning{Social reasoning} \\
\rowcolor{SocialTDAGold!8}
\SocialKnowledge{MMLU Social Sciences}~\citep{hendrycks2021measuringmassivemultitasklanguage} & \num{3077} & 4-shot MC, Accuracy & \SocialKnowledge{Social knowledge} \\
\rowcolor{SocialTDAOrange!8}
\STEMReasoning{ARC-Challenge}~\citep{clark2018think} & \num{1172} & 4-shot MC, Accuracy & \STEMReasoning{STEM reasoning} \\
\rowcolor{SocialTDAGreen!8}
\STEMKnowledge{MMLU STEM}~\citep{hendrycks2021measuringmassivemultitasklanguage} & \num{3018} & 4-shot MC, Accuracy & \STEMKnowledge{STEM knowledge} \\
\bottomrule
\end{tabular}
\end{table}

Each benchmark sample is an attribution query: it asks which training
documents most influence the model's behavior on
that instance. We evaluate with OLMES~\citep{gu2025olmes} following the
OLMo3 base evaluation suite: SocialIQA, MMLU Social Sciences,
ARC-Challenge, and MMLU STEM are scored as multiple-choice accuracy, and
MMLU subscores are macro-averaged. OLMES outputs are converted to
Bergson-compatible JSONL records (query and correct completion), so the
recorded gradients reflect the true cross-entropy loss.
The \num{10000} SocialIQA queries come from the OLMES Hugging Face
\texttt{train} split used for evaluation packaging, not from the standard
SocialIQA validation partition.

\subsection{Corpus Indexing}

Corpus characterization (de-duplication, distribution, fill rates) is in
Appendix~\ref{app:corpus}. For attribution, each document is stored as a
JSONL record with two fields---\texttt{id} (stable corpus identifier) and
\texttt{text} (document text)---sharded round-robin across
\CorpusNumShards{} files for corpus storage. The stratified attribution
working set is re-sharded into ${\sim}$\AttrShardCount{} shard files of
\AttrDocsPerShard{} documents each for gradient indexing
(Appendix~\ref{app:compute}).

\subsection{Gradient Computation}

For each document, we compute per-sample gradients of cross-entropy loss
with respect to all linear-layer weights (Q/K/V/O and MLP projections);
Table~\ref{tab:gradient-params} lists the projection dimension, precision,
and token-batching parameters used at index build time.

\begin{table}[htbp]
\centering
\caption{Gradient computation parameters.}\label{tab:gradient-params}
\begin{tabular}{ll}
\toprule
\rowcolor{SocialTDALightGray}
\textbf{Parameter} & \textbf{Value} \\
\midrule
Projection & Rademacher random \\
Dimension & $d{=}16$ ($256$ per module) \\
Precision & fp32 \\
Token batching & 4{,}096 tokens \\
Truncation & Enabled \\
Normalization & Unit norm \\
\bottomrule
\end{tabular}
\end{table}

\subsection{Preconditioner}\label{app:precond}

We use a two-model attribution setup following~\citet{ruis_procedural_2024}:
document gradients and base-side second-order information are computed with
OLMo3-7B Base, while query gradients are computed with OLMo3-7B Instruct, the
checkpoint used to score benchmark completions. Following that setup, we treat
the supervised-instruction-tuning second-order contribution as identity and
attribute only to pretraining data. This approximation keeps the metric tied
to pretraining curvature rather than adding an unobserved post-training
curvature term. The TrackStar preconditioner therefore supplies the
base-curvature metric in which base-document gradients and instruct-query
gradients are compared:

\[
s(j,i) = \sum_k \tilde{g}^{\mathrm{base}}_k(x_j)^\top
\tilde{g}^{\mathrm{inst}}_k(q_i),
\]

where $\tilde{g}$ denotes projected, unit-normalized, preconditioned
gradients for module $k$.

The mixed preconditioner is constructed in three steps:
\begin{enumerate}
    \item Build a value preconditioner from 100k random training documents.
    \item Build a query preconditioner from evaluation queries.
    \item Combine with 1{,}000-component downweighting.
\end{enumerate}

Exploratory runs skip preconditioning; all reported results include it.
We also computed Base-query variants as calibration checks; the reported
figures and tables use the Base-document/Instruct-query setup because the
Instruct checkpoint follows the OLMES prompts more reliably.

\subsection{Pipeline Architecture}

The pipeline consists of three stages:
\begin{enumerate}
    \item \textbf{Reduce}: compute the mean query gradient per benchmark.
    \item \textbf{Score}: compute document influence scores against the
          prebuilt gradient index.
    \item \textbf{Aggregate}: collect and rank the top-$k$ documents.
\end{enumerate}

GPU work is concentrated in index construction: one build job per
working-set shard (${\sim}$\AttrShardCount{} shards of \AttrDocsPerShard{}
documents each), plus per-benchmark query-gradient builds. Once the index
is built, scoring a query set runs as a parallel CPU pass. Compute
accounting and hardware are in Appendix~\ref{app:compute}.

\FloatBarrier

\section{Bin-Level Influence Heatmaps and Format Marginals}\label{app:format-marginals}

\begin{figure*}[tb]
  \centering
  \includegraphics[width=\MainFigureWidth]{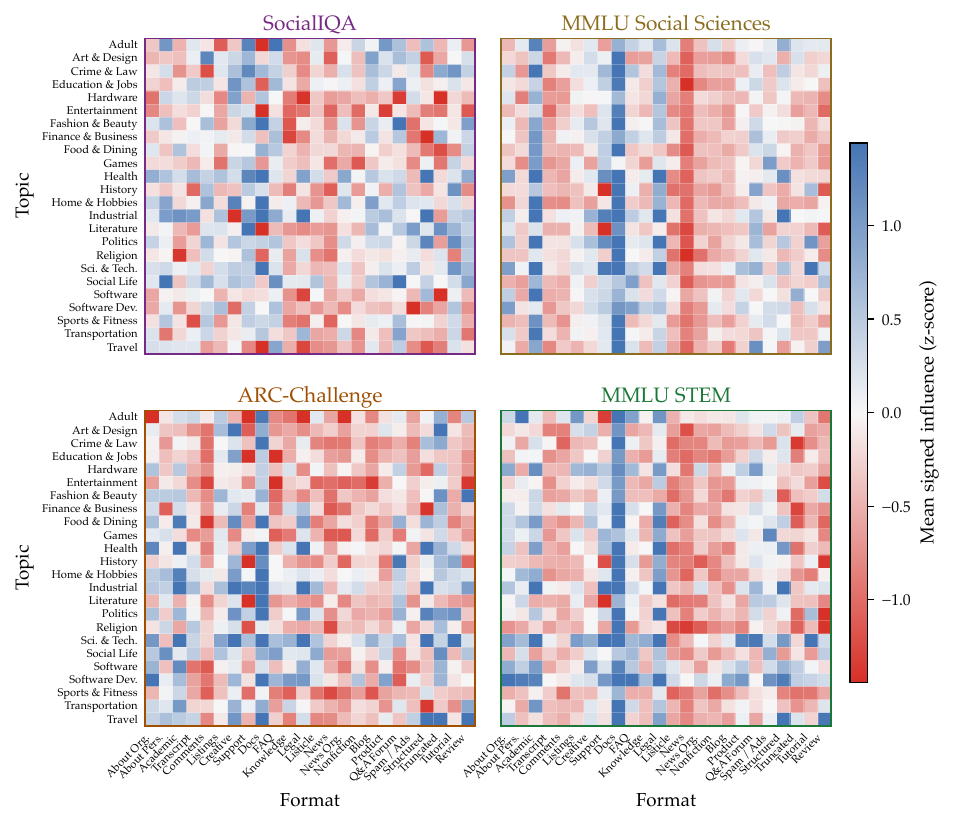}
  \InfluenceSignLegend
  \caption{Signed mean influence (z-scores) by topic (rows) and format (columns). Top row: social benchmarks; bottom row: STEM benchmarks. The shared 95th-percentile color scale uses \PositiveAttribution{blue for positive influence}, white near zero, and \NegativeAttribution{red for negative influence}; more extreme cells saturate. \SocialReasoning{SocialIQA}'s positive mass is dispersed across many topic--format cells, whereas the other three benchmarks---MMLU Social Sciences most visibly---concentrate positive influence in documentation-like formats, giving them a stronger column (format) structure.}
  \label{fig:influence-2x2}
\end{figure*}

\begin{figure*}[!tb]
  \centering
  \includegraphics[width=\MainFigureWidth]{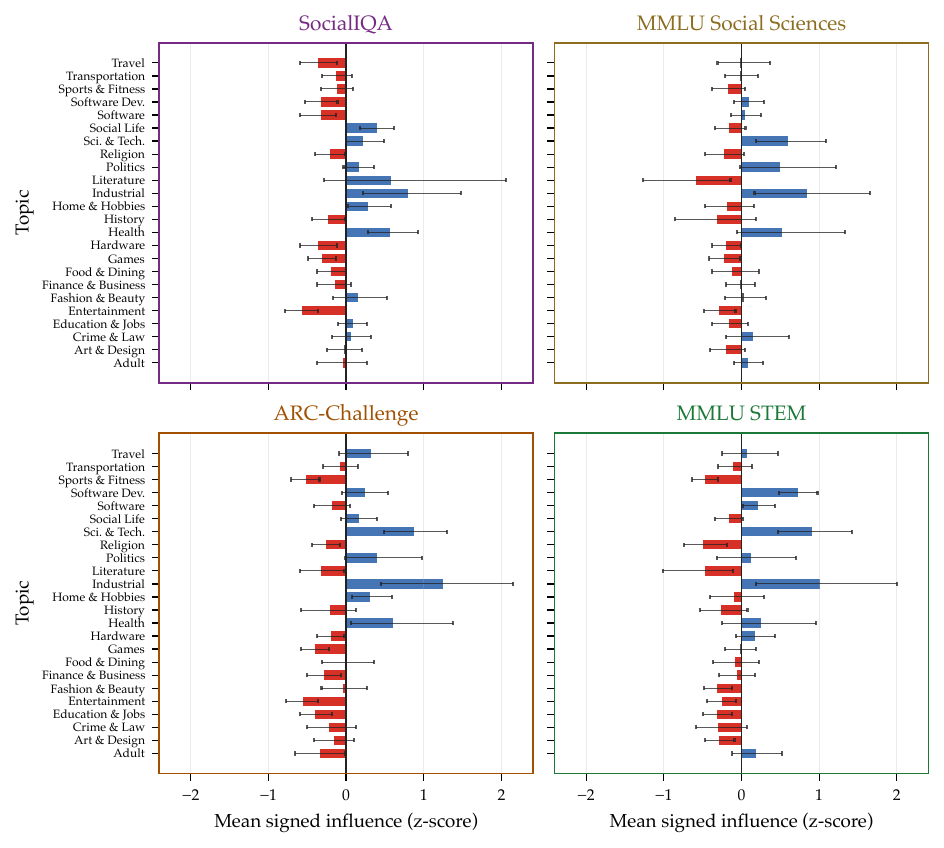}
  \InfluenceSignLegend
  \caption{Mean signed influence by WebOrganizer topic, aggregated across formats. Values are z-scores; bars are colored by influence sign, with positive values indicating supportive influence and negative values indicating suppressive influence.}
  \label{fig:topic-grouped-4way}
\end{figure*}

\begin{figure}[htbp]
  \centering
  \includegraphics[width=\MainFigureWidth]{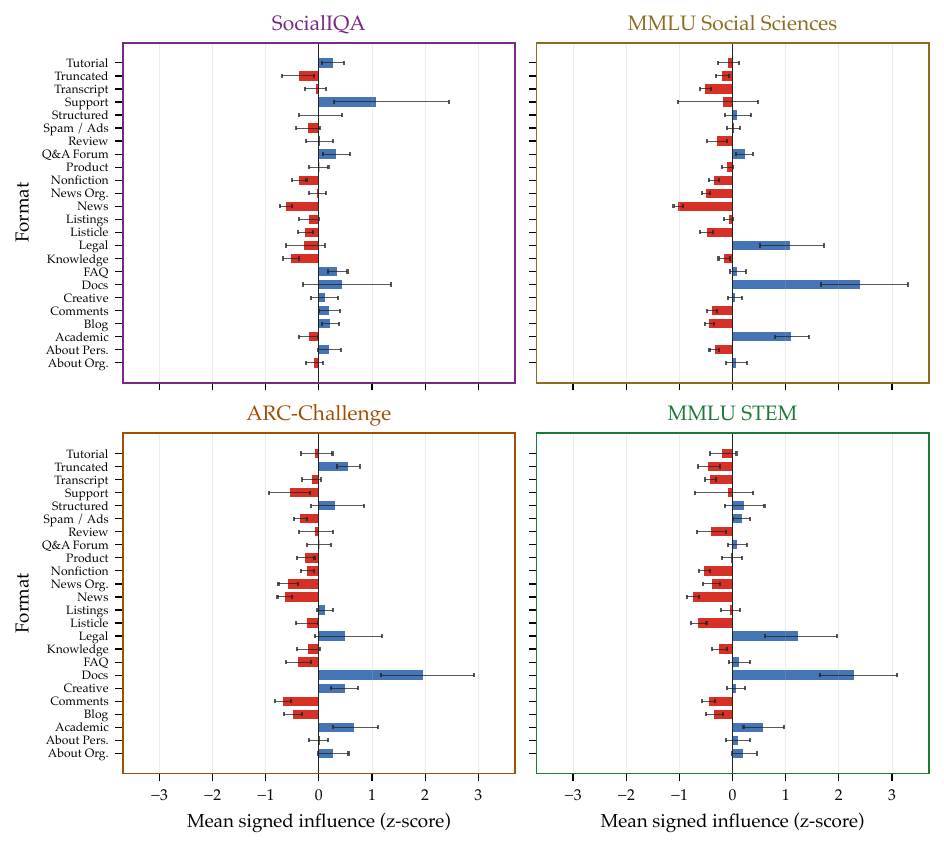}
  \InfluenceSignLegend
  \caption{Mean signed influence by WebOrganizer format, aggregated across topics. Values are within-benchmark z-scores; bars are colored by influence sign. Documentation-like formats dominate the STEM and knowledge benchmarks, while SocialIQA peaks on more dialogue-rich formats.}
  \label{fig:format-grouped-4way}
\end{figure}

Figure~\ref{fig:influence-2x2} shows the full signed bin-level influence
heatmaps for all four benchmarks on a shared color scale; the main text
(Figure~\ref{fig:rq1-format-concentration}) summarizes their most diagnostic
format and topic contrasts. Figures~\ref{fig:topic-grouped-4way}
and~\ref{fig:format-grouped-4way} show the topic and format marginals that
summarize these heatmaps. Table~\ref{tab:format-marginals} names the top-3
positive and top-3 negative formats per benchmark.

\begin{table}[htbp]
\centering
\caption{Top-3 positive and top-3 negative WebOrganizer formats per
benchmark, ranked by mean signed influence z-score (averaged across
topics within each format). Values are derived from the same z-scored
bin scores that drive Figure~\ref{fig:format-grouped-4way}.
\emph{Documentation} appears in every benchmark's top-3;
SocialIQA's top format \emph{Customer Support} does not appear in any
other benchmark's top-3; \emph{News Article} appears in all four
bottom-3 lists.}\label{tab:format-marginals}
\scriptsize
\setlength{\tabcolsep}{3pt}
\renewcommand{\arraystretch}{1.12}
\newcommand{\FormatBenchmark}[2]{%
  \cellcolor{#1!7}\multirow{2}{=}{\textcolor{#1!88!black}{\textbf{#2}}}%
}
\newcommand{\FormatBenchmarkBlank}[1]{\cellcolor{#1!7}}
\newcommand{\FormatMarginalList}[3]{%
  \begin{minipage}[t]{\linewidth}
    \raggedright #1\\[-1pt]#2\\[-1pt]#3
  \end{minipage}%
}
\newcommand{\FormatPositive}[2]{\PositiveAttribution{#1}\quad #2}
\newcommand{\FormatNegative}[2]{\NegativeAttribution{#1}\quad #2}
\begin{tabularx}{\columnwidth}{@{}>{\raggedright\arraybackslash}p{0.27\columnwidth}
                              >{\raggedright\arraybackslash}p{0.18\columnwidth}
                              >{\raggedright\arraybackslash}X@{}}
\toprule
\textbf{Benchmark} & \textbf{Sign} & \textbf{Top-3 formats} \\
\midrule
\FormatBenchmark{SocialTDAPurple}{SocialIQA} &
\cellcolor{SocialTDAPositive!8}\PositiveAttribution{Positive} &
\cellcolor{SocialTDAPositive!5}\FormatMarginalList
  {\FormatPositive{$+1.07$}{Customer Support}}
  {\FormatPositive{$+0.44$}{Documentation}}
  {\FormatPositive{$+0.35$}{FAQ}} \\
\FormatBenchmarkBlank{SocialTDAPurple} &
\cellcolor{SocialTDANegative!7}\NegativeAttribution{Negative} &
\cellcolor{SocialTDANegative!5}\FormatMarginalList
  {\FormatNegative{$-0.38$}{Truncated}}
  {\FormatNegative{$-0.53$}{Knowledge Article}}
  {\FormatNegative{$-0.62$}{News Article}} \\
\addlinespace
\FormatBenchmark{SocialTDAGold}{MMLU Social Sciences} &
\cellcolor{SocialTDAPositive!8}\PositiveAttribution{Positive} &
\cellcolor{SocialTDAPositive!5}\FormatMarginalList
  {\FormatPositive{$+2.40$}{Documentation}}
  {\FormatPositive{$+1.10$}{Academic Writing}}
  {\FormatPositive{$+1.08$}{Legal Notices}} \\
\FormatBenchmarkBlank{SocialTDAGold} &
\cellcolor{SocialTDANegative!7}\NegativeAttribution{Negative} &
\cellcolor{SocialTDANegative!5}\FormatMarginalList
  {\FormatNegative{$-0.49$}{News (Organization)}}
  {\FormatNegative{$-0.51$}{Audio Transcript}}
  {\FormatNegative{$-1.02$}{News Article}} \\
\addlinespace
\FormatBenchmark{SocialTDAOrange}{ARC-Challenge} &
\cellcolor{SocialTDAPositive!8}\PositiveAttribution{Positive} &
\cellcolor{SocialTDAPositive!5}\FormatMarginalList
  {\FormatPositive{$+1.97$}{Documentation}}
  {\FormatPositive{$+0.66$}{Academic Writing}}
  {\FormatPositive{$+0.56$}{Truncated}} \\
\bottomrule
\end{tabularx}
\end{table}

\emph{Documentation} is the strongest positive format for MMLU Social
Sciences ($z{=}{+}2.40$), MMLU STEM ($+2.29$), and ARC-Challenge
($+1.97$), and remains positive for SocialIQA ($+0.44$). This
cross-benchmark consistency parallels the topic-level positivity of
Industrial in Figure~\ref{fig:topic-grouped-4way}. SocialIQA diverges at
the peak: \emph{Customer Support} ($+1.07$) is its top format and does
not appear in any other benchmark's top-3, mirroring the dialogue-rich
character of SocialIQA's top-influence bins
(Appendix~\ref{app:bin-characterization}). \emph{News Article} registers
as a consistent negative across all four benchmarks ($-0.62$ to $-1.02$),
distinct from the topic-level Literature
flip. Substantively, the documentation-like formats the knowledge and STEM
benchmarks rely on carry more direct factual content, whereas SocialIQA's
interpersonal formats (Customer Support, FAQ, Q\&A~Forum) carry dialogue-rich
social interaction.

\subsection{Do the benchmark profiles differ from each other?}\label{app:profile-divergence}

\begin{figure*}[tb]
  \centering
  \includegraphics[width=\MainFigureWidth]{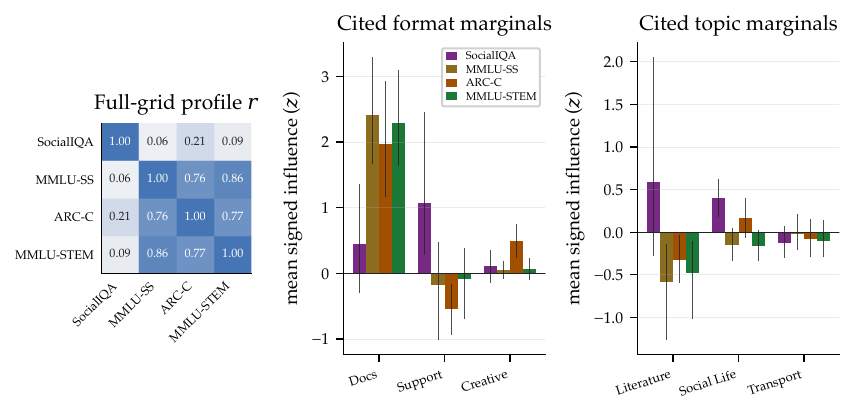}
  \caption{Between-benchmark profile divergence shows \SocialReasoning{SocialIQA}
  is the structural outlier independent of any single format. Each benchmark's
  bin-level influence is a vector over the 576 (topic$\times$format) bins; we
  compare \emph{positions} (which bins carry influence), not magnitudes, since
  scores are $z$-standardized within each benchmark.
  \textbf{Left:} pairwise Pearson correlation of the full-grid profiles, where
  SocialIQA correlates with each comparison benchmark at $r\!\le\!0.21$ while the
  three comparison benchmarks correlate at $r=0.76$--$0.86$ among themselves.
  \textbf{Center, Right:} the format and topic marginals the main text cites,
  with $95\%$ bin-bootstrap CIs. SocialIQA separates from the comparison set on
  Documentation, Customer Support, and Social Life; the Literature gap is
  positive but not CI-separated, and the Creative Writing and Transportation
  negative checks do not separate.}
  \label{fig:profile-divergence}
\end{figure*}

The per-bin tests above ask whether a bin's influence differs from zero within a
benchmark. A complementary question is whether the four profiles differ
\emph{from each other}. Because influence is $z$-standardized within each
benchmark, the marginal score distributions are identical by construction, so we
compare \emph{positions} rather than magnitudes, treating each benchmark's
profile as a vector over the 576 bins and measuring how much two benchmarks
concentrate influence on the same bins (Pearson correlation;
Figure~\ref{fig:profile-divergence}).

\SocialReasoning{SocialIQA}'s full-grid profile correlates with the three
comparison benchmarks at $r=0.06$, $0.21$, and $0.09$, whereas those three
correlate at $r=0.76$--$0.86$ among themselves. An outlier statistic $\Delta_b$,
defined as the mean correlation among the other three benchmarks minus
benchmark $b$'s mean correlation to them, is positive only for SocialIQA at
$\Delta=+0.68$ ($95\%$ bin-bootstrap CI $[+0.34,+0.96]$), with a within-bin
label-permutation null giving $p\approx10^{-4}$. The same ranking holds
independently on the format marginals ($\Delta=+0.70$, $[+0.35,+1.07]$) and the
topic marginals ($+0.37$, $[+0.13,+0.88]$), so SocialIQA's status as the
structural outlier does not rest on any single format. Among the cited
marginals, the SocialIQA-minus-comparison-mean gap is CI-separated for
Documentation ($-1.78$, $[-2.43,-1.15]$), Customer Support ($+1.34$,
$[+0.18,+3.32]$), and Social Life ($+0.45$, $[+0.19,+0.71]$); the Literature
gap is positive but its CI includes zero ($+1.04$, $[-0.13,+2.98]$), as do the
Creative Writing and Transportation negative checks.

\paragraph{What is resampled, and the independence assumption.}
All intervals are nonparametric ($10{,}000$ resamples, fixed seed) and take the
$z$-scored bin as the unit of analysis. The full-grid quantities (the profile
correlations, $c_b$, and $\Delta_b$) use a \emph{bin} bootstrap: each replicate
draws the $N{=}576$ topic$\times$format bins with replacement and applies the
\emph{same} resampled set to all four benchmarks before recomputing the
$4\times4$ correlation matrix, so the cross-benchmark pairing is preserved and the
interval reflects sensitivity to bin composition. The marginal intervals instead
resample the \emph{collapsed} axis: the format-marginal CIs draw the 24 topics
(the axis averaged over to form a format profile) and the topic-marginal CIs draw
the 24 formats; the cited single-cell gap CIs (e.g.\ Documentation) likewise
resample the 24 topics or formats composing that marginal. The omnibus
$p$-value is a separate \emph{benchmark-label} permutation that assumes only
exchangeability, not independence: within each bin we independently permute the
four benchmark labels (reassigning that bin's four standardized values across
benchmarks), recompute $\Delta_{\mathrm{SocialIQA}}$, and report
$p=(1+\#\{\Delta^{\mathrm{perm}}\!\ge\!\Delta^{\mathrm{obs}}\})/(1+10{,}000)$.
The bin bootstrap treats bins as exchangeable resampling units and does
\emph{not} model the topic$\times$format dependence among them, which could make
the full-grid interval mildly anti-conservative; we therefore rely on the
agreement of three schemes resting on different assumptions: the bin bootstrap,
the topic- and format-resampled marginal bootstraps, and the label
permutation. We use bin-level rather than
document-level resampling deliberately: with ${\sim}10^4$ documents per bin a
document bootstrap would render almost any difference significant, so bin
composition is the binding, conservative source of uncertainty, the effect sizes
carry the claim, and the permutation $p$ is confirmatory.

\FloatBarrier

\section{Paired Topic Comparisons}\label{app:paired-topics}

\begin{figure*}[!tb]
  \centering
  \includegraphics[width=\MainFigureWidth,keepaspectratio]{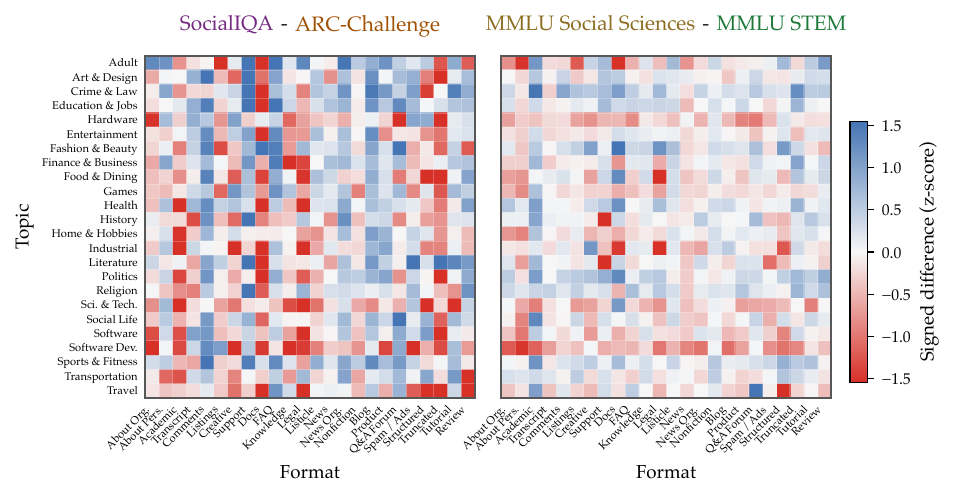}
  \InfluenceSignLegend
  \caption{Bin-level social-vs-STEM separation. Signed differences compare \SocialReasoning{SocialIQA} $-$ \STEMReasoning{ARC-Challenge} and \SocialKnowledge{MMLU Social Sciences} $-$ \STEMKnowledge{MMLU STEM} over WebOrganizer topic--format bins. The shared 95th-percentile color scale uses \PositiveAttribution{blue for positive differences favoring the social benchmark} and \NegativeAttribution{red for negative differences favoring the STEM benchmark}; more extreme cells saturate.}
  \label{fig:diff-combined-canonical}
\end{figure*}

\begin{figure}[htbp]
\centering
\includegraphics[width=\MainFigureWidth]{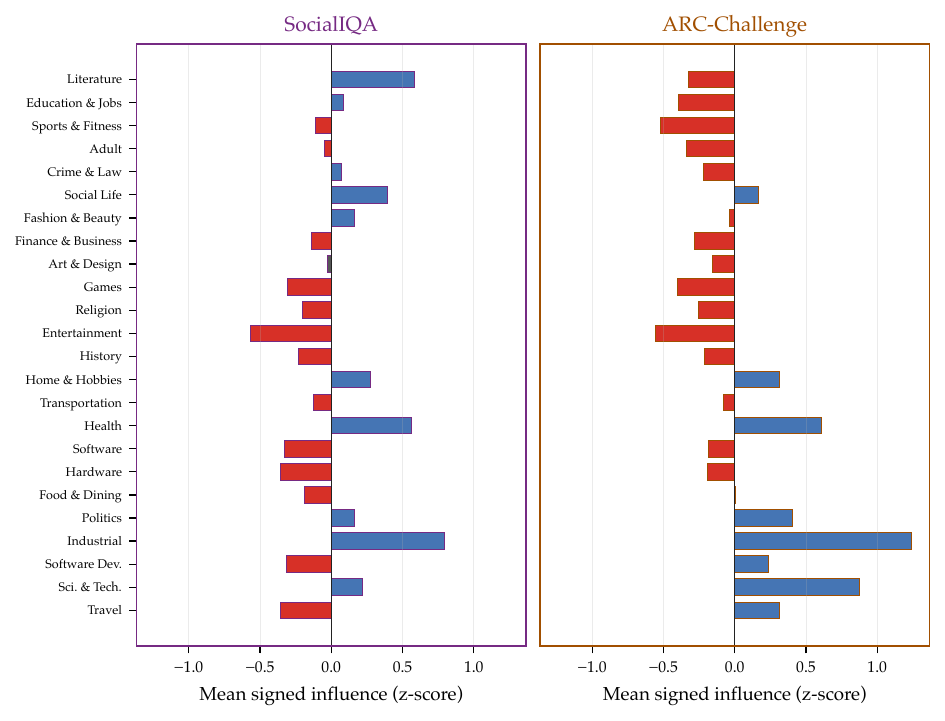}
\InfluenceSignLegend
\caption{Paired topic-level influence comparison: SocialIQA vs.\ ARC-Challenge. Each topic shows signed mean influence for both benchmarks on matched topic axes. Topics where signs diverge (e.g., Literature, Software Development) represent the strongest contrastive signal between social and STEM reasoning.}\label{fig:paired-socialiqa-arc}
\end{figure}

\begin{figure}[htbp]
\centering
\includegraphics[width=\MainFigureWidth]{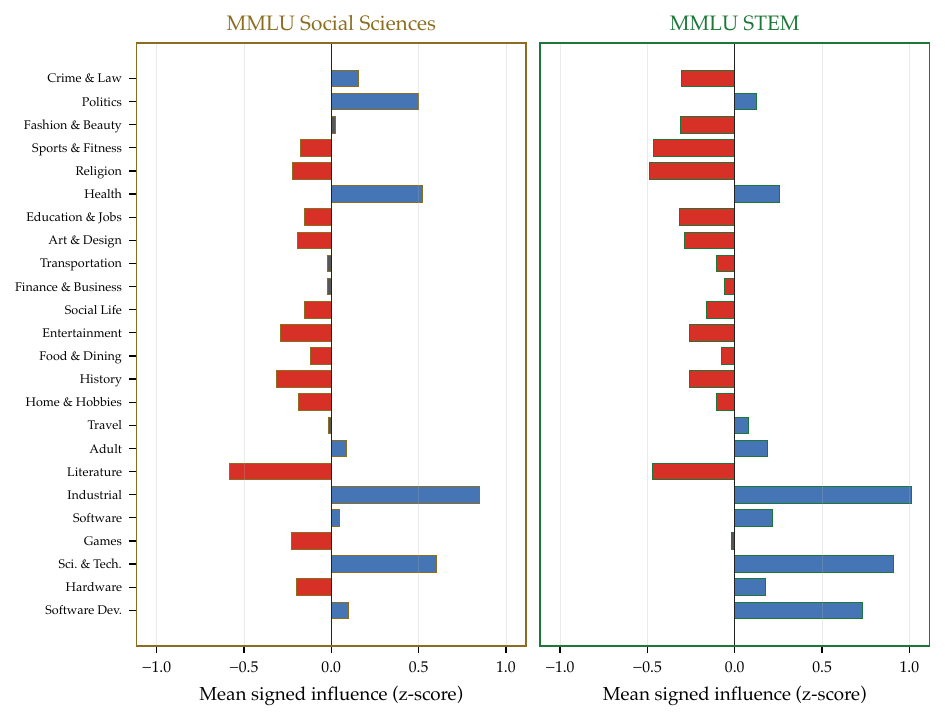}
\InfluenceSignLegend
\caption{Paired topic-level influence comparison: MMLU Social Sciences vs.\ MMLU STEM. The knowledge-domain contrast is directionally consistent with the reasoning contrast (Figure~\ref{fig:paired-socialiqa-arc}) but weaker in magnitude, consistent with the main text findings.}\label{fig:paired-mmlu-ss-stem}
\end{figure}

Figure~\ref{fig:diff-combined-canonical} provides the full bin-level
social-vs-STEM difference heatmaps. Figures~\ref{fig:paired-socialiqa-arc}
and~\ref{fig:paired-mmlu-ss-stem} provide direct side-by-side comparisons
of per-topic influence between contrastive benchmark pairs.
Table~\ref{tab:paired-contrasts} names the three topics most strongly
tilting toward each benchmark in each pair.

\begin{table}[htbp]
\centering
\caption{Top-3 topics tilting toward each side of the two contrastive
benchmark pairs, ranked by the difference $\Delta = z_A - z_B$ of
topic-level marginal influence z-scores. Reasoning contrasts
(SocialIQA vs.\ ARC-Challenge) span $|\Delta|$ up to 0.91, while
knowledge contrasts (MMLU Social Sciences vs.\ MMLU STEM) span $|\Delta|$
up to 0.63---quantifying the attenuation noted in the main
text~(\S\ref{sec:results}). Positive and negative columns follow the
signed-difference convention; \emph{Software Development} is
consistently STEM-leaning in both pairs.}\label{tab:paired-contrasts}
\small
\setlength{\tabcolsep}{4pt}
\renewcommand{\arraystretch}{1.12}
\newcommand{\ContrastBenchmark}[2]{%
  \begingroup
  \setlength{\fboxsep}{2pt}%
  \colorbox{#1!7}{%
    \makebox[\dimexpr\linewidth-2\fboxsep\relax][l]{%
      \textcolor{#1!88!black}{\textbf{#2}}%
    }%
  }%
  \endgroup
}
\newcommand{\ContrastPair}[4]{%
  \begin{minipage}[c]{\linewidth}
    \ContrastBenchmark{#1}{#2}\\[-1pt]
    {\scriptsize\textsf{vs.}}\\[-1pt]
    \ContrastBenchmark{#3}{#4}%
  \end{minipage}%
}
\newcommand{\TopicTiltList}[3]{%
  \begin{tabular}[t]{@{}l@{}}#1\\#2\\#3\end{tabular}%
}
\begin{tabularx}{\linewidth}{>{\raggedright\arraybackslash}p{0.25\linewidth}%
                >{\raggedright\arraybackslash}X%
                >{\raggedright\arraybackslash}X}
\toprule
\textbf{Contrast pair} &
\cellcolor{SocialTDAPositive!8}%
\begin{tabular}[t]{@{}l@{}}\textbf{Top-3 tilting A}\\[-1pt]
{\scriptsize\PositiveAttribution{positive $\Delta$}}\end{tabular} &
\cellcolor{SocialTDANegative!7}%
\begin{tabular}[t]{@{}l@{}}\textbf{Top-3 tilting B}\\[-1pt]
{\scriptsize\NegativeAttribution{negative $\Delta$}}\end{tabular} \\
\midrule
\ContrastPair{SocialTDAPurple}{SocialIQA}{SocialTDAOrange}{ARC-Challenge}
  & \cellcolor{SocialTDAPositive!6}\TopicTiltList{Literature (\PositiveAttribution{$+0.91$})}{Education \& Jobs (\PositiveAttribution{$+0.48$})}{Sports \& Fitness (\PositiveAttribution{$+0.41$})}
  & \cellcolor{SocialTDANegative!5}\TopicTiltList{Travel (\NegativeAttribution{$-0.67$})}{Science \& Technology (\NegativeAttribution{$-0.66$})}{Software Development (\NegativeAttribution{$-0.56$})} \\
\addlinespace
\ContrastPair{SocialTDAGold}{MMLU Social Sciences}{SocialTDAGreen}{MMLU STEM}
  & \cellcolor{SocialTDAPositive!6}\TopicTiltList{Crime \& Law (\PositiveAttribution{$+0.46$})}{Politics (\PositiveAttribution{$+0.38$})}{Fashion \& Beauty (\PositiveAttribution{$+0.34$})}
  & \cellcolor{SocialTDANegative!5}\TopicTiltList{Software Development (\NegativeAttribution{$-0.63$})}{Hardware (\NegativeAttribution{$-0.38$})}{Science \& Technology (\NegativeAttribution{$-0.31$})} \\
\bottomrule
\end{tabularx}
\end{table}

\emph{Software Development} is consistently STEM-leaning in both
contrasts ($\Delta{=}{-}0.56$ for SocialIQA vs.\ ARC-Challenge and
$\Delta{=}{-}0.63$, the most STEM-leaning topic, for MMLU Social Sciences
vs.\ MMLU STEM); in the reasoning contrast the most STEM-leaning topics are
\emph{Travel} (${-}0.67$) and \emph{Science~\&~Technology} (${-}0.66$).
\emph{Literature} is the strongest social-leaning topic in the reasoning
contrast ($\Delta{=}{+}0.91$, sign-flipped between the two benchmarks); the
knowledge contrast has no single dominant social-leaning topic of comparable
magnitude---\emph{Crime~\&~Law} ($+0.46$) and \emph{Politics} ($+0.38$)
lead but at roughly half that size. The maximum gap of 0.91~$z$-units
in the reasoning pair versus 0.63 in the knowledge pair quantifies the
attenuation reported in the main text: reasoning contrasts are sharper
than knowledge contrasts.

\FloatBarrier

\section{Correctness Differential}\label{app:correctness}

\begin{table}[htbp]
\centering
\caption{Per-benchmark correctness-differential extremes: bin with the
strongest positive $\Delta z$ (correct minus incorrect; appears more on
correctly answered queries) and the strongest negative $\Delta z$
(appears more in errors). \emph{Literature~$\times$~Customer Support} is the
top-positive bin for both reasoning benchmarks, SocialIQA ($+13.1$) and
ARC-Challenge ($+6.0$), and the strongest \emph{negative} bin for both
knowledge benchmarks, MMLU Social Sciences and MMLU STEM: a
reasoning-vs-knowledge sign flip on a single bin. Brackets give bootstrap
95\% CIs (2{,}000 resamples over queries); the four top-positive extremes
exclude zero, while all four top-negative extremes include zero. SocialIQA's differential is
the largest, about $2.2\times$ ARC-Challenge's; the two MMLU benchmarks stay
weak ($|\Delta z|\le 2.2$). MMLU STEM's top-positive bin lands in a small,
underfilled Documentation cell (Fashion~\&~Beauty~$\times$~Documentation;
Table~\ref{tab:top-bins}), while MMLU Social Sciences' top-positive
Health~$\times$~Documentation bin is fully
populated.}\label{tab:correctness-extremes}
\small
\setlength{\tabcolsep}{4pt}
\begin{tabular}{l r >{\raggedright\arraybackslash}p{3.1cm}
                  r >{\raggedright\arraybackslash}p{3.1cm}}
\toprule
\textbf{Benchmark} & \multicolumn{2}{c}{\PositiveAttribution{Top positive bin}} & \multicolumn{2}{c}{\NegativeAttribution{Top negative bin}} \\
\cmidrule(lr){2-3}\cmidrule(lr){4-5}
 & $\Delta z$ & Bin (Topic $\times$ Format) & $\Delta z$ & Bin (Topic $\times$ Format) \\
\midrule
\rowcolor{SocialTDAPurple!7}
\SocialReasoning{SocialIQA}        & \cellcolor{SocialTDAPositive!8}\PositiveAttribution{\shortstack[r]{$+13.11$\\{\scriptsize $[2.8,\,24.3]$}}} & Literature $\times$ Customer Support     & \cellcolor{SocialTDANegative!8}\NegativeAttribution{\shortstack[r]{$-4.60$\\{\scriptsize $[-7.7,\,0.8]$}}} & Health $\times$ Documentation \\
\rowcolor{SocialTDAGold!7}
\SocialKnowledge{MMLU Social Sciences} & \cellcolor{SocialTDAPositive!8}\PositiveAttribution{\shortstack[r]{$+0.86$\\{\scriptsize $[0.1,\,1.8]$}}}  & Health $\times$ Documentation           & \cellcolor{SocialTDANegative!8}\NegativeAttribution{\shortstack[r]{$-2.19$\\{\scriptsize $[-5.4,\,1.1]$}}} & Literature $\times$ Customer Support \\
\rowcolor{SocialTDAOrange!7}
\STEMReasoning{ARC-Challenge}    & \cellcolor{SocialTDAPositive!8}\PositiveAttribution{\shortstack[r]{$+6.00$\\{\scriptsize $[0.2,\,10.8]$}}}  & Literature $\times$ Customer Support & \cellcolor{SocialTDANegative!8}\NegativeAttribution{\shortstack[r]{$-1.31$\\{\scriptsize $[-2.6,\,0.3]$}}} & Politics $\times$ Legal Notices \\
\rowcolor{SocialTDAGreen!7}
\STEMKnowledge{MMLU STEM}        & \cellcolor{SocialTDAPositive!8}\PositiveAttribution{\shortstack[r]{$+1.83$\\{\scriptsize $[0.5,\,3.2]$}}}  & Fashion \& Beauty $\times$ Documentation & \cellcolor{SocialTDANegative!8}\NegativeAttribution{\shortstack[r]{$-2.17$\\{\scriptsize $[-6.1,\,1.6]$}}} & Literature $\times$ Customer Support \\
\bottomrule
\end{tabular}
\end{table}

\begin{figure*}[!tb]
  \centering
  \includegraphics[width=\MainFigureWidth,keepaspectratio]{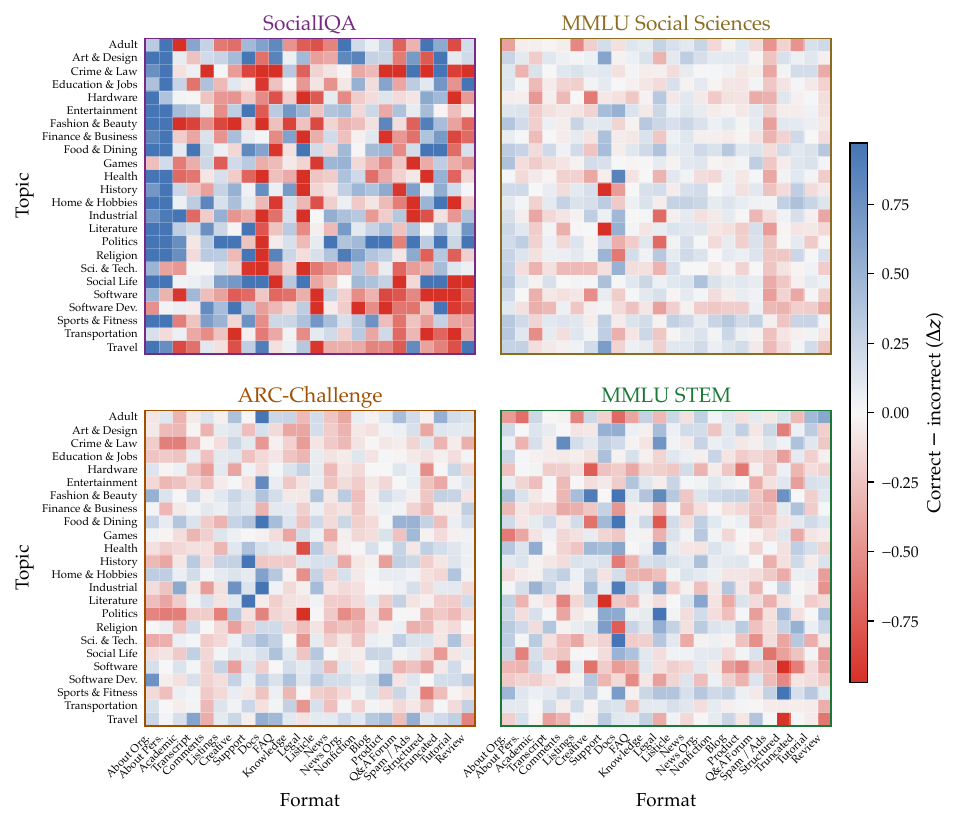}
  \InfluenceSignLegend
  \caption{Correctness differential: mean influence on correctly answered queries minus mean influence on incorrectly answered queries, per WebOrganizer bin. \PositiveAttribution{Positive} values support correct responses more than errors; \NegativeAttribution{negative} values indicate the reverse; more extreme cells saturate on the shared 95th-percentile color scale.}
  \label{fig:diff-correct-2x2}
\end{figure*}

For each benchmark we separately compute influence over the correctly
answered and incorrectly answered query subsets and report the
difference $\Delta z = z_{\text{correct}} - z_{\text{incorrect}}$ per
WebOrganizer~\citep{wettig2025weborganizer} bin. SocialIQA shows the strongest correctness
differential, with maximum $|\Delta z|{=}13.1$, about $2.2\times$ the next
benchmark, ARC-Challenge ($+6.0$); the two MMLU benchmarks stay weak
($|\Delta z|\le 2.2$)
(Table~\ref{tab:correctness-extremes}). The composite heatmap
(Figure~\ref{fig:diff-correct-2x2}) shows high-magnitude positive and negative cells
distributed across many topic--format bins of SocialIQA's panel and
consistent coloring within topic rows. ARC-Challenge's panel is dominated by a
strongly positive \emph{Literature~$\times$~Customer Support} cell, with a few
further cells (for example History~\&~Geography~$\times$~Customer Support and
Industrial~$\times$~Documentation) also reaching the edge of the shared color
scale; the two MMLU panels remain uniformly muted. A provenance caveat applies
to the ARC splits specifically: their per-query correct/incorrect alignment is
reconstructed positionally from recovered evaluation artifacts rather than
re-verified against original per-query identifiers, so the ARC correctness
partitions carry weaker provenance than the other benchmarks' splits.

\paragraph{Reasoning-vs-knowledge sign flip.}
\emph{Literature~$\times$~Customer Support} is the dominant outlier and the
top-positive correctness bin for \emph{both} reasoning benchmarks: it is the
strongest correctness-differential bin for SocialIQA ($\Delta z{=}{+}13.1$) and
for ARC-Challenge ($\Delta z{=}{+}6.0$), while being the strongest bin
associated with \emph{incorrect} answers on the two knowledge benchmarks, MMLU
Social Sciences ($-2.2$) and MMLU STEM ($-2.2$). Bootstrap 95\% CIs over
queries (2{,}000 resamples) qualify the flip: the positive arm is significant
on both reasoning benchmarks (SocialIQA $[2.8,\,24.3]$, ARC-Challenge
$[0.2,\,10.8]$, both excluding zero), whereas the negative arm on the
knowledge benchmarks is not (MMLU Social Sciences $[-5.4,\,1.1]$, MMLU STEM
$[-6.1,\,1.6]$, both spanning zero), so the suppressive side of the flip is
directional rather than statistically resolved. The underlying mechanism
differs by benchmark: for SocialIQA both cohorts are positive
($z_{\text{correct}}{=}{+}17.6$, $z_{\text{incorrect}}{=}{+}4.5$), so this
content actively supports correct social reasoning; for ARC-Challenge both are
negative ($z_{\text{correct}}{=}{-}0.8$, $z_{\text{incorrect}}{=}{-}6.8$), so it
is net-harmful but far less so on correctly answered queries. Either way, the same
corpus content---short, dialogue-rich book-club Q\&A text whose lexical profile
is detailed in Appendix~\ref{app:bin-characterization}---separates correct
reasoning from correct factual recall in this diagnostic. SocialIQA's strongest
\emph{negative} bin is the mirror image: Health~$\times$~Documentation
($\Delta z{=}{-}4.6$), a documentation-format bin that is positive for the
knowledge benchmarks. The two MMLU benchmarks have weak correctness
differentials overall ($|\Delta z|\le 2.2$). MMLU STEM's top-positive bin
additionally lands in a small, underfilled Documentation cell
(Fashion~\&~Beauty~$\times$~Documentation, one of the corpus's empty-tail bins;
Table~\ref{tab:top-bins}), so its signal is support-limited on top of being
weak; MMLU Social Sciences' top-positive Health~$\times$~Documentation bin is
fully populated, so its weakness is not a support artifact.

The asymmetry is itself informative. The strong correctness
differentials of the two reasoning benchmarks (SocialIQA and ARC-Challenge)
suggest that specific corpus regions contribute differently to correct vs.\
incorrect reasoning, while the weaker signal for the two knowledge benchmarks
(both MMLUs) is consistent with a more diffuse relationship between corpus
content and item-level accuracy.

\FloatBarrier

\section{Held-out Probe Analysis for Social Reasoning}\label{app:heldout-suite}

This appendix collects the paper's held-out probes: benchmarks that were not
part of the four core attribution-query design in Table~\ref{tab:query-benchmarks}.
The goal is breadth, not a separate unlearning claim. OLMES emits per-query
predictions, correctness labels, subset metadata, chance baselines, and Wilson
intervals. TrackStar then aggregates signed influence over the
\WebOrgNumTopics{} topic by \WebOrgNumFormats{} format taxonomy. We first check
whether the model solves each probe above chance, because attribution is only
interpretable as capability evidence when the evaluated mask has usable
prediction signal.

\subsection{Setup and Query Construction}
Table~\ref{tab:heldout-query-construction} records the held-out query surface.
The table is separate from Table~\ref{tab:query-benchmarks} because the latter
defines the four core in-paper attribution queries. Here, the query set is an
audit of additional social, pragmatic, moral, bias, and Theory-of-Mind probes.
StereoSet~\citep{nadeem_stereoset_measuring_2021} is included only to document the
boundary: its official evaluation is native likelihood, so it is not used in
the direct-answer TrackStar attribution headline.

\begin{table}[!htbp]
\centering
\scriptsize
\setlength{\tabcolsep}{3pt}
\begin{tabularx}{\linewidth}{@{}p{0.20\linewidth}r p{0.21\linewidth}p{0.16\linewidth}X@{}}
\toprule
Probe & Queries & Scoring format & Role & Caveat \\
\midrule
BBQ & 58,492 & Direct MC plus bias & Main & Bias benchmark; interpret with category-level bias diagnostics. \\
BBH Disambiguation QA & 250 & Forced choice & Main & Small-$N$ BBH row, but clear margin above chance. \\
BBH Causal Judgment & 187 & Forced choice & Caution & Wilson interval reaches chance. \\
BBH Snarks & 178 & 3-shot CoT MC & Supporting & Pragmatic BBH check; attribution is noisier because the query set is small. \\
ToMBench & 2,860 & Direct MC & Main & Public merged source caveat retained from the OLMES card. \\
SimpleToM & 3,441 & Direct suite & Supporting/control & Mental-state QA is positive; behavior and judgment QA are controls. \\
NegotiationToM & 28,560 & Direct MC, desire/belief & Main & Intention and all-round splits are not in the current scored suite. \\
PUB & 26,742 & Direct MC & Main with retained mask & \texttt{pub\_2}/\texttt{pub\_3} are excluded from the main PUB mask. \\
MMLU moral/humanities & 1,652 & OLMES MC subjects & Main & Moral and humanities subject grouping. \\
ETHICS & 19,967 & Direct MC & Secondary/control & Non-justice retained as secondary; justice is a failure control. \\
MORABLES & 709 & Direct MC & Secondary & Low absolute accuracy and CC-BY-NC-4.0 caveat. \\
MoralExceptQA/RBQA & 148 & Forced choice & Secondary & Small-$N$ row with CC-BY-NC-SA-4.0 caveat. \\
StereoSet & 4,229 & Native likelihood & Excluded from attribution headline & Reported only as native-likelihood bias evidence, not direct-answer TrackStar attribution. \\
\bottomrule
\end{tabularx}
\caption{Held-out query construction and inclusion policy. Query counts are the exported OLMES or diagnostic request rows used by the held-out artifact bundle.}
\label{tab:heldout-query-construction}
\end{table}

\subsection{Direct-Answer Gate and Tiering}
Table~\ref{tab:heldout-tier-policy} defines the tiering rule before attribution
figures are interpreted. Main held-out probes must be clearly above chance, relevant
to social, pragmatic, moral, negotiation, or bias attribution, and supported by
the updated OLMES and TrackStar artifacts. Secondary rows remain part of the
audit trail but carry less weight. Failure controls and excluded rows define
the boundary of the direct-answer claims.

\begin{table}[!htbp]
\centering
\scriptsize
\setlength{\tabcolsep}{4pt}
\begin{tabularx}{\linewidth}{@{}p{0.18\linewidth}p{0.36\linewidth}X@{}}
\toprule
Tier & Benchmarks or masks & Role in the paper \\
\midrule
Main & BBQ; BBH Disambiguation QA; ToMBench; MMLU moral/humanities; NegotiationToM desire/belief; PUB excluding \texttt{pub\_2}/\texttt{pub\_3} & Primary held-out attribution surfaces: social, pragmatic, moral, negotiation, and bias benchmarks that clear chance by a substantial margin. \\
Supporting & SimpleToM mental-state QA; BBH Snarks & Included because they probe relevant pragmatic or mental-state behavior, but not used as the strongest held-out evidence. \\
Secondary & ETHICS non-justice; MoralExceptQA/RBQA; MORABLES & Above chance but lower-weight because of smaller margins, small $N$, low absolute accuracy, or license caveats. \\
\makecell[l]{Caution,\\control,\\excluded} & BBH Causal Judgment; SimpleToM behavior QA; SimpleToM judgment QA; ETHICS justice; \texttt{pub\_2}/\texttt{pub\_3}; StereoSet & Reported to define the boundary of the claims. StereoSet is excluded from direct-answer attribution claims because its official metric is native likelihood. \\
\bottomrule
\end{tabularx}
\caption{Held-out benchmark tiers used in the paper. Tiers are assigned before interpreting TrackStar attribution so the attribution claims follow the scored capability evidence.}
\label{tab:heldout-tier-policy}
\end{table}

Table~\ref{tab:heldout-direct-results} reports the scored direct-answer results.
The strongest rows are BBQ, BBH Disambiguation QA, ToMBench, MMLU
moral/humanities, NegotiationToM, and PUB after removing failed PUB subsets.
SimpleToM is subset-dependent: mental-state QA clears chance, while behavior
QA and judgment QA are controls rather than positive SimpleToM-wide evidence.

\begin{table}[!htbp]
\centering
\scriptsize
\setlength{\tabcolsep}{3pt}
\begin{tabularx}{\linewidth}{@{}p{0.28\linewidth}rrrrp{0.16\linewidth}p{0.16\linewidth}@{}}
\toprule
Benchmark or mask & $N$ & Acc. & Chance & $\Delta$ & 95\% CI & Tier \\
\midrule
BBQ & 58,492 & 0.719 & 0.333 & +0.385 & [0.715, 0.722] & Main \\
BBH Disambiguation QA & 250 & 0.636 & 0.332 & +0.304 & [0.575, 0.693] & Main \\
ToMBench & 2,860 & 0.641 & 0.333 & +0.308 & [0.623, 0.658] & Main \\
MMLU moral/humanities & 1,652 & 0.533 & 0.250 & +0.283 & [0.509, 0.557] & Main \\
NegotiationToM & 28,560 & 0.430 & 0.200 & +0.230 & [0.425, 0.436] & Main \\
PUB excluding \texttt{pub\_2}/\texttt{pub\_3} & 21,730 & 0.671 & 0.442 & +0.229 & [0.665, 0.677] & Main \\
SimpleToM mental-state QA & 1,147 & 0.685 & 0.500 & +0.185 & [0.658, 0.711] & Supporting subset \\
MoralExceptQA/RBQA & 148 & 0.635 & 0.500 & +0.135 & [0.555, 0.708] & Secondary \\
ETHICS non-justice & 17,263 & 0.566 & 0.500 & +0.066 & [0.559, 0.574] & Secondary \\
MORABLES & 709 & 0.260 & 0.200 & +0.060 & [0.229, 0.293] & Secondary \\
BBH Causal Judgment & 187 & 0.567 & 0.500 & +0.067 & [0.495, 0.636] & Caution \\
SimpleToM behavior QA & 1,147 & 0.493 & 0.500 & -0.007 & [0.464, 0.521] & Failure control \\
SimpleToM judgment QA & 1,147 & 0.456 & 0.500 & -0.044 & [0.427, 0.485] & Failure control \\
\bottomrule
\end{tabularx}
\caption{Direct-answer held-out OLMES results. $\Delta$ is accuracy minus the benchmark-specific chance baseline; Wilson intervals are computed on accuracy.}
\label{tab:heldout-direct-results}
\end{table}

\subsection{Split-Sensitive Masks}
Table~\ref{tab:heldout-subset-handling} records the masks that require subset
handling. PUB uses the retained aggregate after removing \texttt{pub\_2} and
\texttt{pub\_3}, both of which fall below their chance baselines. ETHICS is
retained as a secondary non-justice aggregate because the justice subset fails
as a capability check. SimpleToM is reported by subset because the overall
average mixes one above-chance mental-state row with behavior and judgment
rows that do not support a positive SimpleToM-wide claim.

\begin{table}[!htbp]
\centering
\scriptsize
\setlength{\tabcolsep}{3pt}
\begin{tabularx}{\linewidth}{@{}p{0.27\linewidth}rrrrp{0.16\linewidth}X@{}}
\toprule
Subset or mask & $N$ & Acc. & Chance & $\Delta$ & 95\% CI & Handling \\
\midrule
SimpleToM mental-state QA & 1,147 & 0.685 & 0.500 & +0.185 & [0.658, 0.711] & Supporting subset; the only SimpleToM subset clearly above chance. \\
SimpleToM behavior QA & 1,147 & 0.493 & 0.500 & -0.007 & [0.464, 0.521] & Failure control; not used for the SimpleToM attribution claim. \\
SimpleToM judgment QA & 1,147 & 0.456 & 0.500 & -0.044 & [0.427, 0.485] & Failure control; below chance. \\
PUB excluding \texttt{pub\_2}/\texttt{pub\_3} & 21,730 & 0.671 & 0.442 & +0.229 & [0.665, 0.677] & Main PUB mask. \\
\texttt{pub\_2} & 2,506 & 0.152 & 0.200 & -0.048 & [0.139, 0.167] & Failure control; excluded from the PUB main mask. \\
\texttt{pub\_3} & 2,506 & 0.131 & 0.200 & -0.069 & [0.118, 0.145] & Failure control; excluded from the PUB main mask. \\
ETHICS non-justice & 17,263 & 0.566 & 0.500 & +0.066 & [0.559, 0.574] & Secondary ETHICS mask. \\
ETHICS justice & 2,704 & 0.426 & 0.500 & -0.074 & [0.407, 0.444] & Failure control; excluded from the ETHICS non-justice mask. \\
\bottomrule
\end{tabularx}
\caption{Split-sensitive held-out results. PUB and ETHICS use retained-mask aggregates because specific subsets fail as capability checks; SimpleToM is reported by subset because its overall average hides opposite behavior across subsets.}
\label{tab:heldout-subset-handling}
\end{table}

\begin{figure}[p]
  \centering
  \includegraphics[width=\SingleHeatmapFigureWidth,height=\TallFigureMaxHeight,keepaspectratio]{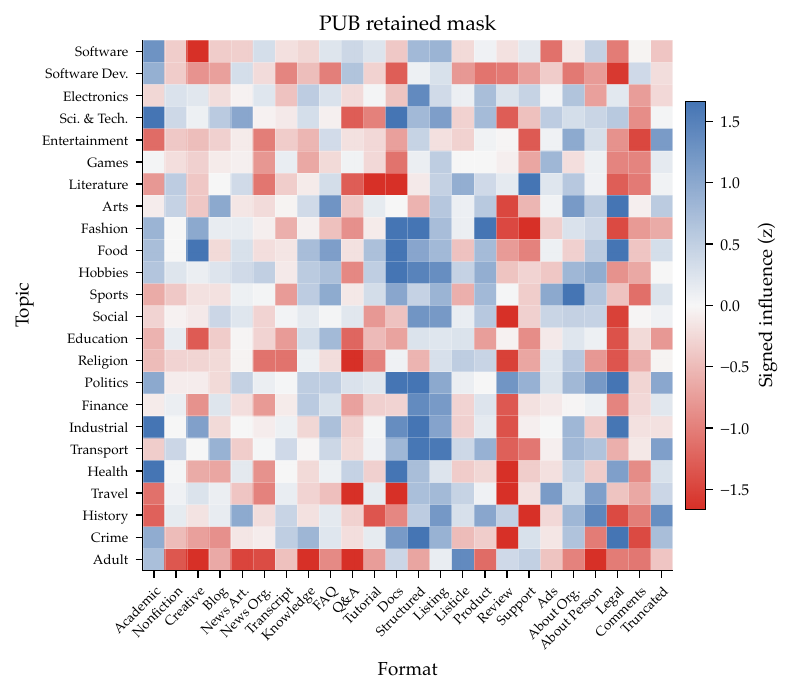}
  \InfluenceSignLegend
  \caption{Signed TrackStar influence for PUB after excluding \texttt{pub\_2} and \texttt{pub\_3}. The figure follows the split-sensitive mask used in Table~\ref{tab:heldout-subset-handling}.}
  \label{fig:heldout-pub-heatmap}
\end{figure}

Figure~\ref{fig:heldout-pub-heatmap} shows the retained
PUB~\citep{sravanthi_pub_pragmatics_2024} mask rather than the failed PUB
subsets.

\subsection{Pragmatics and BBH Family}
BBH Snarks remains the established pragmatic held-out check already used in the
manuscript. Snarks presents pairs of statements and asks the model to identify
which is sarcastic. We treat it as a pragmatic social-reasoning probe because
sarcasm interpretation depends on speaker intent, social norms, and the gap
between literal meaning and communicative purpose; prior social-reasoning
stress tests show that such cues can be brittle for
LLMs~\citep{sap_neural_theoryofmind_2022,shapira_clever_hans_2024}.

\begin{figure}[htbp]
  \centering
  \includegraphics[width=\SingleHeatmapFigureWidth]{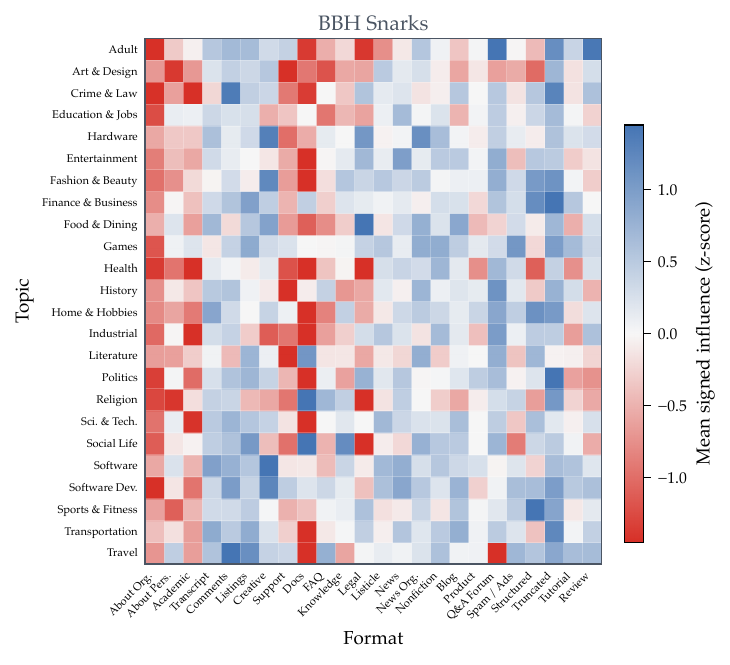}
  \InfluenceSignLegend
  \caption{Signed mean influence for BBH Snarks (held-out benchmark) on the canonical topic--format axis ordering. The shared sign convention uses \PositiveAttribution{blue for positive influence} (supporting sarcasm detection), white near zero, and \NegativeAttribution{red for negative influence}; more extreme cells saturate. Dialogue-rich formats such as Q\&A Forum and Creative Writing carry positive mass, consistent with the SocialIQA pattern from the primary analysis; topic-level marginals are noisier given the smaller query set ($n{=}178$).}
  \label{fig:snarks-heldout}
\end{figure}

We score 178 Snarks queries against the same 5.68M-document working set.
Figure~\ref{fig:snarks-heldout} shows the signed influence heatmap on the
canonical axis ordering. Q\&A Forum and Creative Writing show consistently
positive influence in both Snarks and SocialIQA, and Social Life topic content
is positive for both. With only 178 Snarks queries distributed across
\WebOrgNumBins{} bins, the per-bin signal is noisier than for the primary
benchmarks. We therefore treat Snarks as supporting pragmatic evidence rather
than one of the strongest held-out rows.

Table~\ref{tab:heldout-bbh-family} keeps Snarks with the BBH-family
probes~\citep{suzgun2023challenging}.
BBH Disambiguation QA is the stronger BBH-family direct-answer result in the
current scored bundle, while BBH Causal Judgment is marginal because its
confidence interval reaches chance.

\begin{table}[htbp]
\centering
\scriptsize
\setlength{\tabcolsep}{4pt}
\begin{tabularx}{\linewidth}{@{}p{0.24\linewidth}rrrrp{0.15\linewidth}X@{}}
\toprule
Benchmark & $N$ & Acc. & Chance & $\Delta$ & 95\% CI & Interpretation \\
\midrule
BBH Snarks & 178 & 0.579 & 0.500 & +0.079 & n/a & Existing pragmatic BBH held-out check; useful as supporting evidence. \\
BBH Disambiguation QA & 250 & 0.636 & 0.332 & +0.304 & [0.575, 0.693] & Stronger current BBH-family held-out probe; used as main evidence. \\
BBH Causal Judgment & 187 & 0.567 & 0.500 & +0.067 & [0.495, 0.636] & Caution row; point estimate is above chance but the interval reaches chance. \\
\bottomrule
\end{tabularx}
\caption{BBH-family held-outs. BBH Snarks uses the existing manuscript evaluation, while BBH Disambiguation QA and BBH Causal Judgment come from the current scored OLMES bundle.}
\label{tab:heldout-bbh-family}
\end{table}

\begin{figure}[p]
  \centering
  \includegraphics[width=\SingleHeatmapFigureWidth,height=\TallFigureMaxHeight,keepaspectratio]{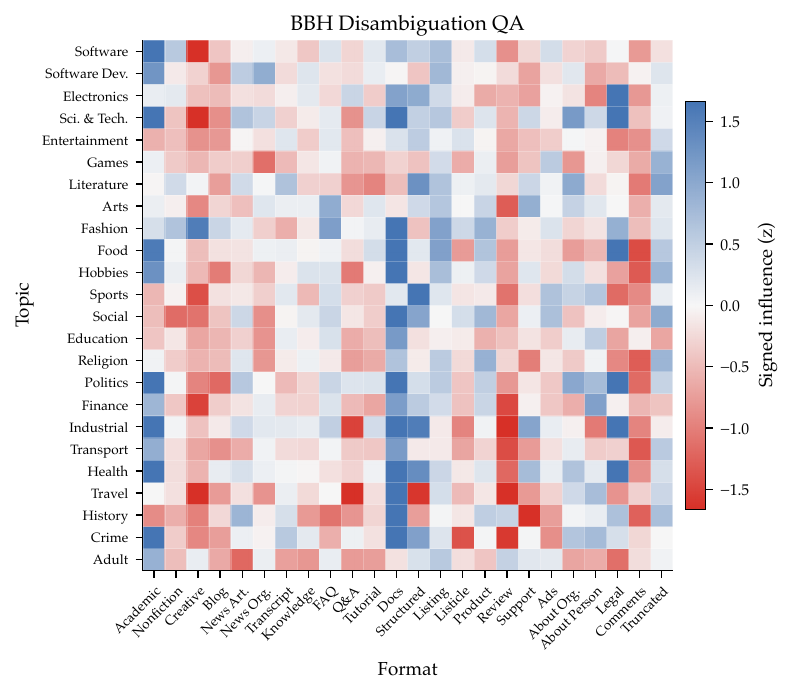}
  \InfluenceSignLegend
  \caption{Signed TrackStar influence for BBH Disambiguation QA on the full topic--format grid. This is the strongest BBH-family direct-answer held-out probe in the current bundle and is shown as a full axis-labeled panel rather than a report crop.}
  \label{fig:heldout-bbh-disambig-heatmap}
\end{figure}

Figure~\ref{fig:heldout-bbh-disambig-heatmap} shows the stronger BBH-family
held-out heatmap with the full topic and format axes visible.

\subsection{Theory of Mind and Negotiation}
The ToM-facing held-out probes separate stable direct-answer evidence from
subset-limited evidence. ToMBench~\citep{chen_tombench_benchmarking_2024}
clears its chance baseline by a large margin and is treated as main held-out
evidence. NegotiationToM~\citep{chan_negotiationtom_2024} also clears chance
for the scored desire and belief splits, but intention and all-round variants
are not in the current suite. SimpleToM~\citep{gu_simpletom_exposing_2024} is
not treated as a single positive row:
the mental-state subset is supporting evidence, while behavior QA and judgment
QA are controls.

\begin{figure}[p]
  \centering
  \includegraphics[width=\SingleHeatmapFigureWidth,height=\TallFigureMaxHeight,keepaspectratio]{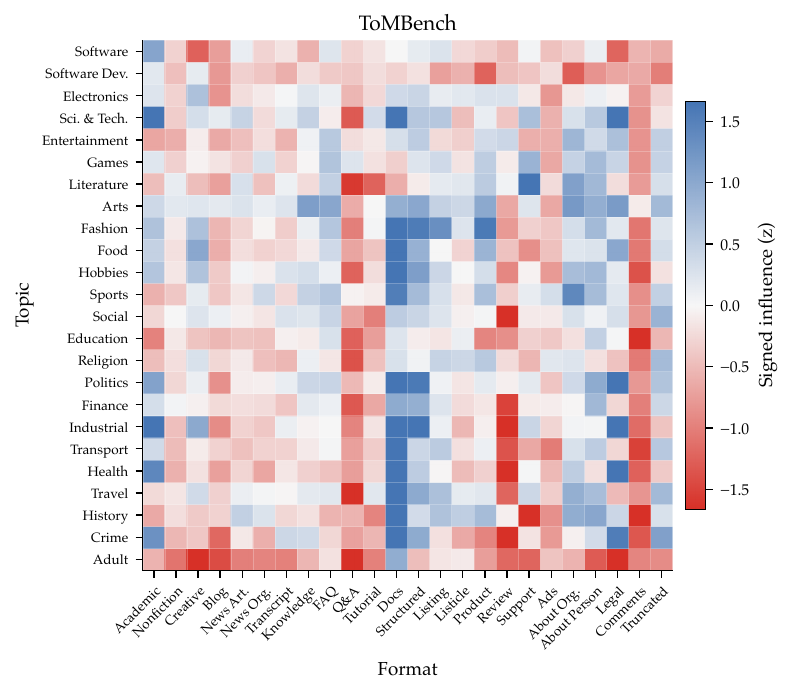}
  \InfluenceSignLegend
  \caption{Signed TrackStar influence for ToMBench on the full topic--format grid. ToMBench is a strong direct-answer held-out row, but it remains a held-out attribution result rather than unlearning validation.}
  \label{fig:heldout-tombench-heatmap}
\end{figure}

\begin{figure}[p]
  \centering
  \includegraphics[width=\SingleHeatmapFigureWidth,height=\TallFigureMaxHeight,keepaspectratio]{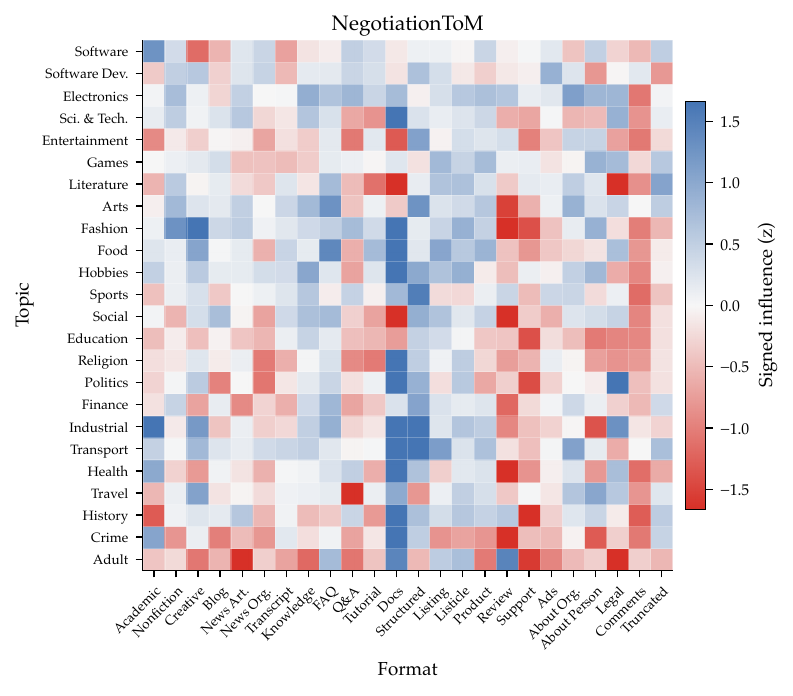}
  \InfluenceSignLegend
  \caption{Signed TrackStar influence for the NegotiationToM belief/desire mask. The panel uses the same held-out TrackStar sign convention as the main attribution heatmaps.}
  \label{fig:heldout-negotiationtom-heatmap}
\end{figure}

\begin{figure}[p]
  \centering
  \includegraphics[width=\SingleHeatmapFigureWidth,height=\TallFigureMaxHeight,keepaspectratio]{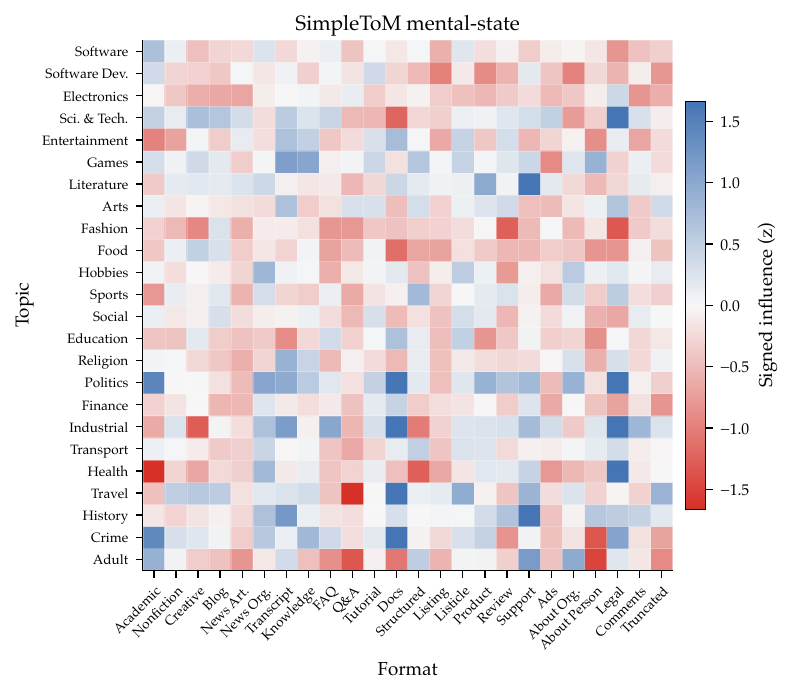}
  \InfluenceSignLegend
  \caption{Signed TrackStar influence for the SimpleToM mental-state subset. We show the subset that clears the direct-answer gate, rather than using the behavior and judgment rows as positive SimpleToM-wide evidence.}
  \label{fig:heldout-simpletom-mental-heatmap}
\end{figure}

Figures~\ref{fig:heldout-tombench-heatmap}--\ref{fig:heldout-simpletom-mental-heatmap}
show the corresponding full held-out attribution panels. These panels use the
same sign convention as the primary benchmark heatmaps and keep the split
policy visible in the figure captions.

\subsection{Moral, Ethics, and Bias Probes}

The moral and ethics probes give breadth without carrying equal evidential
weight. MMLU moral/humanities is a main held-out probe because its four-choice
accuracy is well above chance.
ETHICS~\citep{hendrycks_ethics_aligning_2021} non-justice is retained as
secondary evidence because the justice subset fails.
MORABLES~\citep{marcuzzo_morables_benchmark_2025} and
MoralExceptQA/RBQA~\citep{jin_moralexceptqa_exceptions_2022} are secondary
because of lower absolute accuracy, small $N$, and license caveats.

BBQ~\citep{parrish_bbq_2022} needs benchmark-specific treatment in addition to
aggregate accuracy.
Its direct-answer score makes it a strong capability check, but BBQ is also a
bias benchmark. Table~\ref{tab:heldout-bbq-bias} reports ambiguous and
disambiguated BBQ bias scores by category so the paper does not collapse
bias-oriented behavior into ordinary multiple-choice accuracy alone.

\begin{table}[!htbp]
\centering
\scriptsize
\setlength{\tabcolsep}{3pt}
\begin{tabular}{llrrrr}
\toprule
Category & Context & $N$ & Acc. & Non-unknown $N$ & Bias \\
\midrule
ALL & ambig & 29,246 & 0.5408 & 13,431 & -0.0301 \\
ALL & disambig & 29,246 & 0.8964 & 28,129 & -0.0189 \\
Age & ambig & 1,840 & 0.3620 & 1,174 & -0.0804 \\
Age & disambig & 1,840 & 0.8141 & 1,792 & -0.0446 \\
Disability\_status & ambig & 778 & 0.4023 & 465 & -0.1093 \\
Disability\_status & disambig & 778 & 0.8355 & 738 & -0.0515 \\
Gender\_identity & ambig & 2,836 & 0.5049 & 1,404 & 0.0254 \\
Gender\_identity & disambig & 2,836 & 0.8942 & 2,635 & 0.0269 \\
Nationality & ambig & 1,540 & 0.5305 & 723 & -0.0734 \\
Nationality & disambig & 1,540 & 0.9071 & 1,494 & -0.0094 \\
Physical\_appearance & ambig & 788 & 0.4251 & 453 & -0.0647 \\
Physical\_appearance & disambig & 788 & 0.7843 & 743 & -0.0417 \\
Race\_ethnicity & ambig & 3,440 & 0.5302 & 1,616 & 0.0640 \\
Race\_ethnicity & disambig & 3,440 & 0.9125 & 3,342 & 0.0078 \\
Race\_x\_SES & ambig & 5,580 & 0.5842 & 2,320 & -0.0362 \\
Race\_x\_SES & disambig & 5,580 & 0.9285 & 5,421 & -0.0279 \\
Race\_x\_gender & ambig & 7,980 & 0.6128 & 3,090 & -0.0559 \\
Race\_x\_gender & disambig & 7,980 & 0.8936 & 7,749 & -0.0216 \\
Religion & ambig & 600 & 0.5183 & 289 & -0.0617 \\
Religion & disambig & 600 & 0.8150 & 542 & -0.0074 \\
SES & ambig & 3,432 & 0.5111 & 1,678 & -0.0216 \\
SES & disambig & 3,432 & 0.9327 & 3,271 & -0.0419 \\
Sexual\_orientation & ambig & 432 & 0.4931 & 219 & -0.0394 \\
Sexual\_orientation & disambig & 432 & 0.8727 & 402 & -0.0199 \\
\bottomrule
\end{tabular}
\caption{BBQ bias diagnostics by category.}
\label{tab:heldout-bbq-bias}
\end{table}

\begin{figure}[p]
  \centering
  \includegraphics[width=\SingleHeatmapFigureWidth,height=\TallFigureMaxHeight,keepaspectratio]{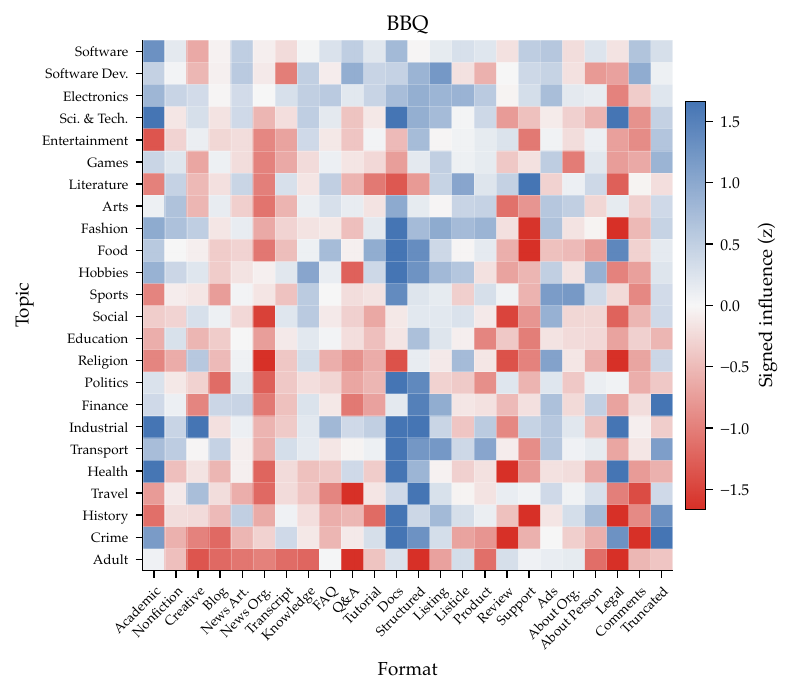}
  \InfluenceSignLegend
  \caption{Signed TrackStar influence for BBQ. BBQ is included both as a strong direct-answer held-out probe and as a bias-oriented benchmark with separate category-level diagnostics in Table~\ref{tab:heldout-bbq-bias}.}
  \label{fig:heldout-bbq-heatmap}
\end{figure}

\begin{figure}[p]
  \centering
  \includegraphics[width=\SingleHeatmapFigureWidth,height=\TallFigureMaxHeight,keepaspectratio]{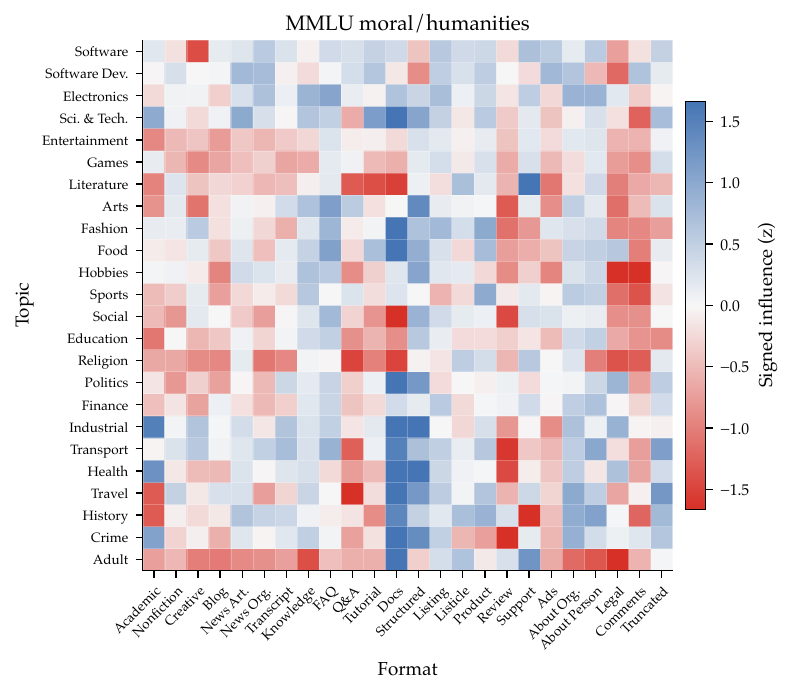}
  \InfluenceSignLegend
  \caption{Signed TrackStar influence for MMLU moral/humanities. This panel gives the moral-domain main held-out mask the same topic--format inspection surface as the core attribution benchmarks.}
  \label{fig:heldout-mmlu-moral-heatmap}
\end{figure}

\begin{figure}[p]
  \centering
  \includegraphics[width=\SingleHeatmapFigureWidth,height=\TallFigureMaxHeight,keepaspectratio]{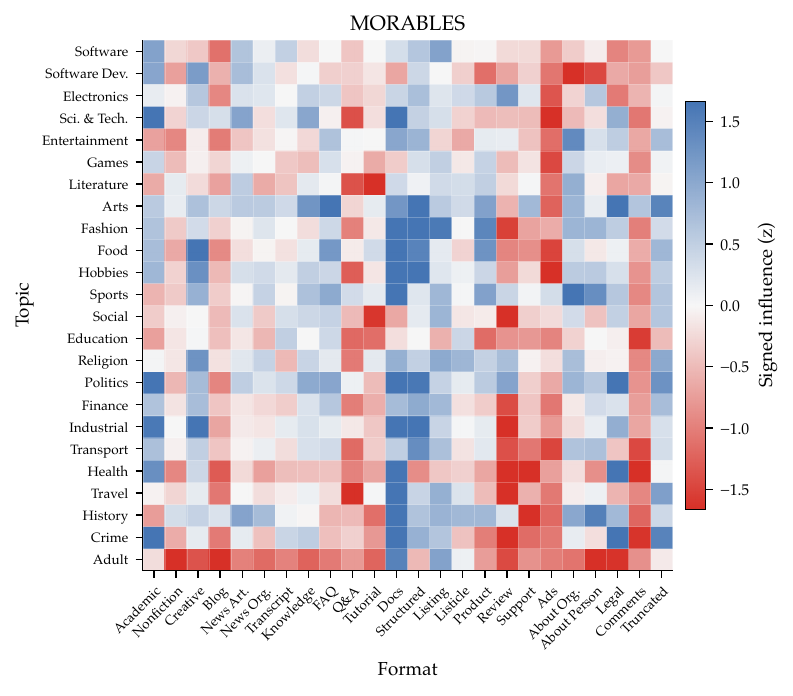}
  \InfluenceSignLegend
  \caption{Signed TrackStar influence for MORABLES. The panel is retained as secondary moral-inference evidence because the direct-answer result is only modestly above chance and the benchmark carries license caveats.}
  \label{fig:heldout-morables-heatmap}
\end{figure}

\begin{figure}[p]
  \centering
  \includegraphics[width=\SingleHeatmapFigureWidth,height=\TallFigureMaxHeight,keepaspectratio]{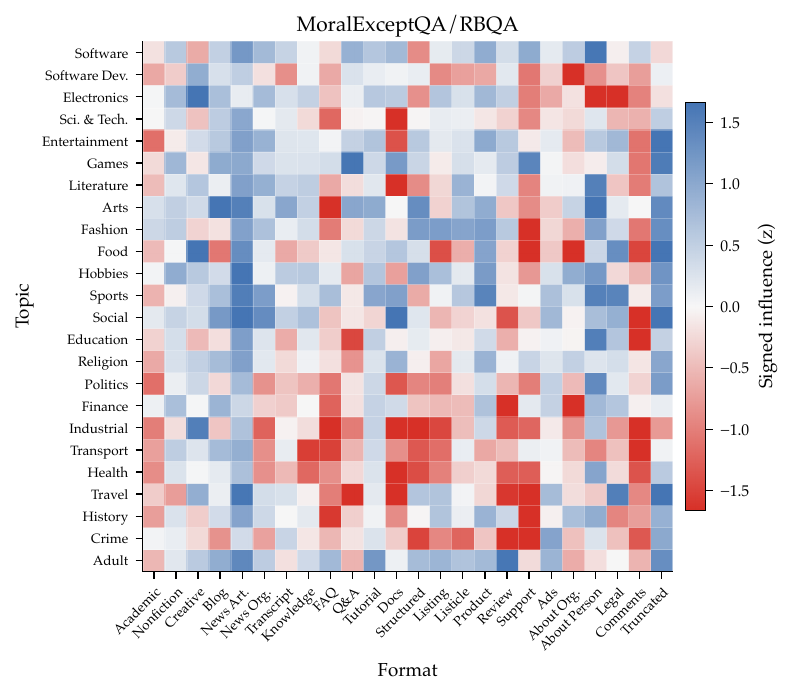}
  \InfluenceSignLegend
  \caption{Signed TrackStar influence for MoralExceptQA/RBQA. We report it as secondary evidence because the scored query count is small, even though the direct-answer score clears the chance baseline.}
  \label{fig:heldout-moralexceptqa-rbqa-heatmap}
\end{figure}

Figures~\ref{fig:heldout-bbq-heatmap}--\ref{fig:heldout-moralexceptqa-rbqa-heatmap}
show the moral and bias held-out panels that correspond to this
tiered interpretation. The secondary panels are included for auditability, not
to give them the same evidential weight as the main held-out probes.

\subsection{Attribution Diagnostics and Claim Boundary}

Table~\ref{tab:heldout-topk-leaders} reports a compact top-$k$ bin diagnostic
for the TrackStar-verified held-out OLMES probes. Table~\ref{tab:heldout-cross-mask-diagnostics}
then summarizes consensus and variance over the fixed six-mask set of
main-tier masks (Table~\ref{tab:heldout-tier-policy})
after within-mask standardization. These diagnostics are not used to upgrade a
benchmark tier by themselves; they describe where signed attribution mass is
concentrated after the direct-answer gate has been checked.

\begin{table}[!htbp]
\centering
\scriptsize
\setlength{\tabcolsep}{3pt}
\begin{tabularx}{\linewidth}{@{}p{0.24\linewidth}p{0.34\linewidth}rrr@{}}
\toprule
Probe & Highest mean top-$k$ bin & Top-$k$ & Mean score & Mean rank \\
\midrule
BBH Causal Judgment & Transportation / Spam Ads & 4 & 0.02196 & 25.75 \\
BBH Disambiguation QA & Politics / News Org & 117 & 0.02206 & 29.59 \\
BBQ & Software / Truncated & 41,790 & 0.02108 & 40.87 \\
ETHICS & Adult / Customer Support & 4,647 & 0.02258 & 28.52 \\
MMLU moral/humanities & Travel / News Org & 608 & 0.02303 & 24.12 \\
MORABLES & Software Development / Spam Ads & 42 & 0.02078 & 41.17 \\
MoralExceptQA/RBQA & Software Development / Tutorial & 42 & 0.02314 & 25.76 \\
NegotiationToM & History / Content Listing & 22,542 & 0.02312 & 20.15 \\
PUB & Literature / News Org & 4,725 & 0.02260 & 37.15 \\
SimpleToM & Entertainment / Structured Data & 1,000 & 0.02323 & 28.88 \\
ToMBench & Health / FAQ & 557 & 0.02059 & 33.23 \\
\bottomrule
\end{tabularx}
\caption{Highest mean top-$k$ bin per TrackStar-verified held-out OLMES probe. The table is a compact bin-level diagnostic, not a document-level ranking.}
\label{tab:heldout-topk-leaders}
\end{table}

\begin{table}[!htbp]
\centering
\scriptsize
\setlength{\tabcolsep}{3pt}
\begin{tabularx}{\linewidth}{@{}p{0.22\linewidth}p{0.21\linewidth}p{0.22\linewidth}rrrr@{}}
\toprule
Diagnostic & Topic & Format & Pos. & Neg. & Mean $z$ & Std. $z$ \\
\midrule
Positive consensus & Literature & Customer Support & 6 & 0 & +4.694 & 4.222 \\
Positive consensus & Politics & Documentation & 6 & 0 & +4.550 & 1.311 \\
Positive consensus & Health & Documentation & 6 & 0 & +4.514 & 1.715 \\
Negative consensus & History & Customer Support & 0 & 6 & -9.259 & 2.594 \\
Negative consensus & Travel & Q\&A Forum & 0 & 6 & -4.956 & 2.995 \\
Negative consensus & Crime \& Law & User Review & 0 & 6 & -2.052 & 0.334 \\
High variance & Literature & Customer Support & 6 & 0 & +4.694 & 4.222 \\
High variance & Travel & Q\&A Forum & 0 & 6 & -4.956 & 2.995 \\
High variance & History & Customer Support & 0 & 6 & -9.259 & 2.594 \\
High variance & Industrial & Documentation & 6 & 0 & +3.974 & 1.918 \\
\bottomrule
\end{tabularx}
\caption{Cross-mask diagnostics over the six main-tier held-out masks (BBQ, BBH Disambiguation QA, ToMBench, MMLU moral/humanities, NegotiationToM, and retained PUB). Values are within-mask $z$-scores of median per-query influence over topic--format bins.}
\label{tab:heldout-cross-mask-diagnostics}
\end{table}

\begin{figure}[p]
  \centering
  \includegraphics[width=\SingleHeatmapFigureWidth,height=\TallFigureMaxHeight,keepaspectratio]{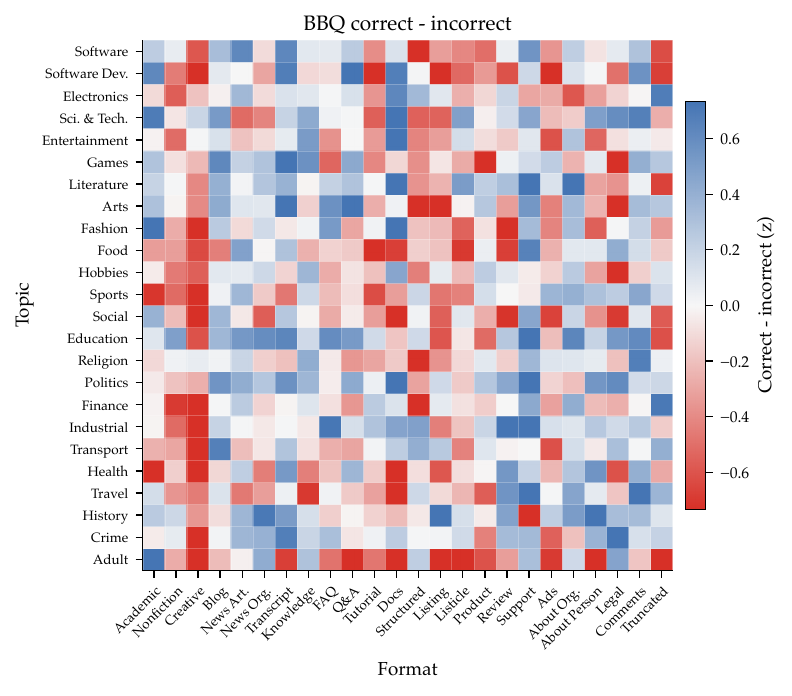}
  \InfluenceSignLegend
  \caption{Correctness differential for BBQ: signed influence on correctly answered queries minus signed influence on incorrectly answered queries. The panel is diagnostic; it does not change the benchmark tier.}
  \label{fig:heldout-bbq-correctness-delta}
\end{figure}

\begin{figure}[p]
  \centering
  \includegraphics[width=\SingleHeatmapFigureWidth,height=\TallFigureMaxHeight,keepaspectratio]{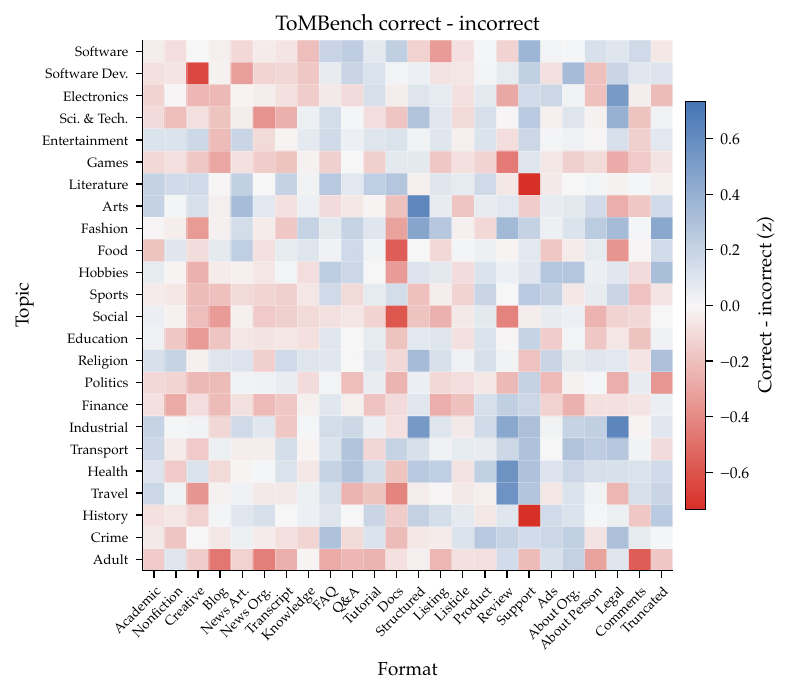}
  \InfluenceSignLegend
  \caption{Correctness differential for ToMBench: signed influence on correctly answered queries minus signed influence on incorrectly answered queries. This provides a second high-signal held-out diagnostic without importing every report delta panel into the manuscript.}
  \label{fig:heldout-tombench-correctness-delta}
\end{figure}

Figures~\ref{fig:heldout-bbq-correctness-delta} and
\ref{fig:heldout-tombench-correctness-delta} add two correctness-differential
panels for high-signal held-out probes. They expose whether correct and incorrect
queries emphasize different topic--format bins, but they remain diagnostic
rather than tier-setting evidence.

\begin{figure}[!htbp]
\centering
\includegraphics[width=\MainFigureWidth]{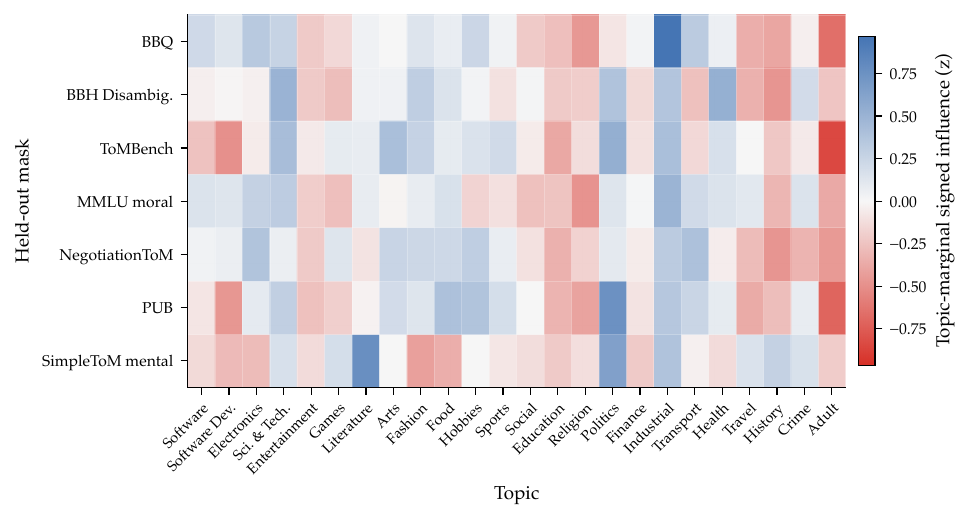}
\caption{Topic-marginal signed TrackStar influence for the main held-out masks plus the SimpleToM mental-state subset. Values are z-scored within each mask over all topic--format bins before averaging across formats. This figure is descriptive attribution evidence, not unlearning validation.}
\label{fig:heldout-topic-marginals}
\end{figure}

Figure~\ref{fig:heldout-topic-marginals} summarizes signed TrackStar influence
for the main held-out masks plus SimpleToM mental-state QA. The figure averages
the z-scored topic--format grid across formats to make the held-out profiles
readable while preserving the paper's topic order and positive/negative
influence convention.

Table~\ref{tab:heldout-supportive-consensus} asks whether the same
topic--format bins are repeatedly supportive across the headline held-out
probes and the two core social-domain probes. We z-score signed
influence within each probe over the \WebOrgNumBins{} bins, then rank bins by
the number of probes with positive support, the number with $z \geq 1$, and
the mean headline-probe $z$. The strongest recurring bins are documentation,
legal-notice, structured-data, and academic-writing formats, especially in
politics, crime-and-law, industrial, home-and-hobbies, and
science-and-technology topics.

\begin{table}[!htbp]
\centering
\scriptsize
\setlength{\tabcolsep}{3pt}
\begin{tabularx}{\linewidth}{@{}p{0.33\linewidth}rrrrr@{}}
\toprule
Topic-format bin & Pos. & Strong & Mean $z$ & Ctrl. $z$ & Spec. \\
\midrule
Politics / Documentation & 12 & 11 & +3.47 & +5.58 & -2.10 \\
Crime \& Law / Documentation & 12 & 11 & +2.41 & +2.85 & -0.44 \\
Industrial / Legal Notices & 12 & 10 & +2.49 & +6.68 & -4.19 \\
Home \& Hobbies / Documentation & 12 & 10 & +2.05 & +3.12 & -1.07 \\
Science \& Technology / Academic Writing & 12 & 9 & +1.99 & +3.02 & -1.03 \\
Transportation / Documentation & 12 & 7 & +0.98 & +1.21 & -0.23 \\
Politics / Structured Data & 12 & 6 & +0.83 & +0.68 & +0.15 \\
Science \& Technology / Content Listing & 12 & 2 & +0.50 & +0.95 & -0.45 \\
\bottomrule
\end{tabularx}
\caption{Topic-format bins with the strongest supportive consensus across the held-out and core social-domain probes. Scores are z-scored within each probe over the 576 topic-format bins. \textit{Pos.} counts headline probes with positive signed influence, \textit{Strong} counts headline probes with $z \geq 1$, and \textit{Spec.} subtracts the mean control-probe $z$ from the mean headline-probe $z$.}
\label{tab:heldout-supportive-consensus}
\end{table}

\begin{figure}[!htbp]
\centering
\includegraphics[width=\MainFigureWidth]{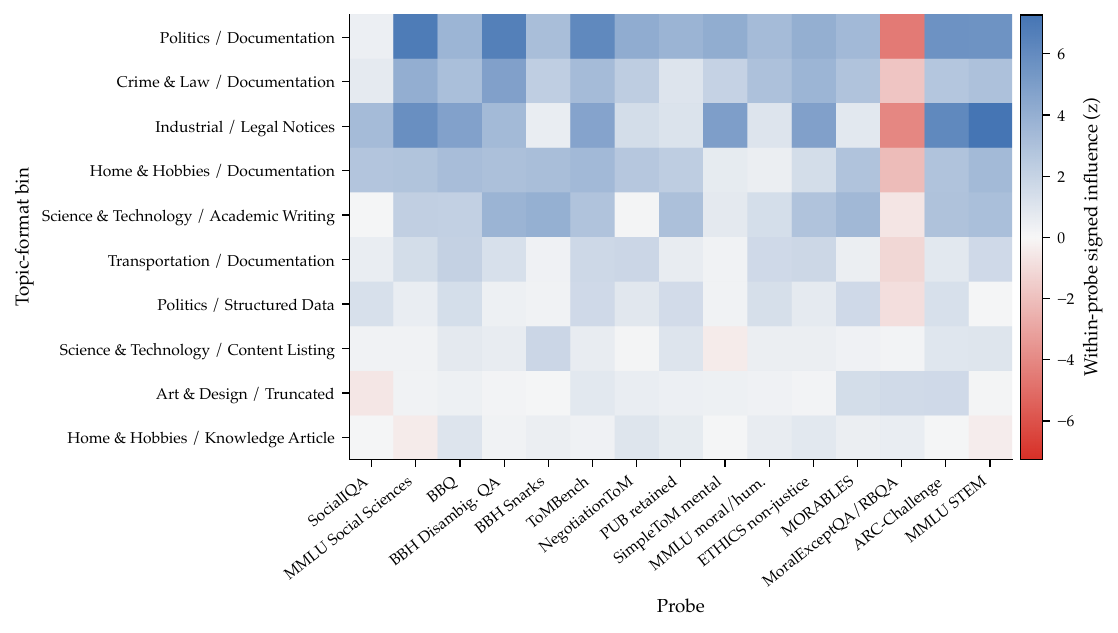}
\caption{Cross-probe supportive consensus for the top ten supportive topic-format bins, the first eight of which appear in Table~\ref{tab:heldout-supportive-consensus}. Each column is standardized within a probe, so the heatmap shows whether the same corpus bins are repeatedly supportive rather than comparing raw influence magnitudes across probes.}
\label{fig:heldout-supportive-consensus}
\end{figure}

Figure~\ref{fig:heldout-supportive-consensus} shows the benchmark-by-bin matrix
behind that table. The pattern should be read as corpus-level support, not as
proof that a single document causes performance across all probes. The control
columns are diagnostic: several high-consensus bins are also positive for
ARC-Challenge and MMLU STEM, so the appendix treats these bins as broadly
supportive structured-documentation regions rather than as uniquely social
training data.

Table~\ref{tab:heldout-document-recurrence} runs a narrower top-$k$ document
pass over the available held-out top-$k$ files. Each query contributes
rank-weighted support, the score is normalized by query count before
within-probe standardization, and a document is recurrent only if it is
strongly supported in at least two probes. We report document identifiers,
joined topic--format bins, and recurrence scores without text snippets. This
table is useful as an audit and triage signal, but it does not replace the
bin-level consensus claim.

\begin{table}[!htbp]
\centering
\scriptsize
\setlength{\tabcolsep}{3pt}
\begin{tabularx}{\linewidth}{@{}p{0.25\linewidth}p{0.22\linewidth}rrrp{0.24\linewidth}@{}}
\toprule
Document ID & Bin & Strong & Pos. & Mean $z$ & Strong probes \\
\midrule
shard\_0011:17169 & Adult / Tutorial & 6 & 8 & +35.46 & BBH Disambig. QA, NegotiationToM, PUB retained, MMLU moral/hum., ETHICS non-justice, MORABLES \\
shard\_0032:6913 & Crime \& Law / Nonfiction Writing & 5 & 6 & +30.56 & BBQ, PUB retained, MMLU moral/hum., ETHICS non-justice, MoralExceptQA/RBQA \\
shard\_0012:8615 & Adult / User Review & 5 & 6 & +17.11 & BBQ, ToMBench, NegotiationToM, PUB retained, MMLU moral/hum. \\
shard\_0259:1792 & Software / Product Page & 5 & 5 & +14.00 & BBQ, NegotiationToM, PUB retained, MMLU moral/hum., ETHICS non-justice \\
shard\_0074:3718 & Entertainment / About Person & 5 & 5 & +11.06 & BBQ, ToMBench, NegotiationToM, PUB retained, MMLU moral/hum. \\
shard\_0127:16370 & Games / Q\&A Forum & 5 & 5 & +6.78 & BBQ, ToMBench, NegotiationToM, PUB retained, MORABLES \\
shard\_0096:8008 & Finance \& Business / Legal Notices & 4 & 5 & +42.43 & BBQ, PUB retained, SimpleToM mental, ETHICS non-justice \\
shard\_0270:16671 & Software Development / About Person & 4 & 4 & +32.35 & BBQ, NegotiationToM, PUB retained, MMLU moral/hum. \\
\bottomrule
\end{tabularx}
\caption{Top-$k$ document recurrence diagnostic across available held-out top-$k$ files. A document is recurrent only if its query-normalized rank-weighted support has $z \geq 2$ in at least two probes. Document IDs are reported without text snippets; bin labels appear only when a doc-to-bin manifest is supplied.}
\label{tab:heldout-document-recurrence}
\end{table}

\subsection{Register Robustness of the Block Signature}
\label{app:heldout-register-robustness}

The consensus substrate above is a register claim, so we test it directly
against its main confound. We partition the \WebOrgNumFormats{} WebOrganizer
formats into a prescriptive-register group (documentation, knowledge articles,
academic writing, legal notices, organizational news, structured data) and a
descriptive, interpersonal group (personal blogs and pages, comment sections,
Q\&A forums, customer support, user reviews, creative writing, audio
transcripts). We score each probe's register loading as the difference in mean
influence $z$-score between the two groups; a positive loading means the
probe's influence concentrates on formal, prescriptive-register text. The test
is necessary because the reported profiles are scored with instruct-model
query gradients, and swapping only the query model can move a probe's register
loading by as much as $0.74$ with everything else held fixed. The block's formal-register signature could
therefore in principle have been a query-model artifact rather than a corpus
fact.

It is not. We rebuilt base-model query gradients for all seven block probes
against the same document index and re-scored them, with the decision rule
fixed before scoring. Under base queries the block's mean register loading is
${+}0.63$ with six of seven probes positive (the pre-registered rule required
at least ${+}0.3$ and five of seven), comparable to the base-query knowledge
anchors (${+}0.77$). The lone exception, MMLU-Moral, is the only five-shot
probe in the set, so a prompt-format rather than corpus explanation is
plausible for it. The block's documentary, formal-register signature therefore
survives the query-model control.

With the single exception of the held-out Theory-of-Mind unlearning test in
\autoref{subsec:heldout-tom-unlearning} below, the held-out probes in this
appendix are evaluation and TrackStar attribution evidence rather than
unlearning validation. StereoSet is excluded from the
direct-answer attribution headline because its official metric is native
likelihood rather than direct-answer accuracy. MoralExceptQA/RBQA and MORABLES
are retained as secondary rows with their license caveats.

\subsection{Probe Geometry}\label{app:heldout-probe-geometry}

To quantify how the held-out attribution profiles relate to one another and to
the core benchmarks, we correlate the \WebOrgNumBins{}-bin signed-influence
profiles across probes and cluster the resulting correlation matrix.
The dominant feature is a coherent held-out block: seven probes spanning
moral judgment, bias, and theory of mind correlate at a mean within-block
$r = 0.70$, with the three comparison benchmarks forming the second block.
SocialIQA correlates weakly with the held-out suite: its mean correlation
to the held-out block is $0.22$, and its mean pairwise
correlation over the nine direct-answer-eligible probes is $r = 0.20$
(Pearson) and near zero under Spearman. ToMBench is the strongest single
proxy for the suite (mean $r = 0.56$ Pearson, $0.47$ Spearman). Hierarchical
clustering assigns seven of the nine held-out probes to the
knowledge-benchmark cluster rather than to SocialIQA; strict pairwise counting
of the same comparison gives five of nine under Pearson and seven of nine
under Spearman, so we report the clustering and pairwise views together. One
design caveat bounds the analysis: query gradients were computed on the
Instruct checkpoint for some probes and on Base for others, and cross-variant
correlations on matched BBH tasks ($r = 0.09$ to $0.29$) bound the size of
this confound; it does not reverse any ordering reported here.

\begin{figure}[t]
  \centering
  \includegraphics[width=\textwidth]{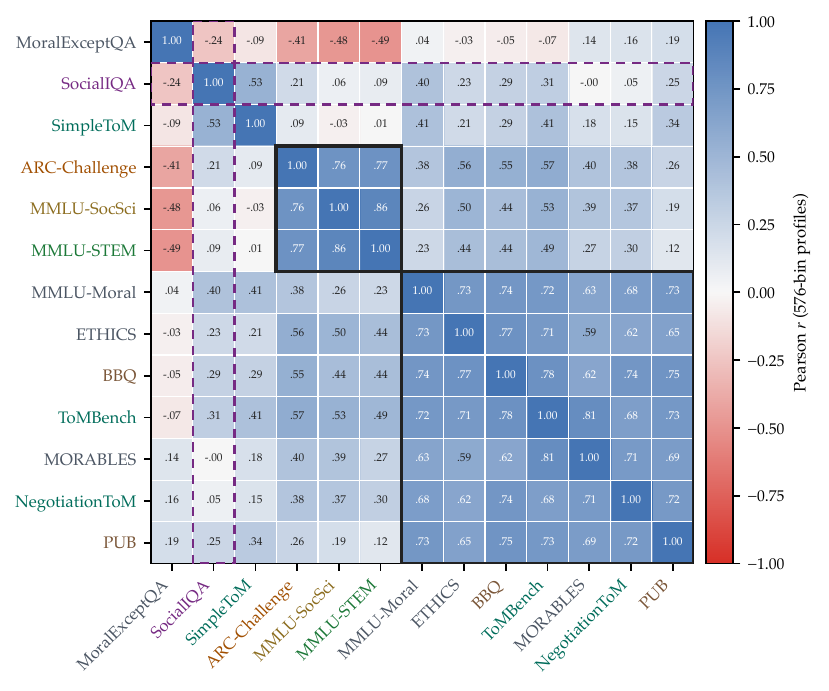}
  \caption{Annotated probe-profile correlation matrix: Pearson $r$ between
  \WebOrgNumBins{}-bin signed-influence profiles, dendrogram-ordered
  (average linkage on $1-r$). Solid outlines mark the held-out and
  knowledge blocks; the dashed row and column mark SocialIQA.}
  \label{fig:probe-similarity-full}
\end{figure}

\subsection{Causal Validation: Held-out ToM Unlearning}
\label{subsec:heldout-tom-unlearning}

Beyond attribution, we run a single held-out \emph{unlearning} test that mirrors
the per-topic single-bin procedure used for the primary benchmarks (NGDiff with
rank-8 LoRA on OLMo3-7B Base, one unlearning run per WebOrganizer topic), now
applied to the ToMBench social-pragmatic subtasks and corrected against a
null-bin control ($\gamma_{\text{net}} = \gamma_{\text{topic}} - \gamma_{\text{null}}$)
over three training seeds. Pooling the three social-pragmatic subtasks (hinting,
faux-pas recognition, and strange-story; $n=9$ subtask--seed replicates),
unlearning the \texttt{social\_life} topic degrades social-pragmatic ToM by a net
$-0.244$ in unconditional accuracy (95\% CI $[-0.380, -0.108]$; sign test $8/9$,
$p=0.039$); the effect holds on the primary metric ($-0.123$, CI
$[-0.208, -0.038]$) and survives a leave-one-subtask-out jackknife.
\texttt{social\_life} is a $z=-4.4$ outlier among the \WebOrgNumTopics{} topics
on this construct.

\begin{figure}[tb]
  \centering
  \includegraphics[width=0.8\linewidth]{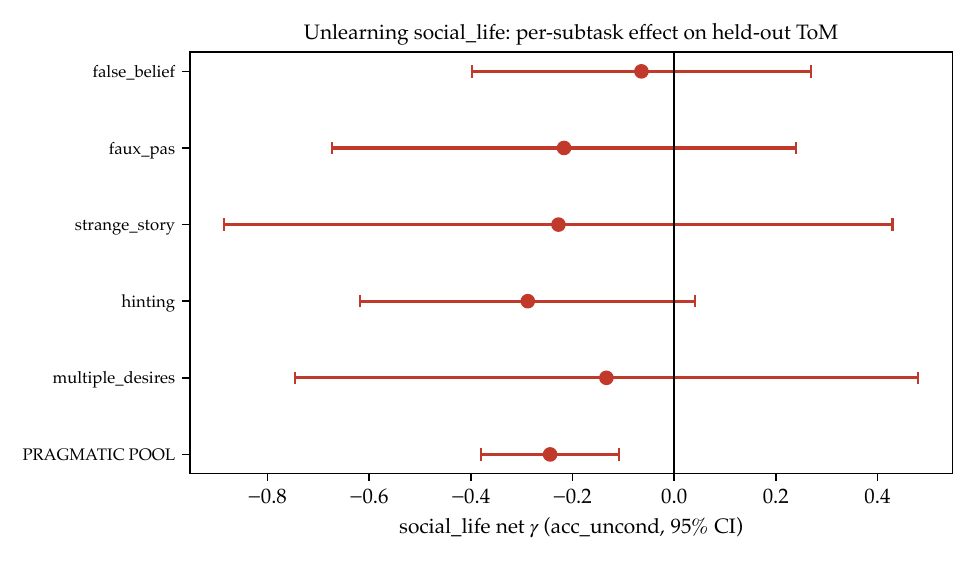}
  \caption{Held-out Theory-of-Mind unlearning. Per-subtask net influence
  ($\gamma_{\text{net}} = \gamma_{\text{topic}} - \gamma_{\text{null}}$, three
  seeds) from unlearning the \texttt{social\_life} topic, with 95\% confidence
  intervals, for the ToMBench social-pragmatic subtasks alongside the
  first-order false-belief and multiple-desires control subtasks. The pooled
  pragmatic effect excludes zero ($-0.244$ unconditional accuracy, $p=0.039$);
  the plotted control subtasks are null, as is the separate
  \texttt{science\_math\_and\_technology} topic control (not shown). The result rests on the causal unlearning test: attribution and
  unlearning do not co-rank for these later-added held-out probes.}
  \label{fig:heldout-tom-unlearning}
\end{figure}

Two controls isolate the effect. Unlearning
\texttt{science\_math\_and\_technology} against the same social-pragmatic pool
gives $-0.008$ (confidence interval includes zero), and first-order false-belief
alone is null, so the degradation is specific to social-pragmatic ToM rather than
a generic capability loss or first-order belief tracking.
Figure~\ref{fig:heldout-tom-unlearning} reports the per-subtask net effects with
confidence intervals.

We state two limits. Several ToMBench subtasks saturate on the Base model under
\texttt{acc\_per\_char}, so this held-out signal is weaker than the non-saturated
primary suite. And for these later-added probes, gradient attribution and
unlearning damage do not co-rank (topic-level Spearman $\approx 0$):
\texttt{social\_life} is the top causal topic but only middling in attribution
support. The held-out ToM claim therefore rests on the causal unlearning test,
and we do not assert attribution--unlearning convergence for it.

\subsection{Data-Side Levers for Safety-Relevant Capabilities}
\label{app:heldout-safety-levers}

Several held-out probes measure capabilities that alignment and safety
evaluation target directly: mental-state inference (ToMBench, SimpleToM),
normative judgment (ETHICS, MMLU-Moral), and social bias (BBQ). The
provenance maps make these capabilities addressable at curation time by
identifying which corpus regions to audit, enrich, or exclude when a training
run should strengthen or limit a specific social capability, and the
\texttt{social\_life} unlearning test shows that at least one such lever is
functionally necessary for social-pragmatic theory of mind
(Appendix~\ref{subsec:heldout-tom-unlearning}). The evidence boundary
matters. For the remaining held-out probes the maps are attributional;
attribution rank and causal damage did not co-rank on the later-added probes;
and locating a capability's corpus support is not the same as showing that
removing it would be safe or effective. We read this as the data-side
complement to internal-state interpretability: where probing and circuit
analysis localize a capability inside the network, corpus-scale attribution
localizes it in the training distribution, and the measurement is possible
only where weights, training checkpoints, and the document-level corpus are
released together, as in the ecosystem studied here.

\FloatBarrier

\section{Bin Characterization}\label{app:bin-characterization}

\subsection{Lexical Profiling Protocol}\label{app:rq4-lexical-enriched}

Lexical profiling is a descriptive layer applied \emph{after} attribution. The
attribution pipeline ranks Dolma3 WebOrganizer bins by signed influence for each
benchmark or probe; lexical profiling then asks what kind of language lives in
those high-influence bins. These features never select bins, compute influence
scores, or tune the attribution pipeline. Their job is interpretive: they turn
``this bin mattered'' into a readable description of the training text that
attribution surfaced.

Because Appendix~\ref{app:heldout-suite} first establishes the direct-answer
gate and tier labels for the social-reasoning probe masks, this appendix can use those tiers
as reader-facing structure. It profiles the four primary benchmarks, the
instruct BBH diagnostics, and the same social-reasoning probe masks without treating
secondary diagnostics, caution diagnostics, or failure controls as main
evidence.

\begin{SocialTDADesignBox}{Reader guide: two grouping axes}
\textbf{Benchmark target groups} are used for pooling, figure labels, and
diagnostic contrasts. They come from the paper roster in
\texttt{configs/rq4\_paper\_benchmarks.yaml}: \SocialReasoning{social
reasoning} (SocialIQA), \textcolor{SocialTDATeal}{\textbf{commonsense
reasoning}} (the instruct BBH probes), and \textbf{expository targets} (MMLU
Social Sciences, ARC-Challenge, and MMLU STEM). The expanded roster also keeps
social-reasoning probe masks in separate main, supporting, secondary, caution, and failure-control
tiers. Figure~\ref{fig:lex-group-contrast}
asks whether the original target groups differ lexically; the OpenLIWC-style heatmaps
then show the social-reasoning probe tiers without merging controls into the main evidence.

\textbf{Format clusters} are a separate, benchmark-agnostic taxonomy over
WebOrganizer formats. Figure~\ref{fig:lex-cluster-composition} uses local
colors for \LegendItem{SocialTDAPurple}{dialogic},
\LegendItem{SocialTDATeal}{personal},
\LegendItem{SocialTDAOrange}{expository},
\LegendItem{SocialTDAGold}{structured},
\LegendItem{SocialTDAGreen}{news}, and
\LegendItem{SocialTDAGray}{boilerplate}. These colors describe text form, not
target-group membership.

\textbf{LIWC-family heatmaps} use the paper's sign convention:
\PositiveAttribution{blue means above} the Dolma working-set baseline and
\NegativeAttribution{red means below} it. Printed values give the exact
full-grid $z$-score, so the figures remain readable without color.
\end{SocialTDADesignBox}

\paragraph{Scope and OpenLIWC-style implementation.}
The LIWC-adjacent run covers the paper's current 21-entry lexical-profile
roster: the four primary benchmarks, the instruct-scored BBH probes (Snarks,
Causal Judgment, and Sports Understanding), and 14 social-reasoning probe masks. The
social-reasoning probe masks are tiered as six main-evidence masks (BBQ, BBH Disambiguation
QA, ToMBench, MMLU moral/humanities, NegotiationToM, and PUB excluding
pub2/pub3), one supporting diagnostic (SimpleToM mental-state), three secondary
diagnostics (ETHICS non-justice, MoralExceptQA/RBQA, and MORABLES), one caution
diagnostic (BBH Causal Judgment), and three failure controls (ETHICS justice,
PUB pub2/pub3, and SimpleToM
behavior/judgment). StereoSet is excluded from correctness-based social-reasoning probe
attribution and is not part of this lexical roster. The checked per-bin matrix
has \WebOrgNumBins{} WebOrganizer bins and 609 columns over \AttrDocCount{} working-set
documents; the benchmark-conditioned top-bin matrix has 420 rows (21 benchmarks
$\times$ top-20 bins) and 617 columns.

We approximate LIWC-22-style constructs with an open implementation rather than
with LIWC dictionaries, LIWC.app, NRC EmoLex, or any closed analyzer. Empath category
unions cover broad semantic families such as affect, social, drives,
perception, lifestyle, and physical content~\citep{fast_empath_understanding_2016}.
In-repo lexicons cover closed-class function words, pronouns, cognition,
informal-register, and remaining perception families. Punctuation, text
statistics, and tense-focus heuristics are computed directly from the same
tokenized text stream. The four LIWC summary variables are open,
direction-preserving proxies: \emph{Analytic-open}, \emph{Tone-open},
\emph{Clout-approx}, and \emph{Authentic-approx}. They follow the construct
directions in prior LIWC summary-variable work~\citep{pennebaker_when_2014,
kacewicz_pronoun_2014,newman_lying_2003} and are mapped to $[0,100]$ with 50 as
corpus-neutral, but they are \emph{not} numerically equivalent to LIWC.app
outputs.

The implementation refuses silent parity claims. Dimensions with no defensible
open reproduction are route-G non-parity entries: \texttt{verb}/\texttt{adj}
(requires a POS tagger we do not ship), \texttt{moral}, \texttt{risk},
\texttt{curiosity}, and mental health. We validate the implementation
statically with Empath super-set checks, pronoun partition checks, punctuation
spot checks, hand-computed summary formulas, and a registry-coverage test. The
matrix, baselines, crosswalk, tokenizer choice, and route registry are released
as artifacts. Regenerating top-bin confidence intervals for the social-reasoning probe masks
is blocked unless the per-document lexical feature parquets are restored; the
checked artifact bundle contains the per-bin and benchmark-conditioned top-bin
matrices, but not the per-document lexical features needed for that bootstrap.

\begin{table}[t]
  \centering
  \scriptsize
  \setlength{\tabcolsep}{4pt}
  \renewcommand{\arraystretch}{1.12}
  \begin{tabular}{c>{\raggedright\arraybackslash}p{0.22\linewidth}>{\raggedright\arraybackslash}p{0.43\linewidth}>{\raggedright\arraybackslash}p{0.21\linewidth}}
    \toprule
    Route & Open mechanism & What it uses & Coverage claim \\
    \midrule
    A & Empath category unions & Open semantic categories are merged into LIWC-like affect, social, drives, perception, lifestyle, and physical families. & 34 realized dimensions. \\
    B & Open closed-class lexicons & In-repo lists cover pronouns, function words, cognition connectives, informal register, and perception words. & 34 realized dimensions. \\
    C & Open summary-variable proxies & Analytic-open, Tone-open, Clout-approx, and Authentic-approx follow published construct directions but are not LIWC.app scores. & 4 realized dimensions. \\
    D & Punctuation regex counts & Punctuation dimensions are transparent regex counts over the same tokenized text stream. & 7 realized dimensions. \\
    E & Text statistics & Word count, words per sentence, big-word share, and dictionary coverage are computed directly from the per-bin text. & 4 realized dimensions. \\
    F & Tense-focus heuristic & Past, present, and future focus use documented morphology and auxiliary-marker heuristics, not a shipped POS tagger. & 3 realized dimensions. \\
    G & Declared non-parity & Dimensions without a defensible open reproduction are listed explicitly rather than silently approximated. & Not realized: \texttt{verb}, \texttt{adj}, \texttt{moral}, \texttt{risk}, \texttt{curiosity}, \texttt{mental}. \\
    \bottomrule
  \end{tabular}
  \caption{OpenLIWC-style route guide and coverage audit for the lexical matrix (\WebOrgNumBins{} WebOrganizer bins, \AttrDocCount{} documents, \num{7073883562} whitespace-tokenized words; the attribution pipeline separately counts ${\sim}$10.5B subword tokens). Routes A--F produce open, LIWC-adjacent approximation columns in the checked per-bin parquet artifact rather than LIWC.app outputs; route G records dimensions we refuse to approximate because doing so would mislabel the construct.}
  \label{tab:liwc-dimension-routes}
\end{table}

\paragraph{Corpus baseline and benchmark-conditioned profiles.}
All lexical rates are computed over the working set
(\AttrDocCount{} documents, ${\sim}$\AttrTokensB{}B tokens) and aggregated to
WebOrganizer topic--format bins. Density features are reported per 1{,}000
words; the token total is the attribution tokenizer's subword count, while the
lexical rates use the whitespace-word count reported in Table~\ref{tab:liwc-dimension-routes}.
OpenLIWC-style family dimensions use the full \WebOrgNumBins{}-bin grid as the reference
distribution, so a positive full-grid $z$-score means the top-attribution bins
are denser in that family than the Dolma working-set baseline. The legacy
pronoun, mental-state, dialogue, social, and affect composites are unchanged by
the lexical enrichment; the LIWC-adjacent dimensions add a wider descriptive layer.

\subsection{Top-20 Lexical Profiles and Format Clusters}

\begin{figure}[h]
\centering
\includegraphics[width=\MainFigureWidth]{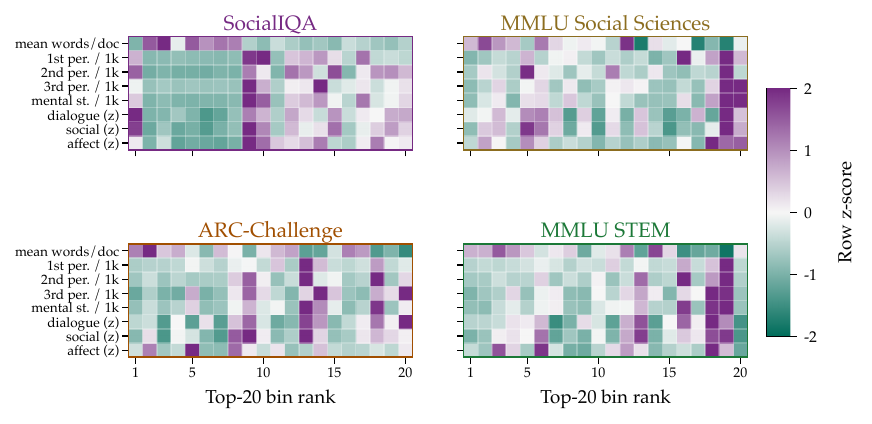}
\caption{Row-normalized lexical profiles for the top-20 bins of the four primary benchmarks. Cell values are within-row $z$-scores on a teal--white--purple diverging scale: teal marks lower-than-row-average values, white marks the row mean, and purple marks higher-than-row-average values; SocialIQA is the only primary benchmark whose high-influence bins straddle a clear interactional vs.\ documentation split.}
\label{fig:rq4-lexical-heatmap}
\end{figure}

Figure~\ref{fig:rq4-lexical-heatmap} gives the four-primary-benchmark view over
the original headline lexical variables, now using the same top-20 support as
the format-cluster and OpenLIWC-style summaries. SocialIQA is the only primary
benchmark whose high-influence bins visibly split between interactional text and
documentation-like text; the other primary benchmarks are dominated by
expository, legal, documentation, structured-data, and academic-writing bins.

\begin{figure}[t]
\centering
\includegraphics[width=\MainFigureWidth]{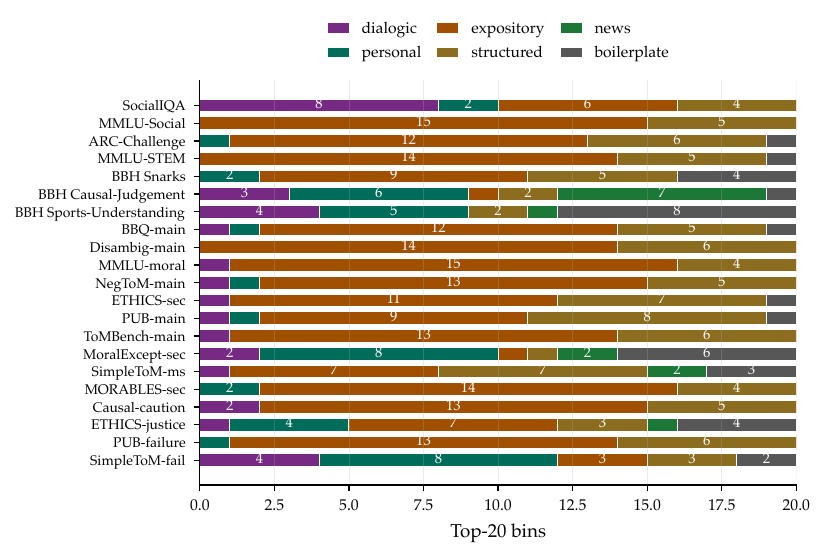}
\caption{Format-cluster composition of the top-20 high-influence bins for the scoped lexical roster. The six colors encode dialogic, personal, expository, structured, news, and boilerplate WebOrganizer formats. These clusters describe communicative function of the format, not the benchmark target groups used in Figure~\ref{fig:lex-group-contrast}. Cluster assignment is the benchmark-agnostic map in \texttt{configs/rq4\_format\_clusters.yaml}.}
\label{fig:lex-cluster-composition}
\end{figure}

\paragraph{Format clusters tell us what kind of text attribution surfaced.}
Figure~\ref{fig:lex-cluster-composition} shows the six-format-cluster
composition of each benchmark's top-20 high-influence bins. SocialIQA is
bimodal: its top bins include both dialogic/personal text and
expository/structured text. The instruct BBH probes are more heterogeneous, but
they still place more mass on dialogic and personal formats than the expository
targets. By contrast, MMLU Social Sciences, ARC-Challenge, and MMLU STEM are
dominated by expository and structured formats. The social-reasoning probe rows extend this
inspection to the probe masks and keep main, secondary, caution, and failure
controls visually separable. This is not a restatement of the paper's
$2{\times}2$ design; it is a descriptive result about the communicative forms of
the training text surfaced by attribution.

\begin{figure}[tbp]
\centering
\includegraphics[width=\MainFigureWidth]{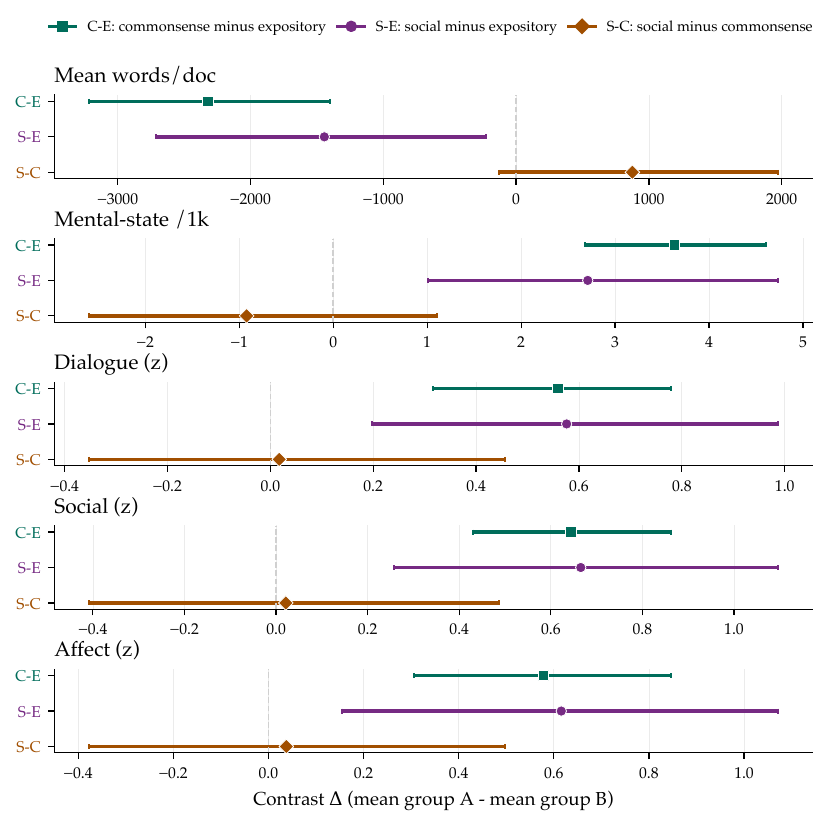}
\caption{Pooled target-group contrasts ($\Delta=$ mean group A $-$ mean group B over top-20 bins), with 95\% document-bootstrap confidence intervals. The legend defines the compact contrast codes: C--E is commonsense reasoning minus expository targets, S--E is social reasoning minus expository targets, and S--C is social reasoning minus commonsense reasoning. Intervals crossing the dashed zero line include zero.}
\label{fig:lex-group-contrast}
\end{figure}

\paragraph{Target-group contrasts test the interpersonal signature directly.}
Figure~\ref{fig:lex-group-contrast} pools the top-20 bins within each benchmark
target group and bootstraps group-mean differences (B$=$1000, seed~42). The
question is simple: compared with expository targets, do social and commonsense
targets surface shorter, more interpersonal, more affective text? Yes. Both the
social-reasoning and commonsense-reasoning groups differ from the expository
comparison group with 95\% CIs excluding zero on all five features. Social vs.\
expository has higher mental-state density ($\Delta{=}+2.71$), dialogue
($\Delta z{=}+0.58$), social language ($\Delta z{=}+0.67$), and affect
($\Delta z{=}+0.62$), with shorter documents ($\Delta{=}-1{,}444$ words/doc).
Commonsense vs.\ expository is comparable or stronger. Social and commonsense
reasoning do \emph{not} separate from each other on these features; their CIs
include zero. The result is a lexical contrast between interpersonal/pragmatic
targets and expository targets.

\subsection{OpenLIWC-Style Result Surfaces}

\begin{figure*}[t]
\centering
\includegraphics[width=\MainFigureWidth]{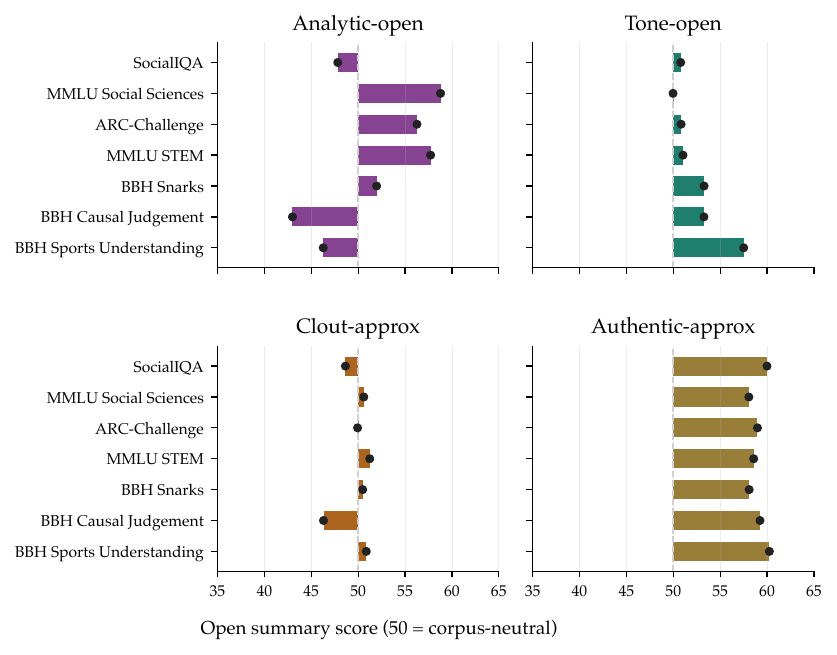}
\caption{OpenLIWC-style summary-variable proxies (Analytic-open, Tone-open, Clout-approx, Authentic-approx) per benchmark, averaged over each benchmark's top-20 high-influence bins. Scores are open, LIWC-adjacent approximations on a $[0,100]$ scale ($50=$ corpus-neutral) that follow the intended construct directions; they are not LIWC.app outputs.}
\label{fig:lex-summary-vars}
\end{figure*}

\begin{figure*}[t]
\centering
\includegraphics[width=\MainFigureWidth]{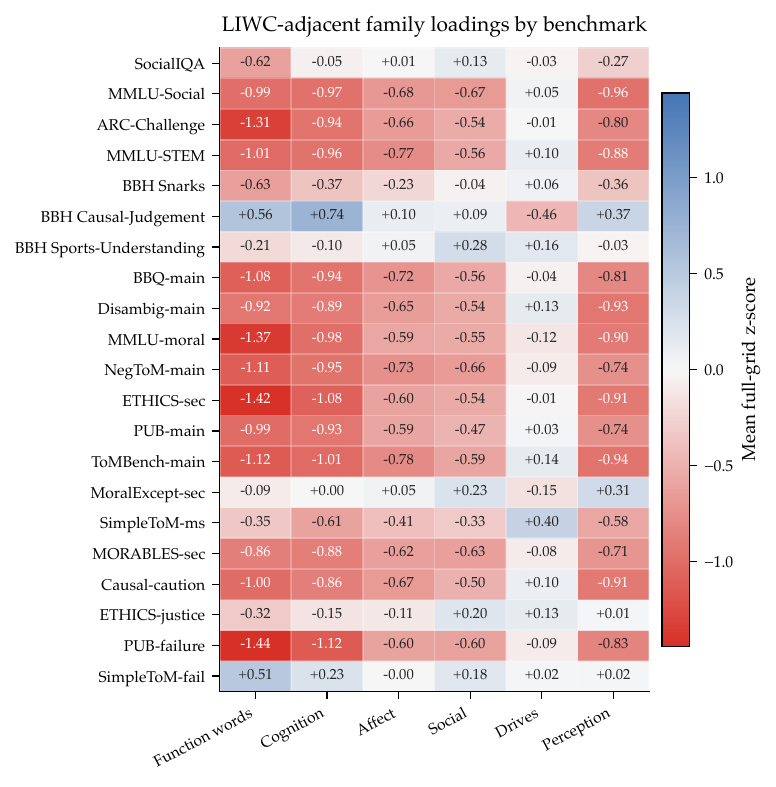}
\caption{Benchmark-level OpenLIWC-style family loadings. Each cell is the mean full-grid $z$-score over a benchmark's top-20 high-influence bins using open, LIWC-adjacent approximations rather than LIWC.app outputs; blue cells are above the Dolma working-set baseline and red cells are below it. Numeric labels make the figure readable without color.}
\label{fig:liwc-families-bybenchmark}
\end{figure*}

\begin{figure}[t]
\centering
\includegraphics[width=\MainFigureWidth]{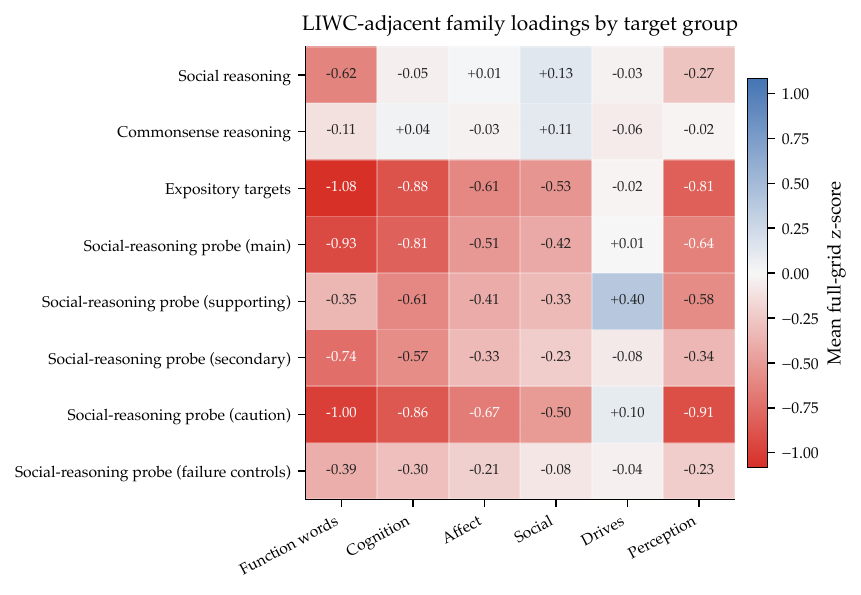}
\caption{Grouped OpenLIWC-style family loadings over pooled top-20 high-influence bins. The values are open, LIWC-adjacent approximations rather than LIWC.app outputs; blue cells are above the Dolma working-set baseline and red cells are below it. Printed values show the mean full-grid $z$-score.}
\label{fig:liwc-families}
\end{figure}

\begin{figure*}[t]
\centering
\includegraphics[width=\MainFigureWidth]{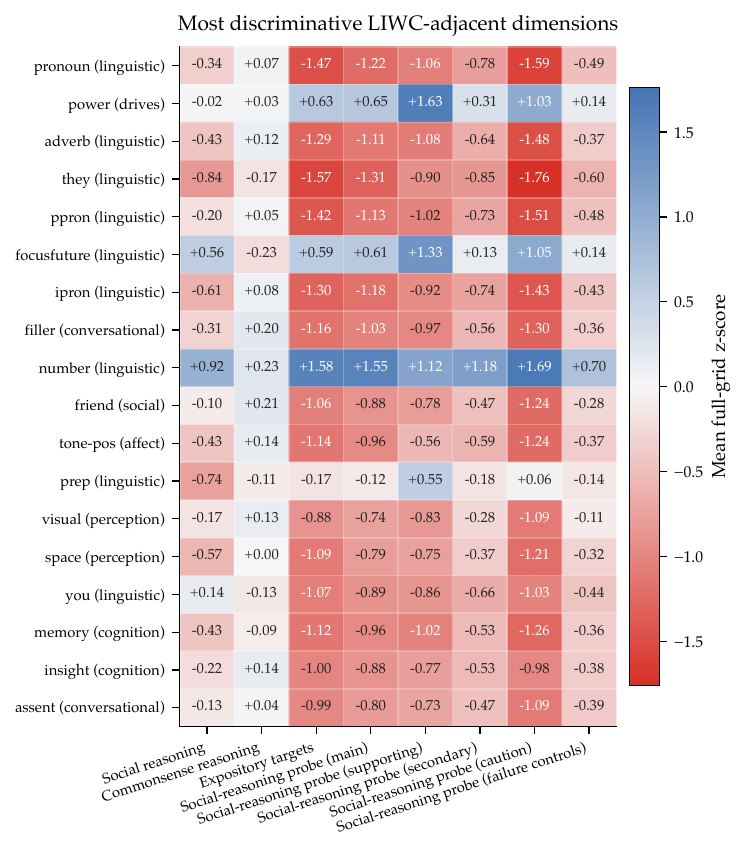}
\caption{Dimension-level OpenLIWC-style loadings for the 18 realized LIWC-adjacent dimensions with the largest target-group range over pooled top-20 bins. Blue cells are above the Dolma working-set baseline and red cells are below it; printed values show the mean full-grid $z$-score. This figure exposes the dimensions behind the family-level summaries in Figures~\ref{fig:liwc-families-bybenchmark}--\ref{fig:liwc-families}; values are not LIWC.app outputs.}
\label{fig:liwc-dimensions}
\end{figure*}

The OpenLIWC-style approximation appears in four places in this appendix. Table~\ref{tab:liwc-dimension-routes}
audits what is implemented and what is declared non-parity. Figure~\ref{fig:lex-summary-vars}
shows the four open summary-variable proxies. Figures
\ref{fig:liwc-families-bybenchmark}--\ref{fig:liwc-families} summarize the
family-level loadings, and Figure~\ref{fig:liwc-dimensions} opens the family
averages into the highest-separation realized dimensions. Together, these plots
answer two questions: whether the top-attribution bins are above or below the
Dolma baseline, and whether that lexical direction changes across the original
target groups and the tiered social-reasoning probe masks.

\paragraph{LIWC-adjacent families broaden the same story.}
Figure~\ref{fig:lex-summary-vars} reports the four open summary variables per
benchmark. Expository targets score higher on \emph{Analytic-open}; social and
commonsense targets are less analytic/formal and closer to the tone and
interpersonal patterns expected from the composite analysis above. Figures
\ref{fig:liwc-families-bybenchmark}--\ref{fig:liwc-families} expose the broader
LIWC-family loadings, with exact values printed in every cell. At the benchmark
level, the expository targets are consistently below the Dolma working-set
baseline on function-word, cognition, affect, social, and perception families.
At the target-group level, social and commonsense reasoning sit near or above
the corpus average on the social family, whereas expository targets are negative
across every family, with drives essentially at baseline ($-0.02$). The social-reasoning probe tiers are also shown as
diagnostics: main and secondary masks are below the corpus baseline on most
OpenLIWC-style families, and the failure controls remain separate rather than being
pooled into the main tier. Figure~\ref{fig:liwc-dimensions} shows which realized
dimensions carry the strongest target-group separation.
Table~\ref{tab:liwc-interpretation} summarizes the intended claim tier for each
target group before the row-level top-bin tables.

\FloatBarrier

\begin{table}[t]
  \centering
  \scriptsize
  \setlength{\tabcolsep}{3.5pt}
  \renewcommand{\arraystretch}{1.12}
  \begin{tabular}{>{\raggedright\arraybackslash}p{0.15\linewidth}>{\raggedright\arraybackslash}p{0.25\linewidth}>{\raggedright\arraybackslash}p{0.29\linewidth}>{\raggedright\arraybackslash}p{0.24\linewidth}}
    \toprule
    Target group & Observed lexical pattern & Strongest evidence & Narrative takeaway \\
    \midrule
    \textbf{Social reasoning} & Original SocialIQA target: mixed dialogic/personal bins and documentation-like bins. & Figure~\ref{fig:liwc-families} gives social-family $z=+0.13$. & Use as the original social-reasoning lexical profile, not as held-out social-reasoning probe evidence. \\
    \textbf{Commonsense reasoning} & Original BBH instruct probes: pragmatic and commonsense diagnostics. & Figure~\ref{fig:liwc-families} gives social-family $z=+0.11$. & Keep separate from the social-reasoning probe tiers in the written claims. \\
    \textbf{Expository targets} & MMLU-Social, ARC-Challenge, and MMLU-STEM are dominated by expository and structured bins. & Figure~\ref{fig:liwc-families} is below baseline on cognition ($z=-0.88$), affect ($z=-0.61$), social ($z=-0.53$), and perception ($z=-0.81$). & The expository group supplies the baseline contrast: high-attribution bins look more like formal reference/documentation text than interpersonal discourse. \\
    \textbf{Social-reasoning probe (main)} & BBQ, Disambiguation QA, ToMBench, moral/humanities MMLU, NegotiationToM, and PUB excluding pub2/pub3. & Figure~\ref{fig:liwc-families} gives social-family $z=-0.42$ and affect-family $z=-0.51$. & Treat this tier as the main social-reasoning probe lexical evidence. \\
    \textbf{Social-reasoning probe (supporting)} & SimpleToM mental-state is retained as a supporting diagnostic. & Figure~\ref{fig:liwc-families} gives social-family $z=-0.33$ and affect-family $z=-0.41$. & Use as relevant mental-state evidence without upgrading the full SimpleToM suite. \\
    \textbf{Social-reasoning probe (secondary)} & ETHICS non-justice, MoralExceptQA/RBQA, and MORABLES. & Figure~\ref{fig:liwc-families} gives social-family $z=-0.23$ and affect-family $z=-0.33$. & Use as secondary diagnostics, not as the headline evidence tier. \\
    \textbf{Social-reasoning probe (caution)} & BBH Causal Judgment is retained as a caution diagnostic. & Figure~\ref{fig:liwc-families} gives social-family $z=-0.50$. & Keep this tier visually available but avoid treating it as positive main evidence. \\
    \textbf{Social-reasoning probe (failure controls)} & ETHICS justice, PUB pub2/pub3, and SimpleToM behavior/judgment. & Figure~\ref{fig:liwc-families} gives social-family $z=-0.08$. & Interpret as controls and diagnostics only. \\
    \bottomrule
  \end{tabular}
  \caption{Reader-facing interpretation of the lexical appendix. The target groups are benchmark groupings used for pooling and contrasts; the social-reasoning probe tiers separate main evidence, supporting diagnostics, secondary diagnostics, caution diagnostics, and failure controls. The format clusters in Figure~\ref{fig:lex-cluster-composition} are a separate communicative-function taxonomy over WebOrganizer formats. Exact OpenLIWC-style family and dimension values are printed in Figures~\ref{fig:liwc-families-bybenchmark}--\ref{fig:liwc-dimensions}; they are open, LIWC-adjacent approximations rather than LIWC.app outputs.}
  \label{tab:liwc-interpretation}
\end{table}

\FloatBarrier

\subsection{Per-Benchmark Top-20 Bin Tables}

Tables~\ref{tab:rq4-profiles}--\ref{tab:rq4-profiles-expanded-socialtda-simpletom-failure-behavior-judgment}
give the row-level top-20 lexical evidence behind the figures for every
benchmark in the scoped lexical roster. We include the four primary benchmarks,
the three instruct BBH probes, and the 14 social-reasoning probe masks. Base-model BBH tables
remain legacy support files and are not part of the current appendix narrative.
Each table uses the same six format clusters as Figure~\ref{fig:lex-cluster-composition},
so the reader can move directly from aggregate composition to the actual bins.

For SocialIQA (Table~\ref{tab:rq4-profiles}), the split is clean and bimodal:
the top-20 divides 10/10 into interactional (dialogic + personal) and expository
(expository + structured) formats. The interactional rows contain short,
dialogue-rich text with higher first-/second-person pronoun density and
mental-state language, while the expository and structured rows contain longer,
formal documentation-like text. Literature/\allowbreak Customer Support remains
the dominant outlier in the correctness-differential analysis
(Appendix~\ref{app:correctness}), pointing to a single training-data substrate
whose effect reverses between social reasoning and factual recall.

\begin{table}[h]
\centering
\small
\resizebox{\linewidth}{!}{%
\begin{tabular}{rlrrrrrrrrr}
\toprule
\RQProfileHeader{SocialTDAPurple}
 & Bin & Docs & \shortstack{Mean\\words} & \shortstack{1st per.\\per 1k} & \shortstack{2nd per.\\per 1k} & \shortstack{3rd per.\\per 1k} & \shortstack{Mental st.\\per 1k} & \shortstack{Dialogue\\($z$)} & \shortstack{Social\\($z$)} & \shortstack{Affect\\($z$)} \\
\midrule
\RQProfileGroupRow{Dialogic formats}
1 & Literature / Customer Support & 10{,}000 & 245 & 26.89 & 33.39 & 9.20 & 6.22 & +2.38 & +1.82 & +0.13 \\
9 & Social Life / Q\&A Forum & 10{,}000 & 1{,}148 & 46.36 & 30.01 & 33.99 & 15.57 & +1.07 & +2.10 & +2.44 \\
12 & Adult / FAQ & 10{,}000 & 1{,}034 & 23.87 & 31.29 & 10.85 & 6.14 & +0.76 & +0.69 & +0.27 \\
13 & Fashion \& Beauty / Q\&A Forum & 10{,}000 & 622 & 26.33 & 26.43 & 11.60 & 6.81 & +0.35 & +0.03 & +0.51 \\
15 & Adult / Customer Support & 9{,}493 & 433 & 20.07 & 36.00 & 5.81 & 4.24 & +0.91 & +1.37 & -0.41 \\
17 & Art \& Design / Comment Section & 10{,}000 & 1{,}202 & 36.77 & 16.15 & 11.79 & 9.43 & +0.22 & +0.02 & +0.11 \\
18 & Health / Customer Support & 10{,}000 & 830 & 9.00 & 26.97 & 6.03 & 3.46 & +0.12 & +0.51 & +1.10 \\
\bottomrule
\end{tabular}%
}
\caption{Lexical profiles of the top-20 \SocialReasoning{SocialIQA} bins by signed mean influence. Densities are per 1{,}000 words; corpus-wide weighted means are 16.1 first-person pronouns, 11.4 second-person pronouns, and 5.1 mental-state terms per 1{,}000 words. Dialogue, social, and affect columns are full-grid $z$-scores against all 576 WebOrganizer bins. Rows use the six format clusters in \texttt{configs/rq4\_format\_clusters.yaml}; only clusters present in the benchmark's top-20 bins are shown.}
\label{tab:rq4-profiles}
\end{table}

\begin{table}[h]
\centering
\small
\resizebox{\linewidth}{!}{%
\begin{tabular}{rlrrrrrrrrr}
\toprule
\RQProfileHeader{SocialTDAGold}
 & Bin & Docs & \shortstack{Mean\\words} & \shortstack{1st per.\\per 1k} & \shortstack{2nd per.\\per 1k} & \shortstack{3rd per.\\per 1k} & \shortstack{Mental st.\\per 1k} & \shortstack{Dialogue\\($z$)} & \shortstack{Social\\($z$)} & \shortstack{Affect\\($z$)} \\
\midrule
\RQProfileGroupRow{Expository formats}
1 & Industrial / Documentation & 10{,}000 & 5{,}738 & 1.46 & 0.61 & 1.46 & 0.64 & -0.66 & -0.93 & -0.89 \\
2 & Health / Documentation & 10{,}000 & 7{,}473 & 1.87 & 1.95 & 2.58 & 1.07 & -0.62 & -0.53 & -0.30 \\
3 & Politics / Documentation & 10{,}000 & 6{,}264 & 2.41 & 1.19 & 5.13 & 1.32 & -0.44 & -0.49 & -0.67 \\
6 & Crime \& Law / Documentation & 10{,}000 & 6{,}735 & 2.19 & 1.85 & 5.09 & 1.44 & -0.28 & -0.28 & -0.51 \\
8 & Industrial / Academic Writing & 10{,}000 & 4{,}655 & 3.24 & 0.50 & 4.51 & 1.57 & -0.98 & -1.01 & -0.85 \\
9 & Science \& Technology / Documentation & 10{,}000 & 4{,}953 & 2.37 & 1.44 & 1.58 & 0.92 & -0.50 & -0.63 & -0.93 \\
10 & Food \& Dining / Documentation & 8{,}381 & 4{,}706 & 1.87 & 1.01 & 2.13 & 0.70 & -0.95 & -1.08 & -1.11 \\
\bottomrule
\end{tabular}%
}
\caption{Lexical profiles of the top-20 \SocialKnowledge{MMLU Social Sciences} bins by signed mean influence. Densities are per 1{,}000 words. Dialogue, social, and affect columns are full-grid $z$-scores against all 576 WebOrganizer bins. Rows use the six format clusters in \texttt{configs/rq4\_format\_clusters.yaml}; only clusters present in the benchmark's top-20 bins are shown.}
\label{tab:rq4-profiles-expanded-mmlu-social-science}
\end{table}

\begin{table}[h]
\centering
\small
\resizebox{\linewidth}{!}{%
\begin{tabular}{rlrrrrrrrrr}
\toprule
\RQProfileHeader{SocialTDAOrange}
 & Bin & Docs & \shortstack{Mean\\words} & \shortstack{1st per.\\per 1k} & \shortstack{2nd per.\\per 1k} & \shortstack{3rd per.\\per 1k} & \shortstack{Mental st.\\per 1k} & \shortstack{Dialogue\\($z$)} & \shortstack{Social\\($z$)} & \shortstack{Affect\\($z$)} \\
\midrule
\RQProfileGroupRow{Personal formats}
17 & Fashion \& Beauty / User Review & 10{,}000 & 493 & 44.53 & 15.62 & 12.00 & 7.46 & -0.39 & -0.29 & +0.77 \\
\midrule
\RQProfileGroupRow{Expository formats}
1 & Industrial / Documentation & 10{,}000 & 5{,}738 & 1.46 & 0.61 & 1.46 & 0.64 & -0.66 & -0.93 & -0.89 \\
2 & Health / Documentation & 10{,}000 & 7{,}473 & 1.87 & 1.95 & 2.58 & 1.07 & -0.62 & -0.53 & -0.30 \\
4 & Politics / Documentation & 10{,}000 & 6{,}264 & 2.41 & 1.19 & 5.13 & 1.32 & -0.44 & -0.49 & -0.67 \\
6 & Industrial / Academic Writing & 10{,}000 & 4{,}655 & 3.24 & 0.50 & 4.51 & 1.57 & -0.98 & -1.01 & -0.85 \\
7 & Travel \& Tourism / Documentation & 816 & 3{,}963 & 2.48 & 3.56 & 4.24 & 1.24 & -0.58 & -0.58 & -0.92 \\
10 & Science \& Technology / Academic Writing & 10{,}000 & 4{,}758 & 7.38 & 0.33 & 2.82 & 1.42 & -0.89 & -0.92 & -0.69 \\
\bottomrule
\end{tabular}%
}
\caption{Lexical profiles of the top-20 \STEMReasoning{ARC-Challenge} bins by signed mean influence. Densities are per 1{,}000 words. Dialogue, social, and affect columns are full-grid $z$-scores against all 576 WebOrganizer bins. Rows use the six format clusters in \texttt{configs/rq4\_format\_clusters.yaml}; only clusters present in the benchmark's top-20 bins are shown.}
\label{tab:rq4-profiles-expanded-arc-challenge}
\end{table}

\begin{table}[h]
\centering
\small
\resizebox{\linewidth}{!}{%
\begin{tabular}{rlrrrrrrrrr}
\toprule
\RQProfileHeader{SocialTDAGreen}
 & Bin & Docs & \shortstack{Mean\\words} & \shortstack{1st per.\\per 1k} & \shortstack{2nd per.\\per 1k} & \shortstack{3rd per.\\per 1k} & \shortstack{Mental st.\\per 1k} & \shortstack{Dialogue\\($z$)} & \shortstack{Social\\($z$)} & \shortstack{Affect\\($z$)} \\
\midrule
\RQProfileGroupRow{Expository formats}
1 & Industrial / Documentation & 10{,}000 & 5{,}738 & 1.46 & 0.61 & 1.46 & 0.64 & -0.66 & -0.93 & -0.89 \\
3 & Health / Documentation & 10{,}000 & 7{,}473 & 1.87 & 1.95 & 2.58 & 1.07 & -0.62 & -0.53 & -0.30 \\
4 & Politics / Documentation & 10{,}000 & 6{,}264 & 2.41 & 1.19 & 5.13 & 1.32 & -0.44 & -0.49 & -0.67 \\
7 & Industrial / Academic Writing & 10{,}000 & 4{,}655 & 3.24 & 0.50 & 4.51 & 1.57 & -0.98 & -1.01 & -0.85 \\
8 & Home \& Hobbies / Documentation & 10{,}000 & 5{,}151 & 1.38 & 1.88 & 1.86 & 0.69 & -0.82 & -0.82 & -1.08 \\
10 & Travel \& Tourism / Documentation & 816 & 3{,}963 & 2.48 & 3.56 & 4.24 & 1.24 & -0.58 & -0.58 & -0.92 \\
11 & Science \& Technology / Academic Writing & 10{,}000 & 4{,}758 & 7.38 & 0.33 & 2.82 & 1.42 & -0.89 & -0.92 & -0.69 \\
\bottomrule
\end{tabular}%
}
\caption{Lexical profiles of the top-20 \STEMKnowledge{MMLU STEM} bins by signed mean influence. Densities are per 1{,}000 words. Dialogue, social, and affect columns are full-grid $z$-scores against all 576 WebOrganizer bins. Rows use the six format clusters in \texttt{configs/rq4\_format\_clusters.yaml}; only clusters present in the benchmark's top-20 bins are shown.}
\label{tab:rq4-profiles-expanded-mmlu-stem}
\end{table}

\begin{table}[h]
\centering
\small
\resizebox{\linewidth}{!}{%
\begin{tabular}{rlrrrrrrrrr}
\toprule
\RQProfileHeader{SocialTDASlate}
 & Bin & Docs & \shortstack{Mean\\words} & \shortstack{1st per.\\per 1k} & \shortstack{2nd per.\\per 1k} & \shortstack{3rd per.\\per 1k} & \shortstack{Mental st.\\per 1k} & \shortstack{Dialogue\\($z$)} & \shortstack{Social\\($z$)} & \shortstack{Affect\\($z$)} \\
\midrule
\RQProfileGroupRow{Personal formats}
1 & Adult / User Review & 10{,}000 & 1{,}192 & 25.37 & 19.47 & 13.96 & 7.39 & +0.00 & +0.66 & +0.41 \\
20 & Food \& Dining / Creative Writing & 10{,}000 & 967 & 30.09 & 11.15 & 32.53 & 7.34 & +0.15 & -0.07 & +0.40 \\
\midrule
\RQProfileGroupRow{Expository formats}
2 & Science \& Technology / Academic Writing & 10{,}000 & 4{,}758 & 7.38 & 0.33 & 2.82 & 1.42 & -0.89 & -0.92 & -0.69 \\
3 & Crime \& Law / Academic Writing & 10{,}000 & 4{,}606 & 4.19 & 0.90 & 11.52 & 2.66 & -0.46 & -0.44 & -0.00 \\
4 & Politics / Documentation & 10{,}000 & 6{,}264 & 2.41 & 1.19 & 5.13 & 1.32 & -0.44 & -0.49 & -0.67 \\
5 & Home \& Hobbies / Documentation & 10{,}000 & 5{,}151 & 1.38 & 1.88 & 1.86 & 0.69 & -0.82 & -0.82 & -1.08 \\
6 & Literature / Documentation & 7{,}164 & 1{,}526 & 9.42 & 4.86 & 7.84 & 2.40 & +1.13 & +0.56 & -0.68 \\
\bottomrule
\end{tabular}%
}
\caption{Lexical profiles of the top-20 BBH Snarks (instruct) bins by signed mean influence. Densities are per 1{,}000 words. Dialogue, social, and affect columns are full-grid $z$-scores against all 576 WebOrganizer bins. Rows use the six format clusters in \texttt{configs/rq4\_format\_clusters.yaml}; only clusters present in the benchmark's top-20 bins are shown.}
\label{tab:rq4-profiles-expanded-bbh-snarks-instruct}
\end{table}

\begin{table}[h]
\centering
\small
\resizebox{\linewidth}{!}{%
\begin{tabular}{rlrrrrrrrrr}
\toprule
\RQProfileHeader{SocialTDASlate}
 & Bin & Docs & \shortstack{Mean\\words} & \shortstack{1st per.\\per 1k} & \shortstack{2nd per.\\per 1k} & \shortstack{3rd per.\\per 1k} & \shortstack{Mental st.\\per 1k} & \shortstack{Dialogue\\($z$)} & \shortstack{Social\\($z$)} & \shortstack{Affect\\($z$)} \\
\midrule
\RQProfileGroupRow{Dialogic formats}
2 & Literature / Customer Support & 10{,}000 & 245 & 26.89 & 33.39 & 9.20 & 6.22 & +2.38 & +1.82 & +0.13 \\
16 & Games / Q\&A Forum & 10{,}000 & 990 & 19.79 & 22.06 & 14.72 & 6.83 & +0.85 & +0.16 & -0.20 \\
17 & Entertainment / Q\&A Forum & 10{,}000 & 1{,}017 & 17.72 & 11.63 & 28.29 & 7.88 & +1.62 & +0.94 & +0.20 \\
\midrule
\RQProfileGroupRow{Personal formats}
3 & History \& Geography / User Review & 10{,}000 & 903 & 12.05 & 3.07 & 18.45 & 4.77 & -0.32 & -0.44 & +0.11 \\
5 & Games / About (Personal) & 10{,}000 & 1{,}115 & 31.63 & 13.71 & 22.40 & 8.40 & +0.05 & -0.03 & +0.23 \\
6 & Entertainment / User Review & 10{,}000 & 952 & 15.28 & 5.73 & 24.95 & 6.69 & +0.27 & +0.39 & +1.16 \\
10 & Home \& Hobbies / Creative Writing & 10{,}000 & 870 & 31.00 & 10.37 & 45.45 & 8.16 & -0.18 & +0.33 & +1.10 \\
\bottomrule
\end{tabular}%
}
\caption{Lexical profiles of the top-20 BBH Causal Judgment (instruct) bins by signed mean influence. Densities are per 1{,}000 words. Dialogue, social, and affect columns are full-grid $z$-scores against all 576 WebOrganizer bins. Rows use the six format clusters in \texttt{configs/rq4\_format\_clusters.yaml}; only clusters present in the benchmark's top-20 bins are shown.}
\label{tab:rq4-profiles-expanded-bbh-causal-judgement-instruct}
\end{table}

\begin{table}[h]
\centering
\small
\resizebox{\linewidth}{!}{%
\begin{tabular}{rlrrrrrrrrr}
\toprule
\RQProfileHeader{SocialTDASlate}
 & Bin & Docs & \shortstack{Mean\\words} & \shortstack{1st per.\\per 1k} & \shortstack{2nd per.\\per 1k} & \shortstack{3rd per.\\per 1k} & \shortstack{Mental st.\\per 1k} & \shortstack{Dialogue\\($z$)} & \shortstack{Social\\($z$)} & \shortstack{Affect\\($z$)} \\
\midrule
\RQProfileGroupRow{Dialogic formats}
1 & Travel \& Tourism / Q\&A Forum & 10{,}000 & 492 & 27.90 & 29.76 & 8.47 & 6.94 & +0.58 & +0.76 & -0.55 \\
2 & History \& Geography / Customer Support & 7{,}740 & 288 & 18.00 & 22.72 & 5.78 & 4.28 & +0.99 & +0.74 & -0.83 \\
4 & Education \& Jobs / FAQ & 10{,}000 & 837 & 16.32 & 24.02 & 8.93 & 4.23 & +0.95 & +0.48 & -0.68 \\
19 & Games / Q\&A Forum & 10{,}000 & 990 & 19.79 & 22.06 & 14.72 & 6.83 & +0.85 & +0.16 & -0.20 \\
\midrule
\RQProfileGroupRow{Personal formats}
9 & Games / About (Personal) & 10{,}000 & 1{,}115 & 31.63 & 13.71 & 22.40 & 8.40 & +0.05 & -0.03 & +0.23 \\
11 & Religion / About (Personal) & 10{,}000 & 565 & 47.25 & 13.24 & 22.61 & 10.49 & +0.03 & +0.56 & +1.69 \\
12 & Travel \& Tourism / User Review & 10{,}000 & 799 & 36.81 & 10.60 & 10.33 & 5.26 & -0.21 & -0.08 & -0.16 \\
\bottomrule
\end{tabular}%
}
\caption{Lexical profiles of the top-20 BBH Sports Understanding (instruct) bins by signed mean influence. Densities are per 1{,}000 words. Dialogue, social, and affect columns are full-grid $z$-scores against all 576 WebOrganizer bins. Rows use the six format clusters in \texttt{configs/rq4\_format\_clusters.yaml}; only clusters present in the benchmark's top-20 bins are shown.}
\label{tab:rq4-profiles-expanded-bbh-sports-understanding-instruct}
\end{table}

\begin{table}[h]
\centering
\small
\resizebox{\linewidth}{!}{%
\begin{tabular}{rlrrrrrrrrr}
\toprule
\RQProfileHeader{SocialTDATeal}
 & Bin & Docs & \shortstack{Mean\\words} & \shortstack{1st per.\\per 1k} & \shortstack{2nd per.\\per 1k} & \shortstack{3rd per.\\per 1k} & \shortstack{Mental st.\\per 1k} & \shortstack{Dialogue\\($z$)} & \shortstack{Social\\($z$)} & \shortstack{Affect\\($z$)} \\
\midrule
\RQProfileGroupRow{Dialogic formats}
5 & Literature / Customer Support & 10{,}000 & 245 & 26.89 & 33.39 & 9.20 & 6.22 & +2.38 & +1.82 & +0.13 \\
\midrule
\RQProfileGroupRow{Personal formats}
16 & Industrial / Creative Writing & 8{,}443 & 2{,}002 & 27.00 & 8.71 & 34.07 & 6.44 & -0.23 & -0.32 & +0.13 \\
\midrule
\RQProfileGroupRow{Expository formats}
1 & Industrial / Documentation & 10{,}000 & 5{,}738 & 1.46 & 0.61 & 1.46 & 0.64 & -0.66 & -0.93 & -0.89 \\
3 & Health / Documentation & 10{,}000 & 7{,}473 & 1.87 & 1.95 & 2.58 & 1.07 & -0.62 & -0.53 & -0.30 \\
6 & Politics / Documentation & 10{,}000 & 6{,}264 & 2.41 & 1.19 & 5.13 & 1.32 & -0.44 & -0.49 & -0.67 \\
7 & Industrial / Academic Writing & 10{,}000 & 4{,}655 & 3.24 & 0.50 & 4.51 & 1.57 & -0.98 & -1.01 & -0.85 \\
8 & Home \& Hobbies / Documentation & 10{,}000 & 5{,}151 & 1.38 & 1.88 & 1.86 & 0.69 & -0.82 & -0.82 & -1.08 \\
\bottomrule
\end{tabular}%
}
\caption{Lexical profiles of the top-20 BBQ main bins by signed mean influence. Densities are per 1{,}000 words. Dialogue, social, and affect columns are full-grid $z$-scores against all 576 WebOrganizer bins. Rows use the six format clusters in \texttt{configs/rq4\_format\_clusters.yaml}; only clusters present in the benchmark's top-20 bins are shown.}
\label{tab:rq4-profiles-expanded-socialtda-bbq-main}
\end{table}

\input{tables/tab-rq4-profiles-expanded-socialtda-bbh-disambiguation-qa-main}

\input{tables/tab-rq4-profiles-expanded-socialtda-mmlu-moral-main}

\input{tables/tab-rq4-profiles-expanded-socialtda-negotiationtom-main}

\input{tables/tab-rq4-profiles-expanded-socialtda-ethics-main-non-justice}

\input{tables/tab-rq4-profiles-expanded-socialtda-pub-main-excluding-pub-2-pub-3}

\input{tables/tab-rq4-profiles-expanded-socialtda-tombench-secondary}

\input{tables/tab-rq4-profiles-expanded-socialtda-moralexceptqa-rbqa-secondary}

\input{tables/tab-rq4-profiles-expanded-socialtda-simpletom-mental-state}

\input{tables/tab-rq4-profiles-expanded-socialtda-morables-secondary}

\input{tables/tab-rq4-profiles-expanded-socialtda-bbh-causal-judgment-caution}

\input{tables/tab-rq4-profiles-expanded-socialtda-ethics-failure-justice}

\input{tables/tab-rq4-profiles-expanded-socialtda-pub-failure-pub-2-pub-3}

\input{tables/tab-rq4-profiles-expanded-socialtda-simpletom-failure-behavior-judgment}

\FloatBarrier

\section{Unlearning Experimental Setup}\label{app:unlearning}

All unlearning experiments in this paper are performed on OLMo3-7B Base (the unmodified pretrained checkpoint). Baseline benchmark scores against which the accuracy drop $\gamma$ is measured come from evaluating the same OLMo3-7B Base checkpoint under the OLMES \texttt{:mc::olmes} protocol; the post-unlearning evaluation uses the identical protocol on the merged base+adapter checkpoint, so the $\gamma$ comparison is config-matched. All reported SocialIQA $\gamma$ values for on-target and random-control conditions use the config-matched OLMES baseline of $0.80334$; earlier fast-eval SocialIQA pilots are not used for reported numbers.

\subsection{Objective and Optimization}

All unlearning experiments use NGDiff~\citep{bu2025ngdiff}. The update direction normalizes
retain and forget gradients independently:
\[
g_{\mathrm{update}} = \frac{g_{\mathrm{retain}}}{\|g_{\mathrm{retain}}\|} -
\frac{g_{\mathrm{forget}}}{\|g_{\mathrm{forget}}\|},
\]
so that neither component dominates regardless of bin size or loss scale. The learning rate
is adapted automatically every 10 steps via two additional forward passes on the retain set
(Hessian-free schedule). We fine-tune only LoRA parameters (rank~8, applied to Q/K/V/O
projections), following the published NGDiff setup for 7B-scale models;
the complete hyperparameter configuration appears in
Table~\ref{tab:unlearn-hyperparams}.

\begin{table}[htbp]
\centering
\caption{Unlearning hyperparameters.}\label{tab:unlearn-hyperparams}
\begin{tabular}{ll}
\toprule
\rowcolor{SocialTDASlate!8}
\textbf{Parameter} & \textbf{Value} \\
\midrule
Method & NGDiff with GeN AutoLR \\
LoRA Configuration & Rank 8; Q, K, V, O projections \\
Batch Size & 16 (per GPU) \\
Effective Batch Size & 160 (via Gradient Accumulation) \\
Precision & bf16 \\
Max Steps & 5,000 \\
\bottomrule
\end{tabular}
\end{table}

\subsection{Data Configuration and Sampling}

We standardize the data pipeline across target topics.

\paragraph{Forget Set ($\mathcal{D}_{F}$)} For each single-bin sweep run, we sample exactly 2,000 documents uniformly at random from the target topic. The faithful per-document replication instead uses the top-200 influence-ranked documents per (topic, benchmark) pair, with 1,000-document random in-topic controls.

\paragraph{Retain Set ($\mathcal{D}_{R}$)} The retain set consists of 9,000 documents sampled uniformly at random from outside the target topic (for corpus-global runs, from the remaining corpus disjoint from the forget set). It is sampled once per run and held fixed for the duration of training.

\paragraph{Length-Normalized Loss} To prevent long documents from disproportionately biasing the gradient, we apply length-normalization to the loss calculation (SimNPO style;~\citealp{fan_simnpo_2025}). The per-document loss is averaged over its token count before aggregating into the batch-level mean.

Table~\ref{tab:unlearn-data} summarizes the forget/retain data configuration
used for the 2,000-document single-bin sweep; the faithful per-document
replication in Appendix~\ref{app:unlearning-paired-significance} uses top-200
influence-ranked forget sets compared against 1,000-document random in-topic
controls.

\begin{table}[htbp]
\centering
\caption{Data and sampling configuration.}\label{tab:unlearn-data}
\begin{tabular}{ll}
\toprule
\rowcolor{SocialTDASlate!8}
\textbf{Parameter} & \textbf{Value} \\
\midrule
Forget Set Size ($k$) & 2,000 documents \\
Retain Set Size & 9,000 documents (uniform, out-of-bin) \\
Retain Refresh Rate & None (fixed for the run) \\
Forget:Retain Ratio & $\sim$1:4.5 \\
\bottomrule
\end{tabular}
\end{table}

\subsection{Stopping Criterion: Randomized-Text Perplexity}

We use randomized-text perplexity as an operational stopping rule. For each forget
set, we compare the model's perplexity on the original forget documents against
shuffled variants that preserve token frequencies while destroying semantic and
logical structure:
\begin{enumerate}
    \item \textbf{Baseline Computation:} For each document in $\mathcal{D}_{F}$, we generate a randomized version by splitting the token sequence into segments and shuffling them.
    \item \textbf{Target Perplexity ($\mathrm{PPL}_{\mathrm{target}}$):} We calculate the original model's perplexity on this randomized $\mathcal{D}_{F}$.
    \item \textbf{Termination:} Unlearning stops when the current model's perplexity on the original (unshuffled) forget set reaches or exceeds $\mathrm{PPL}_{\mathrm{target}}$.
\end{enumerate}
Additionally, we apply a safety guard: if the model's performance on a held-out MMLU subset drops below 90\% of the baseline, or if the process reaches 5,000 steps, training terminates immediately.

\subsection{Experimental Conditions (Validation)}

To test whether attribution scores identify more capability-relevant documents
than topic membership alone, we compare two primary conditions within each target
topic, measured against a global random baseline:

\begin{itemize}
    \item \textbf{Single-Topic Influence (Ours):} The forget set consists of the $k=2,000$ documents from the target topic with the highest aggregate influence scores for the target benchmark. The bin-characterization analysis in Appendix~\ref{app:bin-characterization} characterizes which bins carry the strongest SocialIQA influence, and the correctness-differential analysis in Appendix~\ref{app:correctness} identifies bins whose attribution signal reverses across benchmarks.
    \item \textbf{Single-Topic Random:} The forget set consists of $k=2,000$ documents sampled uniformly at random from the \textit{same} target topic. This controls for the effect of topic-specific distribution and tests whether specific document selection yields stronger unlearning.
\end{itemize}

\paragraph{Global Baseline}
As a secondary control to account for procedural degradation (i.e., the inherent noise introduced by the unlearning process itself), we also conduct \textbf{Global Random} experiments. In this condition, the forget set is sampled uniformly from the entire 6T training corpus without regard to topic labels. This baseline allows us to calculate the net effect of topic-specific unlearning: $\gamma_{\mathrm{net}} = \gamma_{\mathrm{topic}} - \gamma_{\mathrm{global}}$.

\subsection{Evaluation Protocol}\label{app:unlearning-eval}

After each unlearning run converges, we merge the LoRA adapter into
OLMo3-7B Base and evaluate the resulting base-unlearned checkpoint on the
same four benchmarks used for attribution (Table~\ref{tab:query-benchmarks}).
We report raw accuracy damage:
\[
\gamma = s_{\mathrm{baseline}} - s_{\mathrm{unlearned}},
\]
where $s_{\mathrm{baseline}}$ is the score of the unmodified OLMo3-7B Base checkpoint.
For accuracy metrics, $\gamma > 0$ indicates capability degradation; $\gamma \approx 0$
indicates the benchmark was preserved; $\gamma < 0$ indicates an accuracy increase after
unlearning. The per-benchmark baseline scores against
which $\gamma$ is computed are reported in Table~\ref{tab:unlearn-baselines}.

\input{tables/tab-unlearn-baselines}

\FloatBarrier

\section{Unlearning Results}\label{app:unlearning-results}

\begin{figure*}[!tb]
  \centering
  \includegraphics[width=\MainFigureWidth]{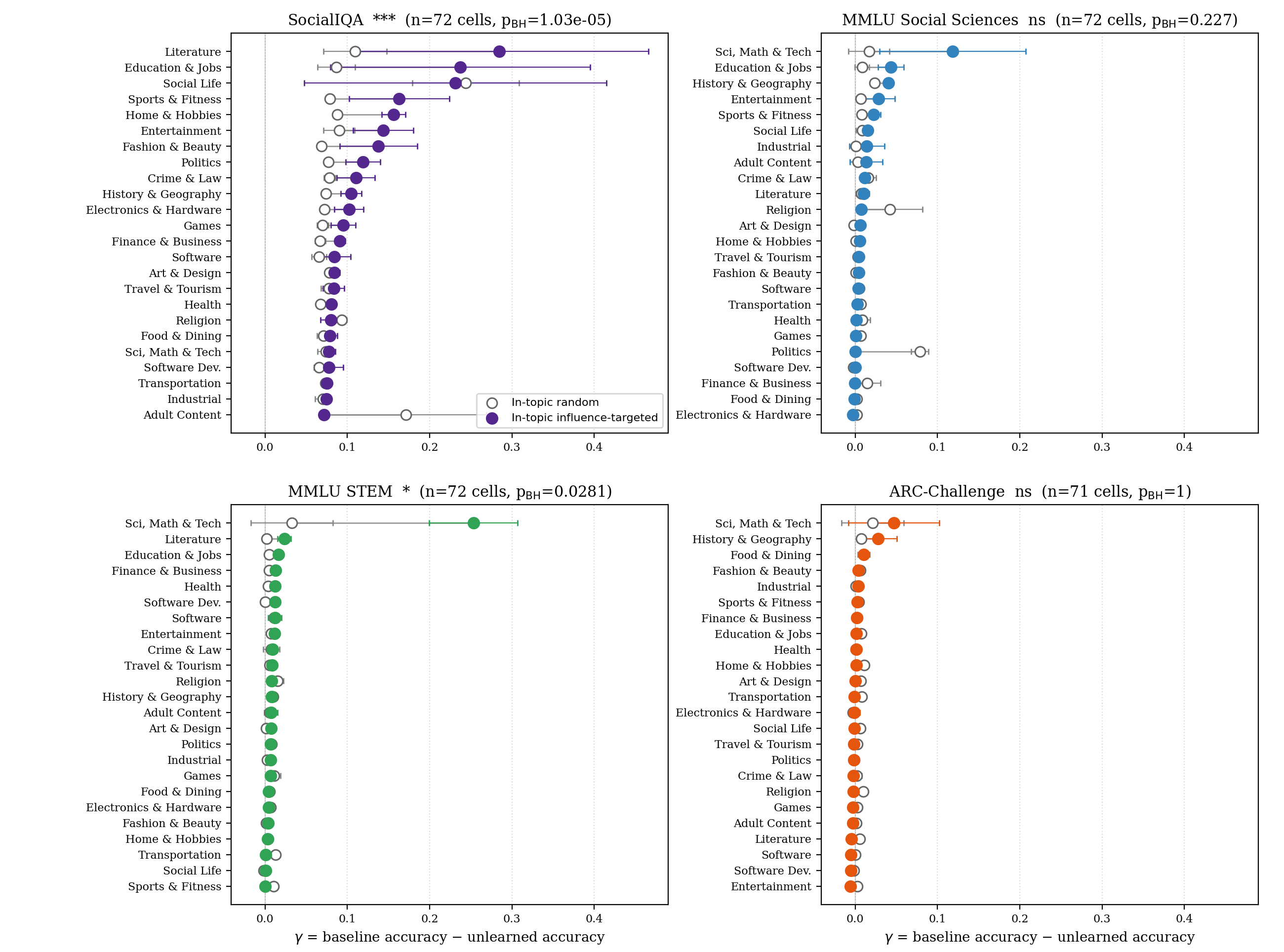}
  \ValidationLegend
  \caption{\ValidationBadge{} Targeted machine unlearning validates bin-level attribution. For each WebOrganizer topic and each primary benchmark, $\gamma_{\mathrm{influence}}$ (colored markers: in-topic, top-200 documents by per-document influence on the target benchmark) is compared against $\gamma_{\mathrm{random}}$ (open markers: in-topic, 1{,}000 random documents, a five-times-larger control forget set), where $\gamma=A_{\mathrm{baseline}}-A_{\mathrm{unlearned}}$ and positive values mean accuracy damage. Markers show the seed-mean across 3 unlearning seeds \{42, 43, 44\}; horizontal whiskers show $\pm 1$ standard deviation. Panel titles report the Benjamini--Hochberg-adjusted Wilcoxon signed-rank significance for the paired contrast across the available matched topic--seed cells (24 topics $\times$ 3 seeds; one ARC-Challenge cell unavailable) (\textsuperscript{*}$p<0.05$, \textsuperscript{**}$p<0.01$, \textsuperscript{***}$p<0.001$). SocialIQA is the clean paired result; MMLU-STEM is weaker because its Wilcoxon test is significant but its bootstrap median interval crosses zero (Appendix~\ref{app:unlearning-results}). The MMLU-SS and ARC-Challenge nulls track the cross-method check in Appendix~\ref{app:selectivity-cross-method}.}
  \label{fig:unlearning-paired}
\end{figure*}

Figure~\ref{fig:unlearning-paired} summarizes the main validation comparison
between influence-targeted and random in-topic unlearning. The remaining
appendix figures break this result down by topic, global control, convergence
efficiency, and influence--drop diagnostics.

\subsection{Paired Significance of the Influence-vs-Random Contrast}\label{app:unlearning-paired-significance}

To put the Figure~\ref{fig:unlearning-paired} pattern on a statistical footing,
we test the one-sided hypothesis that influence-targeted unlearning produces a
larger accuracy drop than random in-topic unlearning on the intended capability.
For each (topic, benchmark, seed) tuple we pair the influence-targeted run
(expA, top-200 documents by per-document influence on the target benchmark) with
the random in-topic run (exp1, same topic and seed, 1,000 documents sampled
uniformly from the same topic),
compute the paired difference $d = \gamma_{\mathrm{influence}} - \gamma_{\mathrm{random}}$,
and aggregate across all \WebOrgNumTopics{} topics and the three unlearning seeds
\{42, 43, 44\}. Per benchmark we run a Wilcoxon signed-rank test on the pooled
(topic\,$\times$\,seed) cells and a paired $t$-test on the per-topic seed-mean
differences; both are one-sided ($d > 0$), and the eight resulting $p$-values
(4 benchmarks $\times$ 2 tests) are Benjamini--Hochberg adjusted within the family.
This faithful per-document analysis is separate from the 2,000-document
single-bin sweep below; it asks whether a small influence-ranked forget set
(200 documents) produces more targeted damage than a five-times-larger random
in-topic control, so the contrast is conservative with respect to forget-set
size.

\paragraph{What uncertainty is measured.}
This paired-significance subsection reports bootstrap 95\% confidence intervals
on the median paired effect $d$ in matched topic--seed cells. Elsewhere in the
selectivity appendix, values written as mean $\pm$ standard deviation summarize
fixed unlearning seeds, not formal confidence intervals. Robustness tables vary
analysis choices such as topic-selection rules and selectivity metrics. We do
not estimate uncertainty over alternative sampled forget-document sets.

Table~\ref{tab:paired-significance} reports the result. The paired contrast is
significantly positive on SocialIQA (Wilcoxon BH-adjusted $p \approx 10^{-5}$,
median $+1.6$ accuracy points, Cohen's $d_z = +0.39$) and on MMLU STEM
($p_{\mathrm{BH}} = 0.028$, $d_z = +0.23$), although MMLU STEM's bootstrap
median interval crosses zero and the paired $t$-test does not reach significance,
so it should be read as weaker evidence. MMLU Social Sciences does not reach
BH-adjusted significance under this paired construction. ARC-Challenge also
does not support the positive one-sided hypothesis, and its median paired
difference is negative, indicating slightly less damage from influence-targeted
forgetting than from random in-topic forgetting. These cases track the
cross-method finding (Appendix~\ref{app:selectivity-cross-method}): the
per-document forget set does not measurably damage MMLU Social Sciences, and ARC-Challenge's
on-target damage sits at the noise floor. We treat both as transparent findings
rather than as failures of the underlying paired-contrast claim.

\input{tables/tab-paired-significance}

\subsection{Single-Bin Topic Sweep}\label{app:unlearning-single-bin}

We run the unlearning pipeline independently for each of the \WebOrgNumTopics{} topic bins.
In each run, the forget set is drawn from a single target topic and the retain set is
stratified across the remaining 23 topics. The checkpoint at the selected stopping
point is evaluated on all four benchmarks.

\begin{figure}[!htbp]
      \centering
      \includegraphics[width=\MainFigureWidth]{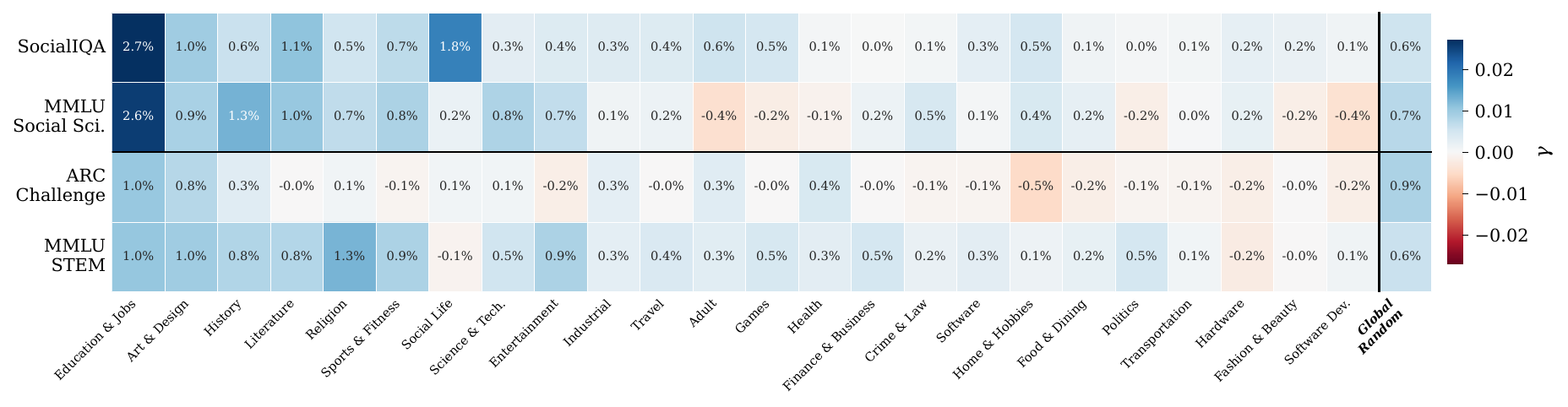}
      \GammaHeatmapSignLegend
      \caption{Cross-benchmark $\gamma$ for single-bin unlearning across all \WebOrgNumTopics{} WebOrganizer topics, with $\gamma=A_{\mathrm{baseline}}-A_{\mathrm{unlearned}}$. Positive values indicate accuracy damage; negative values indicate an accuracy increase after unlearning. The vertical divider separates the topic interventions from the global random control; rows are ordered by the manuscript's 2$\times$2 benchmark design.}
      \label{fig:app-heatmap}
  \end{figure}

Figure~\ref{fig:app-heatmap} presents the per-topic $\gamma$ values. Rows follow the
contrastive design: the two social benchmarks (SocialIQA, MMLU Social Sciences) appear
above the horizontal divider, and the two STEM benchmarks (ARC-Challenge, MMLU STEM)
below it. Topic columns are sorted by mean $\gamma$ across benchmarks, and the vertical
divider separates the topic interventions from the global random control column.

This diagnostic sweep shows modest topic-level effects relative to the paired
influence-vs-random result above. Education \& Jobs has the largest mean
positive damage across the four benchmarks: $+2.7$ points on SocialIQA,
$+2.6$ on MMLU Social Sciences, $+1.0$ on ARC-Challenge, and $+1.0$ on
MMLU STEM. Religion is the largest MMLU STEM drop ($+1.3$ points), while
Art \& Design and History \& Geography also show positive damage on multiple
benchmarks. At the other end, topics such as Software Development,
Transportation, and Fashion \& Beauty show smaller mean $\gamma$, and Adult
Content produces a slight accuracy increase on MMLU Social Sciences in this
single-bin diagnostic view.

\begin{figure}[!htbp]
      \centering
      \includegraphics[width=\MainFigureWidth,height=\TallFigureMaxHeight,keepaspectratio]{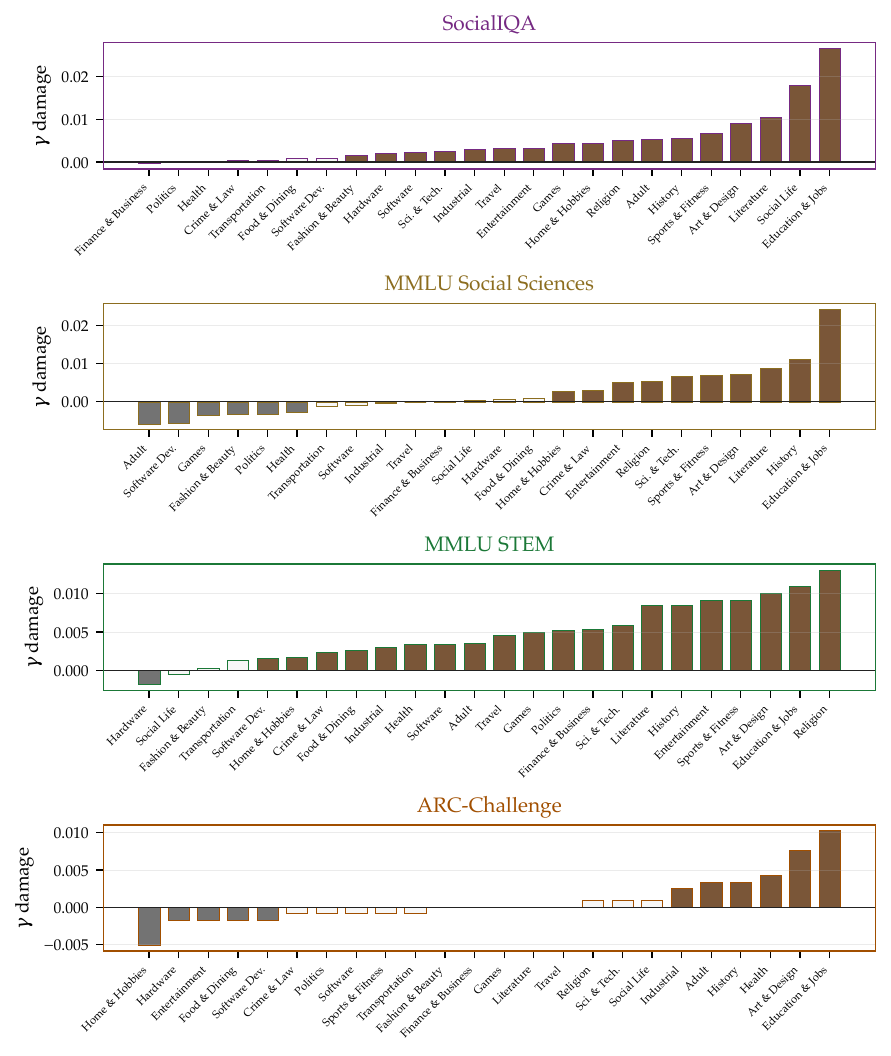}
      \GammaSignLegend
      \caption{Per-benchmark $\gamma$ for each of the \WebOrgNumTopics{} topic bins, sorted independently within each benchmark panel. Panel titles and borders encode benchmark role; muted-brown/gray bars encode positive accuracy damage versus accuracy increases relative to the zero line.}
      \label{fig:app-bars}
  \end{figure}

Figure~\ref{fig:app-bars} shows the per-topic $\gamma$ for each benchmark individually.
Each panel keeps the benchmark-role color in its title and border, while muted-brown/gray bar
fills mark positive accuracy damage versus accuracy increases relative to the zero line. Topics are sorted independently
within each panel. The rank ordering of topic impact is benchmark-dependent: a topic
that strongly degrades SocialIQA does not necessarily degrade MMLU STEM to the same
degree, consistent with the cross-benchmark divergences documented in
Appendix~\ref{app:bin-characterization}.

\subsection{Global Random Control}\label{app:unlearning-control}

To distinguish topic-specific effects from procedural degradation inherent to NGDiff,
we run a global random control where both forget and retain sets are sampled uniformly
from the full corpus without regard to topic labels. The same hyperparameters, stopping
criterion, and evaluation protocol apply.

\begin{figure}[!htbp]
      \centering
      \includegraphics[width=\MainFigureWidth]{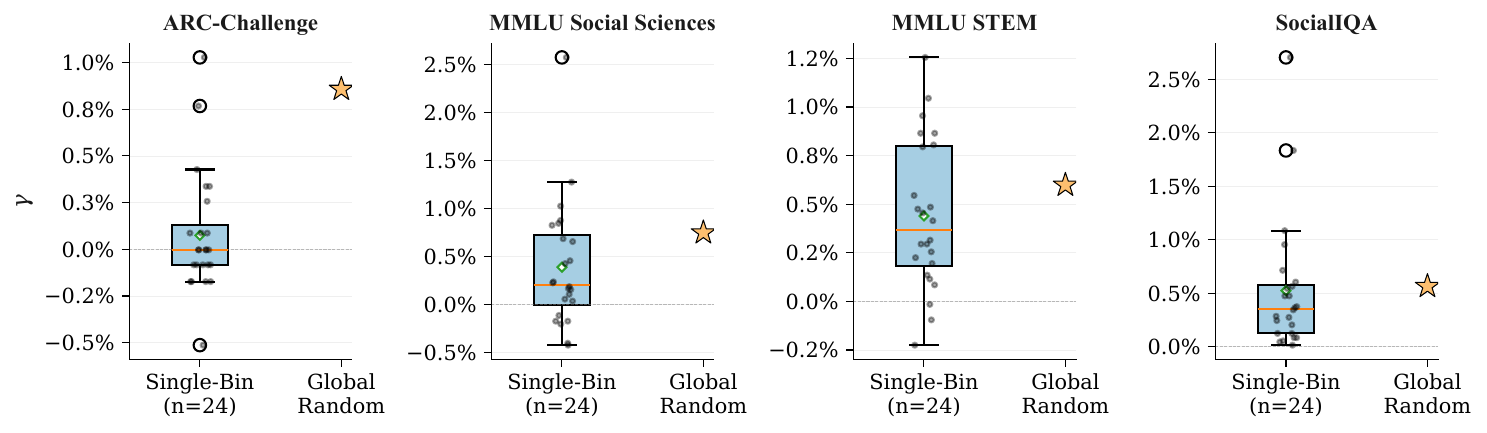}
      \TopicControlLegend
      \caption{Single-bin $\gamma$ distributions (boxplots, $n=\WebOrgNumTopics{}$)
      versus the global random control (star) for each benchmark. The control samples
      forget and retain documents uniformly from the full corpus, controlling for
      procedural degradation. }
      \label{fig:app-control}
  \end{figure}

Figure~\ref{fig:app-control} compares the $\gamma$ distribution from all \WebOrgNumTopics{}
single-bin runs (boxplots) against the global random control (star) for each benchmark.
The control provides a procedural baseline for interpreting the topic-specific
interventions: effects near the control are difficult to separate from ordinary
unlearning degradation, while effects far outside the control range indicate
stronger topic-specific sensitivity. For SocialIQA, MMLU Social Sciences, and
MMLU STEM, the control lies within or close to the upper edge of the single-bin
interquartile range, so the largest topic effects should be read relative to
that background degradation. For ARC-Challenge, the global random control lies
above the single-bin IQR, so the topic-specific ARC effects in this diagnostic
sweep are especially difficult to separate from procedural noise.

\subsection{Topic-Level Efficiency Analysis}\label{app:unlearning-plots}
\begin{figure}[!htbp]
\centering
\includegraphics[width=\MainFigureWidth]{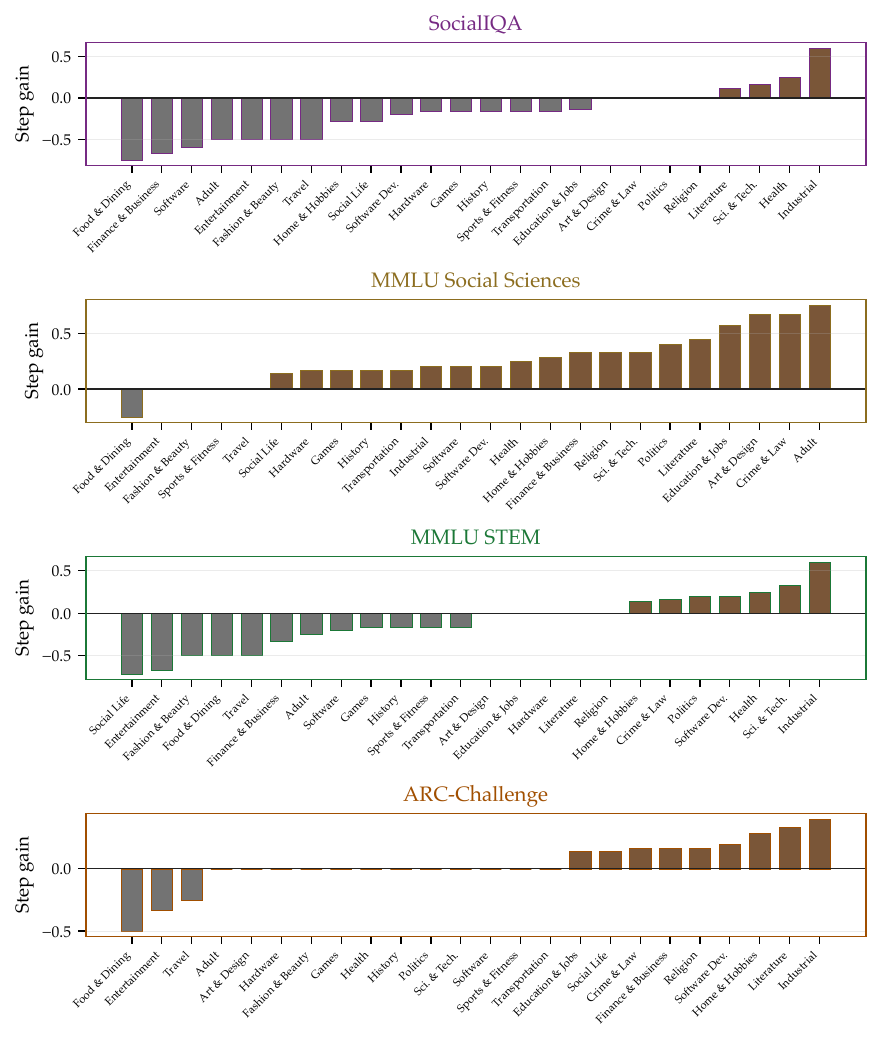}
\EfficiencyLegend
\caption{Topic-level unlearning efficiency gain relative to the in-topic random baseline. Positive values indicate faster convergence with targeted selection; negative values indicate additional steps to reach the randomized-text perplexity threshold. Muted-brown bars mark faster influence-targeted selection and gray bars mark slower outcomes.}
\label{fig:unlearning-efficiency-pct}
\end{figure}

Figure~\ref{fig:unlearning-efficiency-pct} reports convergence speed by topic
and benchmark, showing where influence-guided forget sets reach the stopping
criterion earlier than random in-topic forget sets.
Convergence speed is a different axis from attribution magnitude, so the
fastest-to-unlearn topic need not be the highest-influence one. For SocialIQA,
Industrial shows the largest step gain here; Industrial also happens to lead
the signed topic-level influence marginals
(Figure~\ref{fig:topic-grouped-4way}), but the two rankings answer different
questions and need not coincide in general.

\subsection{Influence--Drop Diagnostic}\label{app:unlearning-scatter-diagnostic}

\begin{figure*}[!tb]
\centering
\includegraphics[width=\MainFigureWidth]{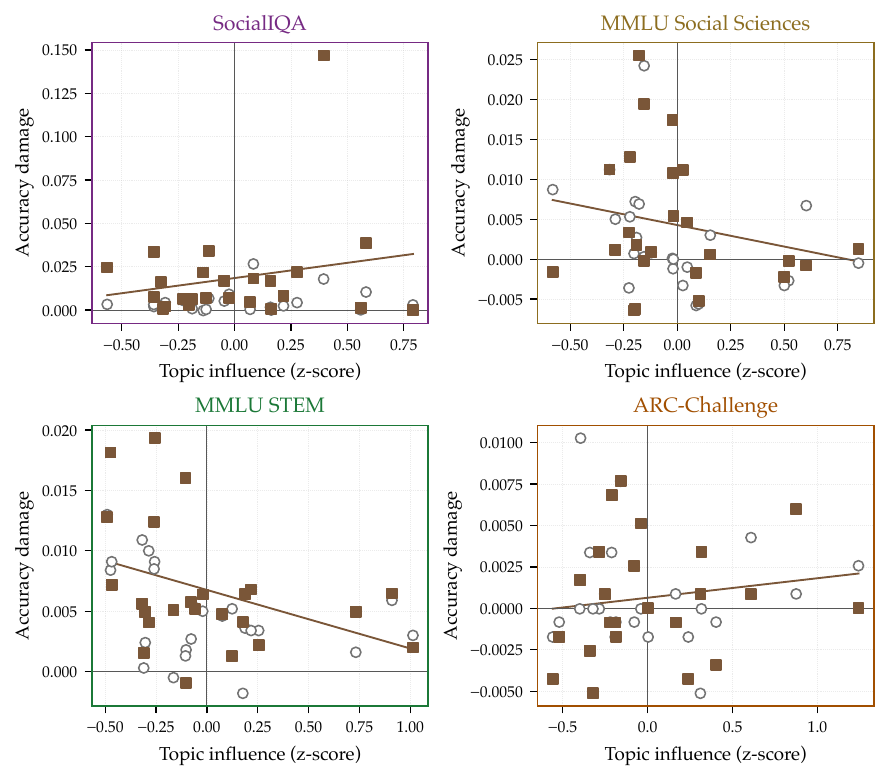}
\ValidationLegend
\caption{Relationship between TrackStar influence scores and raw accuracy damage across four benchmarks. Each point represents a document set ($k{=}2{,}000$), with $\gamma=A_{\mathrm{baseline}}-A_{\mathrm{unlearned}}$. Hollow gray circles denote baseline degradation from random sampling within the target topic bin, while solid muted-brown squares show the impact of selecting high-influence documents. Solid lines indicate the linear regression for the targeted condition.}
\label{fig:unlearning-scatter}
\end{figure*}

Figure~\ref{fig:unlearning-scatter} reports the diagnostic relationship
between TrackStar influence scores and observed accuracy drop. The scatter is
kept in the appendix because the paired unlearning comparison carries the
main validation claim, while this view is useful for checking whether higher
influence is associated with stronger measured degradation.

In this single-bin diagnostic the influence--drop association is weak
throughout: SocialIQA shows a mild positive trend across topics, while the
remaining benchmarks show no consistent positive association. We treat this as
a diagnostic pattern rather than a separate causal claim; the paired
influence-vs-random comparison above carries the validation evidence.

\FloatBarrier

\section{Selective Unlearning: Does Taxonomy Decomposition Help?}\label{app:selectivity}

A natural question about our method is whether decomposing the corpus into a topic taxonomy actually buys anything, or whether one could get the same control by simply unlearning the most-influential documents regardless of topic. This appendix answers that question with direct unlearning experiments. After correcting the SocialIQA baseline and using one gamma convention throughout, the selectivity evidence is mixed: SocialIQA remains favorable, MMLU STEM is favorable only under the faithful per-document replication, MMLU Social Sciences is an agreeing loss, and ARC-Challenge shows no selectivity advantage because the target damage is at the noise floor.

\paragraph{What ``selective'' means.}
We unlearn a set of documents, then measure the accuracy drop $\gamma(b')$ on each benchmark $b'$. For a recipe intended to degrade target benchmark $b$, we define selectivity as the damage on the intended target divided by the average collateral damage on the other three primary benchmarks:
\begin{equation*}
\text{sel}(b) \;=\; \frac{|\gamma(b)|}{\tfrac{1}{3}\sum_{b'\neq b}|\gamma(b')|}.
\end{equation*}
A selectivity of 1 means damage is spread evenly; higher values mean the intended capability is degraded while the others are spared.

\begin{SocialTDACallout}{Reading selectivity results}
Values above $1$ mean the intended benchmark takes more damage than the
average collateral benchmark. The strongest evidence is therefore a large
drop on the intended benchmark paired with low collateral damage across the
other three cells of the 2$\times$2 benchmark design.
\end{SocialTDACallout}

\paragraph{Experimental design.}
We compare four unlearning recipes that form a $2\times2$ factorial over the two ingredients of bin-targeted unlearning: \emph{which documents to forget} (the most-influential ones, ranked by training-data attribution, versus a random sample) and \emph{whether to restrict the forget set to one topic} (a single WebOrganizer topic versus the whole corpus).

\begin{center}
\begin{tabular}{lll}
\toprule
 & one topic & whole corpus \\
\midrule
influence-ranked & bin-targeted (our method) & na\"ive top-$K$ \\
random           & random-in-topic           & random-global (baseline) \\
\bottomrule
\end{tabular}

\end{center}

All four are realized unlearning runs (NGDiff~\citep{bu2025ngdiff} with a rank-8 LoRA adapter, $k{=}2{,}000$ forget documents each, identical configuration; only the forget-set selection differs), evaluated under the same OLMES protocol as our primary suite. Unless a table states otherwise, the numbers below are measured raw accuracy damage values or ratios derived from them.

\paragraph{What uncertainty is measured.}
The $\pm$ values in the faithful per-document selectivity tables are standard
deviations across fixed unlearning seeds. They estimate optimizer-run
variability, not confidence intervals over alternative forget-document samples.
The paired-significance table in Appendix~\ref{app:unlearning-paired-significance}
is the source for bootstrap 95\% intervals on the matched influence-vs-random
median effect. The robustness rows below are sensitivity checks: they vary
analysis choices over the same training runs.

\paragraph{Result 1: the strict blind top-1 factorial is mixed.}
Table~\ref{tab:selectivity-factorial} reports the $2\times2$. Comparing the two influence-ranked recipes isolates the value of \emph{topic filtering}; under the corrected OLMES gamma matrix, this helps on 0 of the 3 scorable benchmarks at the blind top-1 attribution topic. Comparing the two topic-filtered recipes isolates the value of \emph{influence ranking}; this helps on 1 of 3. The strict blind top-1 factorial therefore does not support a uniform selectivity claim. It is still useful diagnostically: the MMLU pair fails at the Industrial top-1 topic, while ARC-Challenge sits at the noise floor ($\gamma \approx 0$) so its ingredient ratios are uninterpretable and it is excluded from the counts.

\paragraph{Result 2: bin-targeted vs.\ na\"ive top-$K$ head-to-head.}
Table~\ref{tab:selectivity} reports the direct head-to-head between bin-targeted unlearning and na\"ive top-$K$ under the corrected Path-B analysis, using the simplest possible way of choosing which topic to target: pick the single topic with the highest attribution score for the benchmark. Under this attribution-only choice and the config-matched OLMES \texttt{:mc::olmes} baselines, bin-targeted unlearning is more selective than na\"ive top-$K$ on 1 of the 3 scorable benchmarks: SocialIQA. The two losses, MMLU STEM and MMLU Social Sciences, both occur at the Industrial top-1 topic; ARC-Challenge sits at the noise floor ($\gamma \approx 0$) and its head-to-head ratio is not interpretable.

\paragraph{Pilot topic choice changes the picture, but not uniformly.}
Because the attribution score is only a \emph{predictor} of unlearning effect, not a measurement of it, the simplest topic choice leaves selectivity on the table. We therefore evaluate topic-selection procedures in Table~\ref{tab:selectivity-robust}. The corrected cross-method results do not recover a uniform four-benchmark win.

\begin{itemize}[leftmargin=1.2em, topsep=2pt, itemsep=2pt]
\item \textbf{Corrected bin-level pilot.} The refreshed bin-level single-seed pilot is favorable on SocialIQA ($1.43\times$) but not on ARC-Challenge ($0.82\times$), MMLU STEM ($0.62\times$), or MMLU Social Sciences ($0.77\times$).
\item \textbf{Faithful per-document pilot.} The faithful replicated pipeline is favorable on SocialIQA ($2.17 \pm 0.61\times$) and MMLU STEM ($1.45 \pm 0.10\times$), loses on MMLU Social Sciences ($0.15 \pm 0.02\times$), and shows no advantage on ARC-Challenge ($0.87 \pm 0.24\times$), whose on-target damage is at the noise floor.
\item \textbf{Best-in-hindsight (upper bound).} If the unlearning effect of every topic were known, picking the most selective one is favorable on all four faithful-pipeline targets. This is an upper bound, not a deployable selection rule, and should not be used as the main result.
\end{itemize}

The gap between attribution-only selection and the pilot rows is itself a finding: attribution scores predict \emph{which} documents matter but are imperfect predictors of \emph{how much} unlearning them degrades a capability.

\paragraph{Robustness.}
The faithful per-document robustness sweep in Table~\ref{tab:selectivity-robust} is also mixed. SocialIQA is favorable under the mean, worst-case, and excluding-ARC ratio metrics, but not under the signed ratio. MMLU STEM is favorable in the faithful pilot row but not at the highest-attribution topic across the six metric definitions. MMLU Social Sciences is a consistent loss in the ratio metrics, despite winning under the entropy concentration diagnostic. ARC-Challenge ratios sit at or below $1\times$ across the deployable metrics and are not interpretable because the target damage is at the noise floor. Full numbers and the per-metric sweep are in Table~\ref{tab:selectivity-robust}.

\paragraph{Limitations.}
\textit{Method-dependent benchmarks.} The faithful per-document pipeline supports SocialIQA and MMLU STEM but not MMLU Social Sciences. ARC-Challenge remains difficult to interpret because selectivity ratios are sensitive when on-target damage is near zero. We report these cases directly in the cross-method check (\S\ref{app:selectivity-cross-method}) rather than treating them as artifacts to suppress.
\textit{Topic vs.\ bin granularity.} Our unlearning is implemented per WebOrganizer topic (24 levels); the selectivity gains may be larger at the finer (topic\,$\times$\,format) bin granularity that the attribution profiles (\S\ref{sec:results}) operate on, but per-bin unlearning is left to future work.
\textit{Target-topic choice.} The current audit shows that topic choice is load-bearing. A purely predictive selection rule that matches the best empirical pilot remains open.

\subsection{Cross-Method Check: Bin-Level vs.\ Per-Document Forget-Sets}\label{app:selectivity-cross-method}

The corrected bin-level pilot above is computed under a single forget-set-construction method (bin-level z-scoring at topic\,$\times$\,format granularity, single seed). To distinguish methodological advantages of bin-targeting from artifacts of that specific construction, we re-run the pilot under a second, independently-implemented forget-set construction: \emph{faithful per-document} selection, in which the forget set for each (target, candidate topic) pair is the top-200 documents by per-document influence on the target benchmark, replicated across the available unlearning seeds. Both methods use the identical 24-topic grid, the identical pilot-validation procedure (shortlist 3 candidate topics by unique-z, run a $\gamma$ pilot on each, keep the most selective), and the identical OLMES \texttt{:mc::olmes} baselines and evaluation protocol; only the forget-set construction differs.

Table~\ref{tab:selectivity-cross-method} reports the cross-method comparison. The four targets sort into four distinct patterns.

\begin{itemize}[leftmargin=1.2em, topsep=2pt, itemsep=2pt]
\item \textbf{SocialIQA.} Bin-level pilot $1.43\times$ and faithful pilot $2.17 \pm 0.61\times$ agree in direction. This remains the clearest favorable selectivity result.
\item \textbf{MMLU STEM.} The bin-level pilot loses ($0.62\times$), while the faithful per-document pilot wins ($1.45 \pm 0.10\times$). This is favorable only under the faithful forget-set construction.
\item \textbf{MMLU Social Sciences.} The bin-level pilot ($0.77\times$) and faithful per-document pilot ($0.15 \pm 0.02\times$) both lose. This is an agreeing loss under the current cross-method comparison.
\item \textbf{ARC-C.} Neither construction shows a selectivity advantage (bin-level $0.82\times$, faithful $0.87 \pm 0.24\times$). Because on-target damage is at the noise floor ($\gamma \approx 0$), the ratio is not interpretable; we treat ARC-Challenge as a noise-floor non-result.
\end{itemize}

\paragraph{Reframed headline.} The cross-method-aware reading is: SocialIQA is favorable in both constructions; MMLU STEM is faithful-only favorable; MMLU Social Sciences is an agreeing loss; and ARC-Challenge is a noise-floor non-result (no selectivity advantage under either construction). This is the reading used throughout the paper.

\begin{table}[!htbp]
\centering
\scriptsize
\caption{Selectivity of four empirical unlearning recipes, forming a $2\times2$ factorial that separates the two ingredients of bin-targeted unlearning: \emph{influence ranking} (which documents to forget) and \emph{topic filtering} (whether to restrict the forget set to one WebOrganizer topic). Each cell is a realized NGDiff unlearning run; we report selectivity $\text{sel}(b){=}|\gamma(b)|/\mathrm{mean}_{b'\neq b}|\gamma(b')|$ (damage on the intended benchmark relative to mean collateral on the other three), evaluated at the top-1 attribution topic for each target benchmark $b$ under the config-matched OLMES \texttt{:mc::olmes} protocol. Higher is more selective. The two right-hand columns isolate each ingredient: the \emph{influence effect} compares the two topic-filtered recipes (top-left vs.\ bottom-left), and the \emph{topic effect} compares the two influence-ranked recipes (top-left vs.\ top-right).}\label{tab:selectivity-factorial}
\setlength{\tabcolsep}{2pt}
\renewcommand{\arraystretch}{1.1}
\begin{tabularx}{\linewidth}{@{}>{\raggedright\arraybackslash}p{0.16\linewidth}>{\centering\arraybackslash}p{0.15\linewidth}>{\centering\arraybackslash}X>{\centering\arraybackslash}p{0.11\linewidth}>{\centering\arraybackslash}p{0.11\linewidth}>{\centering\arraybackslash}p{0.11\linewidth}@{}}
\toprule
& \multicolumn{3}{c}{\textbf{Selectivity by recipe}} & \multicolumn{2}{c}{\textbf{Ingredient effect}} \\
\cmidrule(lr){2-4}\cmidrule(lr){5-6}
\rowcolor{SocialTDALightGray}
Target $b$ & \cellcolor{SocialTDASlate!8}\makecell{Bin-targeted\\(influence + topic)} & \makecell{Na\"ive top-$K$\\(influence, no topic)} & \makecell{Random\\in-topic} & \makecell{Influence\\effect} & \makecell{Topic\\effect} \\
\midrule
\BenchmarkTargetTableCell{SocialTDAPurple}{SocialIQA} & \cellcolor{SocialTDASlate!8}8.63 & \textbf{14.37} & 11.58 & 0.75$\times$ & 0.60$\times$ \\
\BenchmarkTargetTableCell{SocialTDAOrange}{ARC-Challenge} & \cellcolor{SocialTDASlate!8}1.11 & 1.34 & 0.03 & n/a ($\gamma{\approx}0$) & n/a ($\gamma{\approx}0$) \\
\BenchmarkTargetTableCell{SocialTDAGreen}{MMLU STEM} & \cellcolor{SocialTDASlate!8}0.04 & \textbf{0.78} & 0.10 & 0.38$\times$ & 0.05$\times$ \\
\BenchmarkTargetTableCell{SocialTDAGold}{MMLU Social Sciences} & \cellcolor{SocialTDASlate!8}0.06 & \textbf{0.88} & 0.04 & \textbf{1.33$\times$} & 0.06$\times$ \\
\midrule
\textit{count helping} & --- & --- & --- & 1/3 & 0/3 \\
\bottomrule
\end{tabularx}
\vspace{2pt}
\begin{minipage}{0.95\linewidth}
\scriptsize
\textbf{Influence effect} $=$ selectivity(bin-targeted)\,/\,selectivity(random-in-topic): isolates the value of ranking documents by influence, holding the topic fixed. \textbf{Topic effect} $=$ selectivity(bin-targeted)\,/\,selectivity(na\"ive top-$K$): isolates the value of restricting to one topic, holding influence-ranking fixed. Under the corrected OLMES gamma matrix, influence ranking helps on 1 of the 3 scorable benchmarks and topic filtering on 0 at the strict blind top-1 topic. ARC-Challenge is excluded from the ingredient ratios because its on-target damage is at the noise floor ($\gamma \approx 0.003$), which makes the near-zero/near-zero ratios uninterpretable. Both MMLU Social Sciences and MMLU STEM topic effects are below 1.0 because the blind top-1 attribution topic (Industrial) is a weak unlearning target for both. The pilot and faithful per-document checks in Table~\ref{tab:selectivity-robust} should therefore be read as a mixed robustness audit, not as a uniform four-benchmark win.
\end{minipage}
\end{table}

\begin{table}[!htbp]
\centering
\scriptsize
\caption{Selectivity of bin-targeted vs.\ na\"ive top-$K$ unlearning under the corrected \textbf{Path B} analysis ($k{=}2{,}000$ documents per condition). For target benchmark $b$, the damage ratio $\text{sel}_c(b){=}|\gamma_{c,b}(b)|/\mathrm{mean}_{b'\neq b}|\gamma_{c,b}(b')|$ measures how concentrated the unlearning damage is on the intended target relative to the mean magnitude of collateral damage on the other three primary benchmarks (higher~$=$~more selective). Bin-targeted uses the single-topic intervention at each benchmark's top-1 topic $k_\star$ (parent topic of the max-$z$ bin in z-scored bin influence). Na\"ive top-$K$ is the Path-B linear prediction $\gamma_{\mathrm{naive},b}(b')=\sum_k p_b[k]\gamma_{b,k}(b')$, where $p_b[k]$ is the empirical topic distribution of the global top-$2{,}000$ documents ranked by influence on $b$. All $\gamma$ values are raw accuracy damage under the config-matched OLMES \texttt{:mc::olmes} protocol.}\label{tab:selectivity}
\setlength{\tabcolsep}{3pt}
\renewcommand{\arraystretch}{1.08}
\begin{tabularx}{\linewidth}{@{}>{\raggedright\arraybackslash}p{0.18\linewidth}
  >{\raggedright\arraybackslash}X
  >{\centering\arraybackslash}p{0.12\linewidth}
  >{\centering\arraybackslash}p{0.13\linewidth}
  >{\centering\arraybackslash}p{0.10\linewidth}@{}}
\toprule
\rowcolor{SocialTDALightGray}
Target $b$ & Top-1 topic--format bin &
\cellcolor{SocialTDASlate!8}\makecell{$\text{sel}_{\text{bin}}$\\bin-targeted} &
\makecell{$\text{sel}_{\text{na\"ive}}^{\text{emp}}$\\na\"ive top-$K$} &
Ratio \\
\midrule
\BenchmarkTargetTableCell{SocialTDAPurple}{SocialIQA} &
Literature $\times$ Customer Support &
\cellcolor{SocialTDASlate!8}\textbf{1.28} & 0.64 & \textbf{1.99$\times$} \\
\BenchmarkTargetTableCell{SocialTDAOrange}{ARC-Challenge} &
Industrial $\times$ Documentation &
\cellcolor{SocialTDASlate!8}1.11 & 0.36 & n/a ($\gamma{\approx}0$) \\
\midrule
\BenchmarkTargetTableCell{SocialTDAGreen}{MMLU STEM} &
Industrial $\times$ Documentation &
\cellcolor{SocialTDASlate!8}0.10 & 0.28 & 0.36$\times$ \\
\BenchmarkTargetTableCell{SocialTDAGold}{MMLU Social Sciences} &
Industrial $\times$ Documentation &
\cellcolor{SocialTDASlate!8}0.06 & 0.18 & 0.34$\times$ \\
\bottomrule
\end{tabularx}
\vspace{2pt}
\begin{minipage}{0.95\linewidth}
\scriptsize
Top two rows: bin-targeting concentrates damage more than na\"ive top-$K$ on one of the three scorable benchmarks (SocialIQA) under the corrected Path-B top-1 rule; ARC-Challenge sits at the noise floor ($\gamma{\approx}0$) and its ratio is not interpretable. Bottom two rows: MMLU STEM and MMLU Social Sciences lose under this attribution-only choice because their highest-attribution bin (Industrial $\times$ Documentation) is a weak unlearning target for both. The pilot and cross-method checks in Tables~\ref{tab:selectivity-robust} and~\ref{tab:selectivity-cross-method} give the current mixed reading rather than a uniform four-benchmark win. Sign convention: $\gamma > 0$ indicates accuracy drop on the named benchmark after unlearning; absolute values are taken for the selectivity ratio so help (negative $\gamma$) and hurt (positive $\gamma$) on non-target benchmarks both count as collateral.
\end{minipage}
\end{table}

\input{tables/tab-selectivity-robustness}

\input{tables/tab-selectivity-cross-method}

\FloatBarrier

\subsection{ARC Unlearning Across Task Definitions (Coverage Robustness)}\label{app:arc-rescore-3var}

The ARC attribution and its bin-level forget-sets are computed over the full
$5{,}678{,}621$-document working set, matching the other primary benchmarks. To
check that the near-zero ARC on-target effect is a property of the benchmark
rather than an artifact of the earlier partial-coverage attribution, we re-run
bin-level unlearning under three ARC definitions: ARC-Easy, ARC-Challenge, and
ARC-Combined (the union of the two query sets). Each definition uses the same
NGDiff + rank-8 LoRA pipeline, the same 24-topic grid plus a corpus-wide recipe,
and the same config-matched OLMES \texttt{:mc::olmes} baselines.

\begin{figure}[!htbp]
\centering
\includegraphics[width=0.62\linewidth]{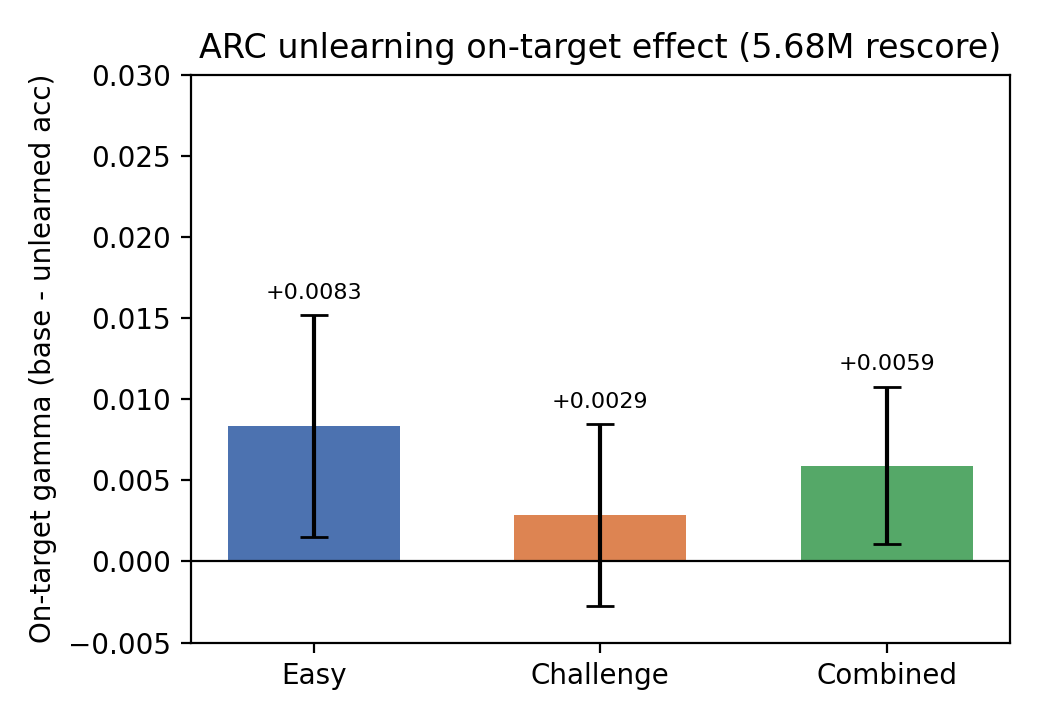}
\caption{On-target $\gamma$ (base $-$ unlearned accuracy) for ARC unlearning under
three task definitions on the full $5.68$M working set, mean\,$\pm$\,std over 25
recipes. All three sit at the noise floor.}\label{fig:arc-rescore-3var}
\end{figure}

Table~\ref{tab:arc-rescore-3var} and Figure~\ref{fig:arc-rescore-3var} report the
result. On-target $\gamma$ stays below one accuracy point under all three
definitions (ARC-Easy $+0.0083 \pm 0.0068$, ARC-Challenge
$+0.0029 \pm 0.0056$, ARC-Combined $+0.0059 \pm 0.0049$), with ARC-Challenge
itself within one standard deviation of zero, while the held-out benchmarks are
preserved. The noise-floor reading of ARC-Challenge therefore
holds across task definitions and under the corrected full-coverage attribution;
it is not produced by the earlier partial scoring.

\begin{table}[!htbp]
\centering
\small
\caption{ARC unlearning on-target effect across three task definitions on the full $5{,}678{,}621$-document working set. Bin-level NGDiff + rank-8 LoRA unlearning is run per WebOrganizer topic (24 topics + corpus-wide) for each ARC definition: ARC-Easy, ARC-Challenge, and ARC-Combined (union of the two query sets). On-target $\gamma = (\text{base acc} - \text{unlearned acc})$, mean\,$\pm$\,std over the 25 recipes; positive $\gamma$ means accuracy dropped after unlearning. Retain columns report mean $\gamma$ on the held-out benchmarks. All three definitions leave the on-target benchmark essentially unchanged, confirming the noise-floor reading is not an artifact of the earlier partial-coverage attribution.}\label{tab:arc-rescore-3var}
\setlength{\tabcolsep}{4pt}
\renewcommand{\arraystretch}{1.1}
\begin{tabular}{lcccccc}
\toprule
\rowcolor{SocialTDALightGray}
ARC definition & $n$ & On-target $\gamma$ & Base acc & SocialIQA $\gamma$ & MMLU-STEM $\gamma$ & MMLU-SS $\gamma$ \\
\midrule
ARC-Easy      & 25 & $+0.0083 \pm 0.0068$ & 0.942 & $+0.0206$ & $+0.0032$ & $+0.0026$ \\
ARC-Challenge & 25 & $+0.0029 \pm 0.0056$ & 0.834 & $+0.0184$ & $+0.0034$ & $+0.0008$ \\
ARC-Combined  & 25 & $+0.0059 \pm 0.0049$ & 0.888 & $+0.0189$ & $+0.0035$ & $+0.0022$ \\
\bottomrule
\addlinespace[3pt]
\multicolumn{7}{@{}p{\linewidth}@{}}{\footnotesize On-target $\gamma$ is the mean over 24 per-topic recipes plus the corpus-wide recipe; for ARC-Combined the on-target accuracy is the mean of ARC-Easy and ARC-Challenge. Baselines are the same-pipeline OLMo3-7B Base checkpoint under \texttt{:mc::olmes}. Across all three definitions the on-target effect stays below one accuracy point; ARC-Challenge itself is within one standard deviation of zero.} \\
\end{tabular}
\end{table}

\FloatBarrier

\section{Compute and Resource Accounting}\label{app:compute}

\subsection{Taxonomy Enrichment}\label{app:taxonomy-enrichment}

Classifying \CorpusUniqueDocsApprox{} documents with the official WebOrganizer
classifiers~\citep{wettig2025weborganizer} required processing \CorpusNumShards{}
shards across a mixed-tier HPC GPU cluster. Per-shard task wall-clock and
allocated GPU type were instrumented at run time; aggregated totals
appear in Table~\ref{tab:compute-summary}.

\paragraph{Classifier accuracy.}
On the validation set used by \citet{wettig2025weborganizer} (10k web pages,
$\geq$75\% teacher confidence), the topic and format classifiers achieve
93.5\% and 91.8\% average accuracy respectively (87.1\% and 80.5\%
worst-group). The 405B teacher's own self-agreement under random prompt
ordering is 97--98\%, placing classifier-induced error at ${\sim}$4--5\%
above the annotation-noise floor. Per-classifier throughput by GPU tier is
itemized in Table~\ref{tab:enrichment-throughput}.

\input{tables/tab-enrichment-throughput}

\subsection{Attribution Pipeline}

We use Bergson's index-reuse pipeline: training gradients are computed
once on GPU and stored as a persistent index, then scored against any number of
query sets via \emph{CPU} dot products. The one-time index build amortizes over
all subsequent benchmark scoring passes. The attribution pipeline thus
decomposes into four distinct GPU-bound sub-phases plus a separate CPU-bound
scoring stage:

\begin{enumerate}[label={},leftmargin=1em,topsep=2pt,itemsep=0pt]
  \item \textbf{2a. Training gradient index} (\ComputeAttrIndexEquiv{} GPU-hr):
    per-document gradients for the \AttrDocCount{}-document
    stratified working set (${\sim}$\AttrShardCount{} shards $\times$
    \AttrDocsPerShard{} docs), built once on H100 80\,GB and
    H200 144\,GB across two HPC allocations.
  \item \textbf{2b. Query gradient indexes} (\ComputeAttrQueryEquiv{} GPU-hr):
    model-specific query gradient indexes, in base and instruct variants,
    for the four primary benchmarks (SocialIQA, MMLU Social Sciences,
    ARC-Challenge, MMLU~STEM) and the additional diagnostic probe suites,
    including the instruct-query builds used for reported two-model scoring.
  \item \textbf{2c. Preconditioner construction} (\ComputeAttrPrecondEquiv{}
    GPU-hr): TrackStar mixed preconditioner artifacts for the reported
    scoring runs, with base-side curvature as the pretraining-data
    metric and supervised-instruction-tuning curvature treated as identity.
    This identity treatment is a modeling approximation that keeps the
    reported metric anchored to pretraining-data curvature rather than adding
    an unobserved post-training curvature term.
    Each build uses 8$\times$ H100 80\,GB with FSDP over a
    \PrecondSampleSize-document sample.
  \item \textbf{2d. Calibration and exploratory runs}
    (\ComputeAttrCalibEquiv{} GPU-hr): smoke tests, calibration baselines
    (\AttrGPUHoursExplore{} GPU-hr without preconditioning), and
    preconditioner-hyperparameter sweeps that informed the final pipeline.
  \item \textbf{2e. Production scoring (CPU)}: each benchmark-vs-index
    scoring pass runs as a parallel CPU array, so no GPU-hours apply
    to this stage.
\end{enumerate}

\paragraph{Training index storage.}
The gradient index requires ${\sim}$8\,KB per document. For the working set
(\AttrDocCount{} documents), this is ${\sim}$44\,GB.

\subsection{Unlearning Experiments}

Each single-bin unlearning run trains for up to 5{,}000 steps on a single
\textbf{H200 144\,GB} GPU. The 24-topic single-bin sweep
($24$ topics $\times$ $3$ chained jobs $=72$ runs) is the largest
intervention cost; influence-guided and random-baseline conditions
roughly double it. Additional cost comes from multi-bin runs,
hyperparameter sweeps, null-bin controls, and per-checkpoint
evaluation passes (every 250 steps across 4 benchmarks). A subset of
single-topic runs was executed on rented GPU infrastructure off the
primary HPC environment; those hours are included in the totals above.

\subsection{Aggregate Compute Summary}

Total compute across all phases was approximately \ComputeTotalRaw{}
raw GPU-hours (\ComputeTotalEquiv{} \ComputeRefGPU-equivalent after
FLOPS normalization). Table~\ref{tab:compute-summary} reports the
per-phase breakdown. The raw column sums heterogeneous accelerator
time; the H200-equivalent column applies bf16 FLOPS ratios per phase
to support like-for-like comparison with other corpus-scale attribution
works. We round \emph{up} on partial hours to avoid under-reporting. The totals above account for GPU-bound
work only; substantial CPU compute for corpus preparation,
de-duplication, and classifier-inference orchestration was a non-trivial
share of the overall research effort but is tracked separately and not
folded into the GPU-hour totals.

\input{tables/tab-compute-summary}

\FloatBarrier

\section{LLM Disclosure}

Large language models were used during the preparation of this manuscript for
drafting and revising prose, editing LaTeX, writing analysis and visualization
code, and managing project coordination. All LLM-generated content was reviewed
and verified by the authors. LLMs were not used to generate data. Research
questions, experimental design, and original ideas are entirely the work of the
authors.

\stopcontents[appendix]

\end{document}